\documentclass{article}

\usepackage{amsmath,amsfonts,bm}

\def\eqref#1{equation~\ref{#1}}

\def\1{\bm{1}}

\def\eps{{\epsilon}}

\def\va{{\bm{a}}}
\def\vb{{\bm{b}}}
\def\vc{{\bm{c}}}

\def\vv{{\bm{v}}}
\def\vw{{\bm{w}}}
\def\vx{{\bm{x}}}

\def\vz{{\bm{z}}}

\def\mA{{\bm{A}}}

\def\mV{{\bm{V}}}
\def\mW{{\bm{W}}}

\DeclareMathAlphabet{\mathsfit}{\encodingdefault}{\sfdefault}{m}{sl}
\SetMathAlphabet{\mathsfit}{bold}{\encodingdefault}{\sfdefault}{bx}{n}

\def\sB{{\mathbb{B}}}

\def\sH{{\mathbb{H}}}

\def\sV{{\mathbb{V}}}

\def\sX{{\mathbb{X}}}

\usepackage{latexsym,eucal,bbm,color}
\usepackage{url}
\usepackage{comment}
\usepackage{float}
\usepackage{tcolorbox}
\usepackage{booktabs}
\usepackage{blindtext}
\usepackage{multirow}
\usepackage{booktabs}
\usepackage{array}
\usepackage{soul}
 
\usepackage[normalem]{ulem}
\usepackage{stackengine}
\usepackage{parskip}
\usepackage{bm}
\usepackage{balance}
\usepackage{comment}
\usepackage{soul}
\usepackage{pgffor}

\usepackage{caption}

\def \ba{\boldsymbol{a}}

\def \bV{\boldsymbol{V}}

\def\Indic{\mathbbm{1}}

\usepackage{listings}

\definecolor{codegreen}{rgb}{0,0.6,0}
\definecolor{codegray}{rgb}{0.5,0.5,0.5}
\definecolor{codepurple}{rgb}{0.58,0,0.82}
\definecolor{backcolour}{rgb}{0.95,0.95,0.92}

\lstdefinestyle{mystyle}{
  backgroundcolor=\color{backcolour}, commentstyle=\color{codegreen},
  keywordstyle=\color{magenta},
  numberstyle=\tiny\color{codegray},
  stringstyle=\color{codepurple},
  basicstyle=\ttfamily\footnotesize,
  breakatwhitespace=false,         
  breaklines=true,                 
  captionpos=b,                    
  keepspaces=true,                 
  numbers=left,                    
  numbersep=5pt,                  
  showspaces=false,                
  showstringspaces=false,
  showtabs=false,                  
  tabsize=2
}

\usepackage{microtype}
\usepackage{graphicx}
\usepackage{subfigure}
\usepackage{booktabs} 
\usepackage{comment}

\usepackage{hyperref}

\usepackage[accepted]{icml2024}

\usepackage{amsmath}
\usepackage{amssymb}
\usepackage{mathtools}
\usepackage{amsthm}

\usepackage[capitalize,noabbrev]{cleveref}

\theoremstyle{plain}

\theoremstyle{definition}

\theoremstyle{remark}

\usepackage[textsize=tiny]{todonotes}

\icmltitlerunning{Deep Networks Always Grok and Here is Why}

\newcounter{todocounter}

\begin{document}

\twocolumn[
\icmltitle{Deep Networks Always Grok and Here is Why}




\begin{icmlauthorlist}
\icmlauthor{Ahmed Imtiaz Humayun}{xxx}
\icmlauthor{Randall Balestriero}{yyy}
\icmlauthor{Richard Baraniuk}{xxx}
\end{icmlauthorlist}

\icmlaffiliation{xxx}{Rice University}
\icmlaffiliation{yyy}{Brown University}

\icmlcorrespondingauthor{Ahmed Imtiaz Humayun}{imtiaz@rice.edu}

\icmlkeywords{grokking}

\vskip 0.3in
]



\printAffiliationsAndNotice{}  

\begin{abstract}
Grokking, or {\em delayed generalization}, is a phenomenon where generalization in a deep neural network (DNN) occurs long after achieving near zero training error. Previous studies have reported the occurrence of grokking in specific controlled settings, such as DNNs initialized with large-norm parameters or transformers trained on algorithmic datasets. We demonstrate that grokking is actually much more widespread and materializes in a wide range of practical settings, such as training of a convolutional neural network (CNN) on CIFAR10 or a Resnet on Imagenette. We introduce the new concept of {\em delayed robustness}, whereby a DNN groks adversarial examples and becomes robust, long after interpolation and/or generalization. We develop an analytical explanation for the emergence of both delayed generalization and delayed robustness based on the {\em local complexity} of a DNN's input-output mapping. Our \textit{local complexity} measures the density of so-called ``linear regions’’ (aka, spline partition regions) that tile the DNN input space and serves as a utile progress measure for training. We provide the first evidence that, for classification problems, the linear regions undergo a phase transition during training whereafter they migrate away from the training samples (making the DNN mapping smoother there) and towards the decision boundary (making the DNN mapping less smooth there). Grokking occurs post phase transition as a robust partition of the input space thanks to the linearization of the DNN mapping around the training points. Web: \href{https://bit.ly/grok-adversarial}{bit.ly/grok-adversarial}.
\end{abstract}

\begin{figure}[!thb]
\begin{minipage}{0.03\linewidth}
    \centering
    \rotatebox[]{90}{{Local Complexity \hspace{4em} Accuracy}}
\end{minipage}%
\begin{minipage}{.97\linewidth}
    \centering
    \includegraphics[width=\linewidth]{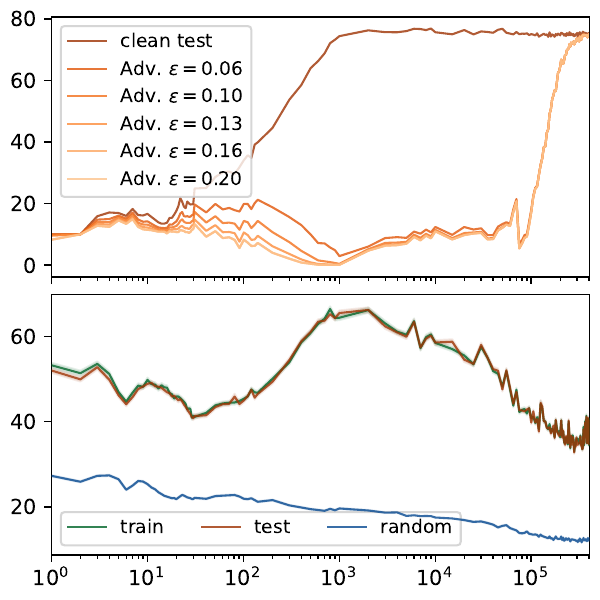}
\end{minipage}\\
\begin{minipage}{\linewidth}
    \centering
    Optimization Steps
\end{minipage}
    \centering
    \caption{\textbf{Deep Neural Networks grok robustness.} When training a ResNet18 on CIFAR10, without any controlled initialization as in \citet{liu2022omnigrok}, the network starts grokking adversarial examples generated using Projected Gradient Descent \cite{madry2017towards} after $10^4$ optimization steps (top) and attains almost equal robustness and generalization performance after $2\times10^5$ steps.  We see that, prior to grokking, the network undergoes a phase change during training in the \textit{local complexity}, i.e., the local density of spline partition regions
    in the input space (bottom). After test accuracy converges, the network starts \textit{migrating} its non-linearities 
    away from the data points and closer to the decision boundary (see \cref{fig:mnist-splinecam}), eventually reducing the complexity of the learned function around the data points. This increase and subsequent decrease in local non-linearity is a phenomenon visible for a wide variety of networks and training settings (see \cref{fig:resnet-cifar10-imagenette}). In this paper, we show that this particular training dynamic always results in delayed generalization or robustness.}
    \label{fig:resnet-cifar10-grok-robustness}
\end{figure}

\section{Introduction}

\begin{figure*}
\centering
    \begin{minipage}{.03\linewidth}
        \centering
    \rotatebox[]{90}{{\small{Local Complexity} \hspace{1em} \small{Accuracy}}}
    \end{minipage}%
    \begin{minipage}{0.37\linewidth}
        \centering
    \includegraphics[width=\linewidth]{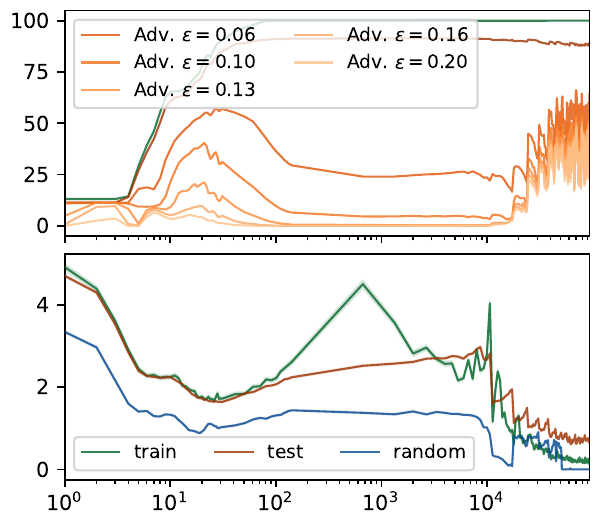}\\
    \begin{minipage}{\linewidth}
        \centering
        \small{Optimization Steps}
    \end{minipage}
    \end{minipage}%
    \begin{minipage}{0.6\linewidth}
        \centering
        \includegraphics[width=0.5\linewidth]{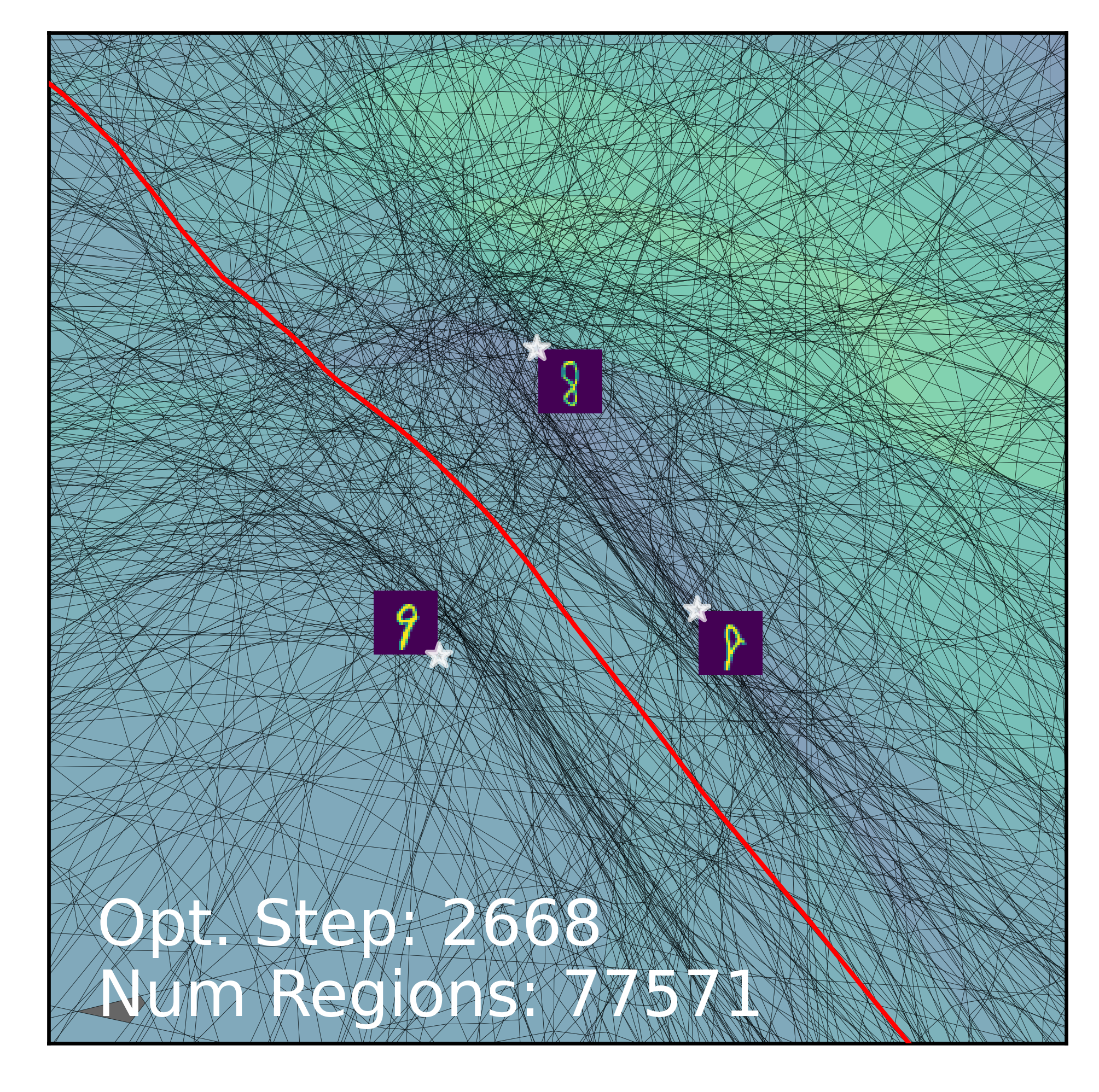}%
        \includegraphics[width=0.5\linewidth]{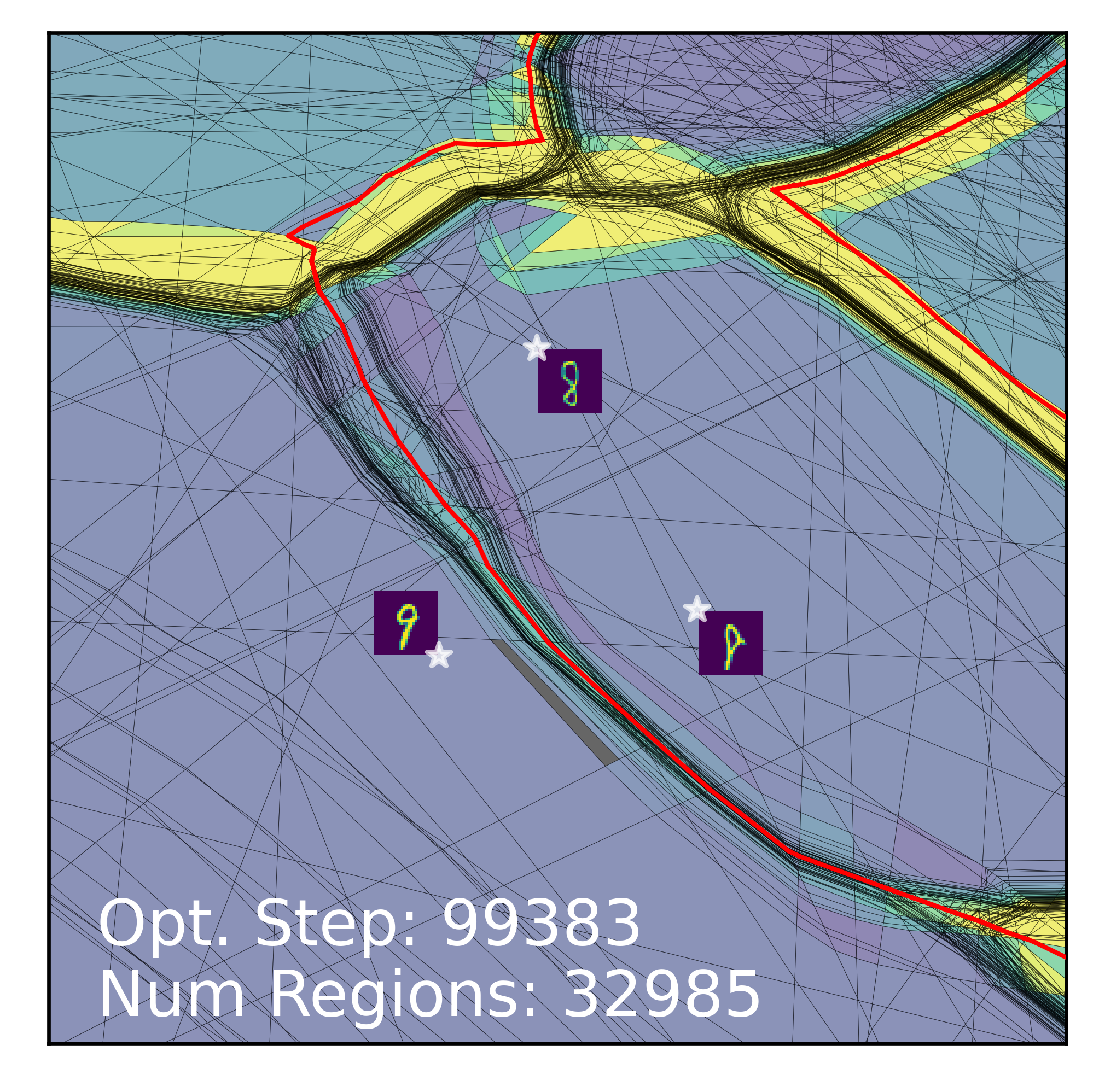}%
        \vspace{2.3em}
    \end{minipage}
    \caption{\textbf{Emergence of Robust Partition.} We train a 4-layer ReLU Multi Layer Perceptron (MLP) of $200$ width, on $1K$ samples from MNIST for $10^5$ optimization steps, with batch size $200$.
    We see that the network starts grokking adversarial examples after approximately $10^4$ optimization steps (top-left). The local complexity around data points (bottom-left) follows a double descent curve with the final descent starting approximately after $10^4$ optimization steps as well. {\em Where do the non-linearities migrate to?} In the {\bf middle} and {\bf right} images we present analytically computed visualizations of the DNN input space partition \cite{Humayun_2023_CVPR}. The partition or \textit{linear regions} are visualized across a 2D domain in the input space, that intersects three training samples. We see that during the final descent in local complexity, a unique structure emerges in the DNN partition geometry, where a large number of non-linearities (black lines) therefore linear regions, have concentrated around the decision boundary (red line). We dub this phenomenon \textit{Region Migration}. Animation for an entire training run in \href{https://bit.ly/grok-splinecam}{bit.ly/grok-splinecam}.
    }
    \label{fig:mnist-splinecam}
\end{figure*}

Grokking is a surprising phenomenon related to representation learning in Deep Neural Networks (DNNs) whereby DNNs may learn generalizing solutions to a task long after interpolating the training dataset, i.e., reaching near zero training error. It was first demonstrated by \cite{power2022grokking} on simple Transformer architectures performing modular addition or division. Subsequently, multiple studies have reported instances of grokking for settings outside of modular addition, e.g., DNNs initialized with large weight norms for MNIST, IMDb \cite{liu2022omnigrok}, or XOR cluster data \cite{xu2023benign}. For all the reported instances, DNNs that grok show a standard behavior in the training loss/accuracy curves approaching zero error as training progresses. The test error however, remains high even long after training error reaches zero. After a large number of training iterations, the DNN starts grokking--or generalizing--to the test data. This paper concerns the following question: 

\textbf{Question.}\textit{ How subjective is the onset of grokking on the test data? When grokking does not manifest as a measurable change in the test set performance, could there exist an alternate set of samples for which grokking would occur?}

To find an answer to the question, we look past the test dataset towards progressively generated adversarial samples, i.e., we generate adversarial samples after each training update by using PGD \cite{madry2017towards} attacks on the test data and monitor accuracy on adversarial samples. Note that it is not guaranteed that robustness towards adversarial samples would emerge with generalization, quite the contrary has been demonstrated in previous papers. For example, \citet{tsipras2018robustness} introduced the generalization-robustness trade-off, \citet{ilyas2019adversarial} demonstrated that robust networks learn fundamentally different representations. On the other hand, \citet{li2022robust} introduced the notion of 'robust generalization' and provided theoretical proof of its existence under linear separability conditions, indicating that robustness may be achieved alongside generalization. We report the following observation:

\textbf{Observation.} \textit{For a number of training settings, with standard initialization with or without weight decay, DNNs grok adversarial samples long after generalizing on the test dataset. 
We dub this novel, previously unreported form of grokking \textbf{delayed robustness}.
}

\begin{figure}[!thb]
\begin{minipage}{.04\linewidth}
    \centering
    \rotatebox[]{90}{{\small{Accuracy}}}
\end{minipage}%
\begin{minipage}{.92\linewidth}
    \centering
    \includegraphics[width=.95\linewidth]{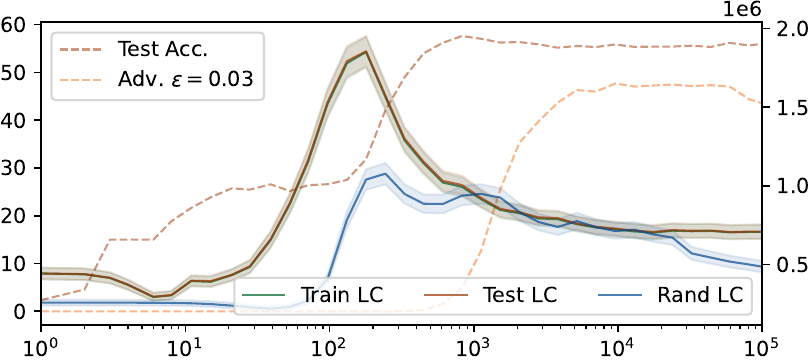}
\end{minipage}%
\begin{minipage}{.04\linewidth}
    \centering
    \rotatebox[]{270}{{\small{Local Complexity}}}
\end{minipage}\\
\begin{minipage}{\linewidth}
    \centering
    \small{Optimization Steps}
\end{minipage}
    \caption{\textbf{Delayed robustness in LLMs.} Grokking observed in a GPT architecture with 12 heads and 12 layers trained on next character prediction using the Shakespeare Text Dataset. We see that the second local complexity descent starts prior to the test acc. peak, and descent continues while the network groks $\eps=0.03$ $\ell_\infty$-PGD adversarial examples, generated in the token embedding space. Approximate input space partition visualized in \cref{fig:shakespeare_ascentpeak} and \cref{fig:shakespeare_aftergrok}.}
    \label{fig:grokking-llm}
\end{figure}

We make this observation for a number of training settings including for fully connected networks trained on MNIST (\cref{fig:mnist-splinecam}), Convolutional Neural Networks (CNNs) trained on CIFAR10 and CIFAR100 (\cref{fig:resnet-cifar10-imagenette}), ResNet18 without batch-normalization, trained on CIFAR10 (\cref{fig:resnet-cifar10-grok-robustness}) and Imagenette (\cref{fig:resnet-cifar10-imagenette}), and a GPT-based Architecture trained on Shakespeare Text (\cref{fig:grokking-llm}). We generate adversarial examples after each training step using $\ell_{\infty}$-PGD with varying $\epsilon \in \{0.03,0.06,0.10,0.13,0.16,0.20\}$, $\alpha = 0.0156$ and $10$ ($100$ for MNIST) PGD steps. This observation answers our initial question: indeed there can exist a dataset other than the test dataset for which grokking manifests in classification accuracy. Moreover, we observe that the same phenomenon occurs when test set grokking is induced via initialization scaling (\cref{fig:grokking-splinecam}) or when training transformers on Modular Addition (\cref{fig:grokking-transformer}).

\textbf{Question.} \textit{How can we explain both delayed generalization and delayed robustness?}    

\begin{figure}[!tbh]
\begin{minipage}{.4\linewidth}
    \centering
    \includegraphics[width=\linewidth]{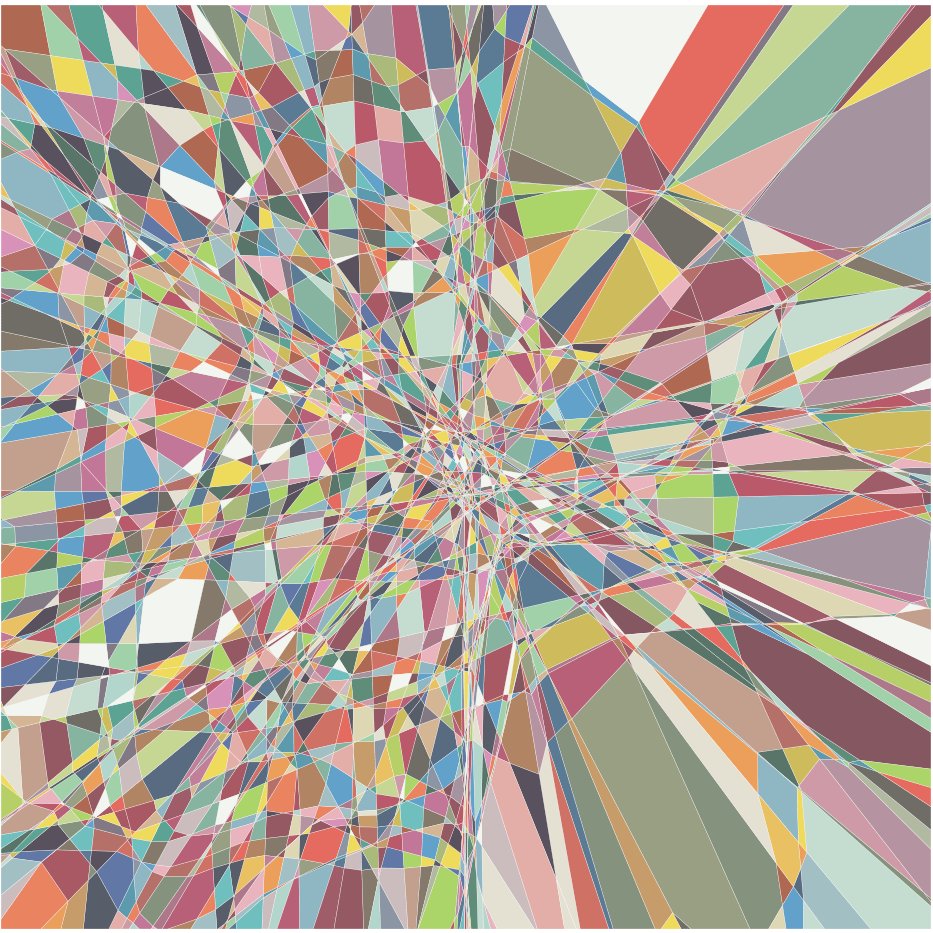}
\end{minipage}%
\begin{minipage}{.6\linewidth}
    \centering
    \includegraphics[width=\linewidth]{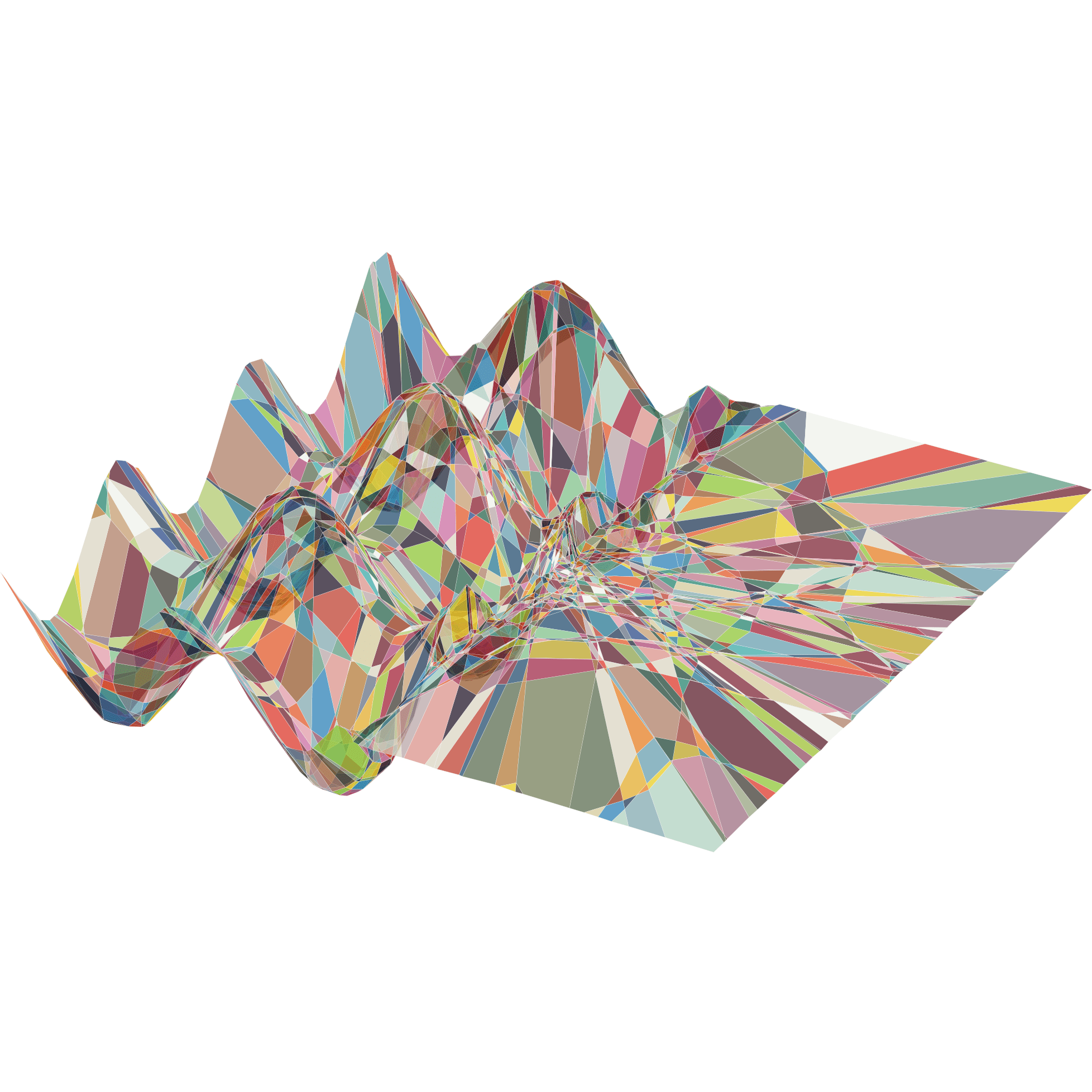}
\end{minipage}
    \caption{\textbf{Curvature and complexity.} Visual depiction of \cref{eq:CPA} with a toy affine spline $S : \mathbb{R}^2 \rightarrow \mathbb{R}$, obtained by training an MLP to regress the piecewise function $f(x_1,x_2) = \{\sin(x_1)+\cos(x_2)\}\Indic_{x_1<0}$. Regions in the input space partition $\Omega$ (left) and the graph of the affine spline function (right) are randomly colored. The spline partition has significantly higher density of non-linearities for $x_1<0$, i.e., the local complexity is higher where the learned function has more curvature.}
  \label{fig:2dexact}
\end{figure}

It has previously been established that both robustness and generalization are a function of the expressivity \cite{xu2012robustness,li2022robust} as well as the local linearity \cite{qin2019adversarial,balestriero2023police,humayun2023provable} of a DNN. To explain grokking, we propose a novel complexity measure based on the local non-linearity of the DNN. Our novel measure does not rely on the dataset, labels, or loss function that is used during training. It behaves as a progress measure \cite{barak2022hidden,nanda2023progress} that exhibits dynamics correlating with the onset of \textit{both delayed generalization and robustness}, opening new avenues to study grokking and DNN training dynamics. We show that DNNs undergo a phase change in the local complexity (LC) averaged over data points. Based on these dynamics, we come to the following conclusion:

\textbf{Claim.} \textit{Grokking occurs due to the emergence of a robust input space partition by a DNN, through a linearization of the DNN function around training points as a consequence of the training dynamics. This leads to larger linear regions around training points, and accumulation of non-linearities/linear regions around the decision boundary.}

We summarize our contributions as follows:

\begin{itemize}
    \item We observe for the first time \textbf{delayed robustness}, a novel form of grokking for DNNs that occurs for a wide range of training settings and co-occurs with delayed generalization. 
    
    \item  We develop a novel \textit{progress measure} 
    \cite{barak2022hidden} for DNN's based on the local complexity of a DNN's input space partition. Our proposed measure is a proxy for the DNN's expressivity, it is task agnostic yet informative of training dynamics. Using our measure, we detect three phases in training: two descent phases and an ascent phase. This is the first time that such dynamics in a DNN's partition are reported. We crucially observe that a DNN's partition regions concentrate around the decision boundary long after interpolation, a phenomenon we term \textit{region migration}.
    
    \item We pinpoint the origin of grokking via the spline viewpoint of DNNs \cite{balestriero2018spline}, connect it with the circuits viewpoint \cite{olah2020zoom}, and show that grokking always occurs during region migration.
    \item 
    Through a number of ablation studies we connect the training phases with DNN design parameters and study their changes during memorization/generalization.
\end{itemize}

We organize the rest of the paper as follows. In Section. 2 we overview the spline interpretation of deep networks and introduce our proposed local complexity measure. We also draw contrasts with common interpretability frameworks, e.g., the commonly used notion of circuits \cite{olah2020zoom} in mechanistic interpretability.
In Section. 3 we introduce the double descent characteristics of local complexity and connect region migration, i.e., the final phase of the double descent LC dynamics with grokking. We also present results showing that grokking does not happen when using batch normalization and provide theoretical justification. We present results connecting grokking with parameterization and memorization. Finally we draw conclusions from our results and discuss the limitations of our analysis.

\section{Local Complexity: A New Progress Measure}
\citet{barak2022hidden} introduced the notion of \textit{progress measures} for DNN training, as scalar quantities that are causally linked with the training state of a network.
The spline framework enables us to introduce our proposed progress measure, the local complexity of a DNN's partition. In later sections we show that local complexity dynamics are directly linked to grokking and present results showing its dependence on training and architectural parameters.

\begin{figure*}[!thb]
    \centering
    \includegraphics[width=\textwidth]{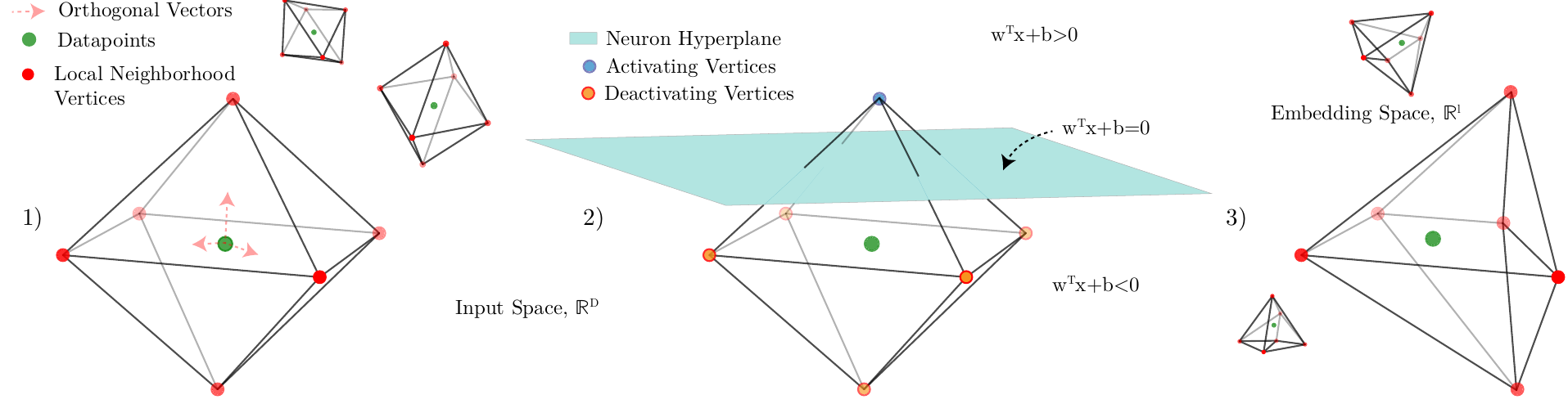}
    \caption{\textbf{Local Complexity Approximation.} 1) Given a point in the input space $x\in \mathbb{R}^D$, we start by sampling $P$ orthonormal vectors $\{v_1,v_2,...,v_P\}$ to obtain cross-polytopal frame $\mV_x=\{x \pm r*v_p \forall p\}$ centered on $x$, where $r$ is a radius parameter. We consider the convex hull $conv(\mV_x)$ as the local neighborhood of $x$. 2) If any neuron hyperplane intersects the neighborhood $conv(\mV_x)$ then the pre-activation sign will be different for the different vertices. We can therefore count the number neurons for a given layer, which results in sign changes in the pre-activation of $\mV_x$ to quantify local complexity $x$ for that layer. 3) By embedding $\mV_x$ to the input of the next layer, we can obtain a coarse approximation of the local neighborhood of $x$ and continue computing local complexity in a layerwise fashion.}
    \label{fig:complexity-schematic}
\end{figure*}%

\subsection{Deep Networks are Affine Spline Operators}
\label{sec:spline-intro}
DNNs primarily perform a sequential mapping of an input vector $\vx$ through $L$ nonlinear transformations, i.e., layers, as in
\begin{multline}
    f_{\theta}(\vx)\triangleq \mW^{(L)} \dots \va\left(\mW^{(2)}\va\left(\mW^{(1)}\vx+\vb^{(1)}\right)+\vb^{(2)}\right) \\
     \dots + \vb^{(L)},
    \label{eq:no_BN}    
\end{multline}
starting with some input $\vx$. For any layer $\ell \in \{1,\dots,L\}$, the $\mW^{(\ell)}$ weight matrix, and the $\vb^{(\ell)}$ bias vector
can be parameterized to control the type of operation for that layer, e.g., a circulant matrix as $\mW^{(\ell)}$ results in a convolutional layer. The operator $\ba$ is an element-wise nonlinearity, e.g., ReLU, and $\theta$ is the set of all parameters of the network. 
According to \citet{balestriero2018spline}, for any $\ba$ that is a continuous piecewise linear function, $f_{\theta}$ is a continuous piecewise affine spline operator. That is,  
there exists a partition $\Omega$ of the DNN's input space $\mathbb{R}^D$ (for example, \cref{fig:2dexact} left) comprised of non-overlapping regions that span the entire input space. On any region of the partition $\omega \in \Omega$, the DNN's input-output mapping is a simple affine mapping with parameters $(\mA_{\omega},\vb_{\omega})$. In short, we can express $f_{\theta}$ as 
\begin{equation}
    f_{\theta}(\vx) = \sum_{\omega \in \Omega}(\mA_{\omega}\vx+\vb_{\omega})\Indic_{\{\vx \in \omega\}}\label{eq:CPA},
\end{equation}
where, $\Indic_{\{\vx \in \omega\}}$ is an indicator function that is non-zero for $x \in \omega$.

\paragraph{Curvature and Linear Regions.} Formulations like that in \cref{eq:CPA} that represent DNNs as continuous piecewise affine splines, have previously been employed to make theoretical studies amenable to actual DNNs, e.g. in generative modeling \cite{humayun2022polarity}, network pruning \cite{you2021max}, and OOD detection \cite{ji2022test}. Empirical estimates of the density of linear regions in the spline partition have also been employed  in sensitivity analysis \cite{novak2018sensitivity}, quantifying non-linearity \cite{gamba2022all}, quantifying expressivity \cite{raghu2017expressive} or to estimate the complexity of spline functions \cite{hanin2019complexity}.
We demonstrate the relationship between function curvature and linear region density through a toy example in \cref{fig:2dexact}. In \cref{fig:2dexact}-left and \cref{fig:mnist-splinecam}-(middle,right), any contiguous line is a non-linearity in the input space, corresponding to a single neuron of the network. All the non-linearities re-orient themselves during training to be able to obtain the target function (\cref{fig:2dexact}-right). Therefore, in \cref{fig:2dexact}, we see that DNN partitions have higher density of linear regions/non-linearities/knots in the spline partition, where the target function curvature is non-zero. 

\subsection{Measuring Local Complexity using the Deep Network Spline Partition}
\begin{figure}[!thb]
    \centering
    \includegraphics[width=\linewidth]{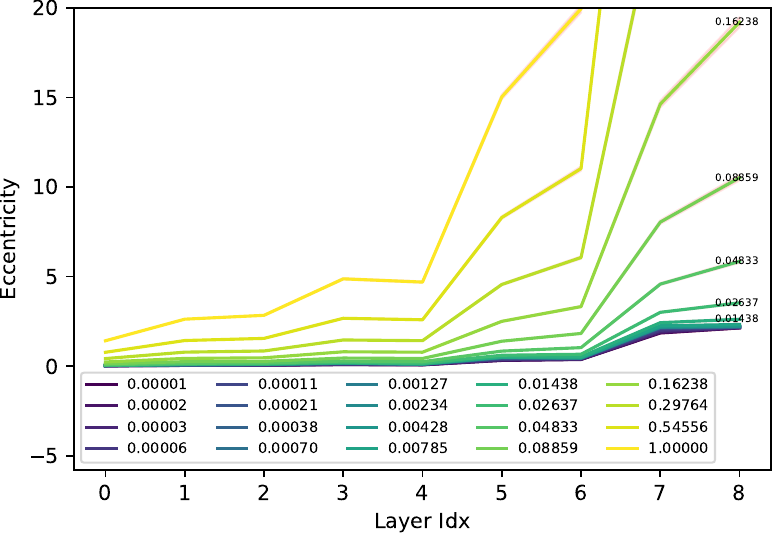}%
    \caption{\textbf{Deformation with depth.} Change of average eccentricity \cite{xu2021comparing} of the input space neighborhoods $\mV_x$ by different layers of a CNN trained on the CIFAR10 dataset, for different radius $r$.
    We see that, for larger radius, the deformation increases with depth almost exponentially. For $r\leq 0.014$ deformation is low, indicating that smaller radius neighborhoods are reliable for LC computation on deeper networks. Values are averaged over neighborhoods sampled for $1000$ training points from CIFAR10. For ResNet18, see \cref{fig:appendix_deformation_resnet}.}
    \label{fig:main_deformation_cnn}
\end{figure}
\begin{figure*}[!thb]
\begin{minipage}{0.03\textwidth}
    \centering
    \rotatebox[]{90}{\small{Local Complexity \hspace{2.5em} Accuracy}}
\end{minipage}%
\begin{minipage}{0.33\textwidth}
    \centering
    \includegraphics[width=\textwidth]{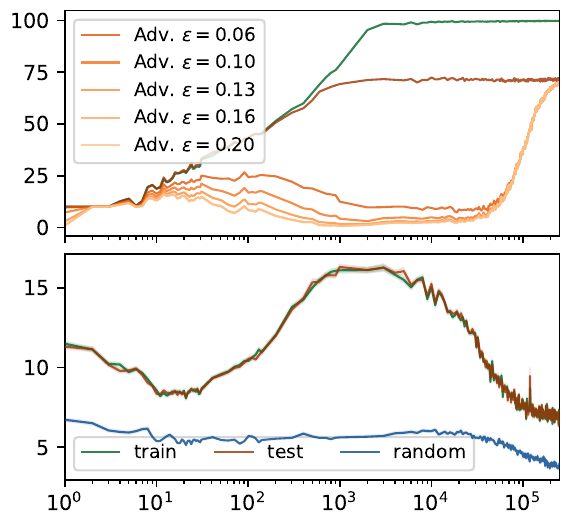}
\end{minipage}%
\begin{minipage}{0.33\textwidth}
    \centering
    \includegraphics[width=\textwidth]{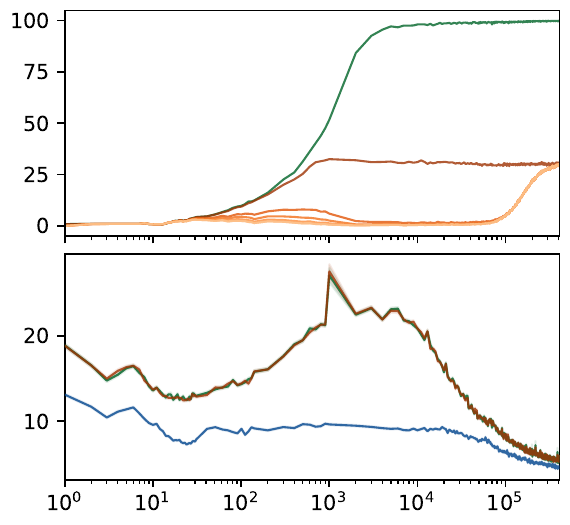}
\end{minipage}%
\begin{minipage}{0.33\textwidth}
    \centering
    \includegraphics[width=\textwidth]{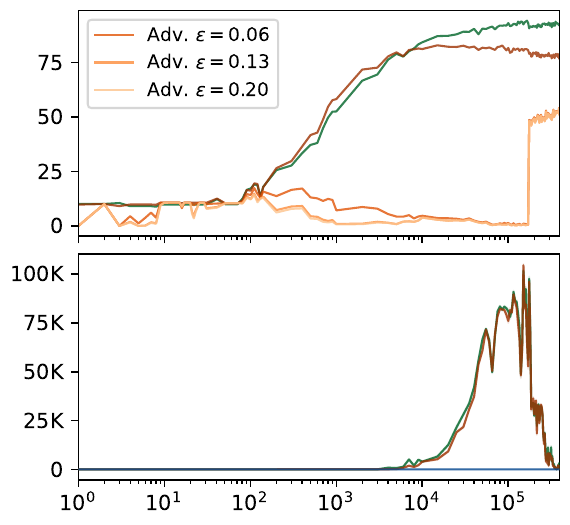}
\end{minipage}\\
\begin{minipage}{\textwidth}
    \centering
    \vspace{1em}
    \small{Optimization Steps}
\end{minipage}%
    \caption{\textbf{Grokking across datasets and architectures.} From left to right, examples of delayed robustness emerging late in training for a CNN trained on CIFAR10, CNN trained on CIFAR100, and ResNet18 trained on the Imagenette\footnote{\href{https://github.com/fastai/imagenette}{github.com/fastai/imagenette}} dataset. Clear double descent behavior visible in the local complexity of CNN with CIFAR10 and CIFAR100. The ResNet18 trained with Imagenette obtains a very high local complexity during the ascent phase of training. To compute local complexity we consider $25$ dimensional neighborhoods centered on $1024$ train, test or random samples. We use $r=0.005$ for CNN and $r=10^{-4}$ for ResNet18.}
    \label{fig:resnet-cifar10-imagenette}
\end{figure*}%

Suppose a domain is specified as the convex hull of a set of vertices $\bm{V} = \left[ \vv_1, \ldots \vv_p \right]^T$ in the DNN's input space. We wish to compute the local complexity or smoothness \cite{hanin2019complexity} for neighborhood $\mathcal{V} = conv(\bV)$. Consider a single hidden layer of a network. Let's denote the DNN layer weight as $W^{(\ell)}\triangleq [\vw^{(\ell)}_1,\dots, \vw^{(\ell)}_{D^{(\ell)}}]$, $b^{(\ell)}$ where $\ell$ is the layer index, $\vw^{(\ell)}_i$ is the $i$-th row of $W^{(\ell)}$ or weight of the $i$-th neuron, and $D^{(\ell)}$ is the output space dimension of layer $\ell$. The forward pass through this layer for $\mV$ can be considered an inner product with each row of the weight matrix $W^{(\ell)}$ followed by a continuous piecewise linear activation function. Without loss of generality,
let's consider ReLU as the activation function in our network. The partition at the input space of layer $\ell$ can therefore be expressed as the set of all hyperplane equations formed via the neuron weights such as: 
\begin{align}
    \partial \Omega  &= \bigcup_{i=1}^{D^{(\ell)}} \mathcal{H}^{(\ell)}_i \\
    \mathcal{H}^{(\ell)}_i &=
    \left\{\vx \in \mathbb{R}^{D^{(\ell-1)}} : \langle \vw_i^{(\ell)} , \vx \rangle + \vb_i^{(\ell)} = 0 \right\}, \label{eq:hyperplane}
\end{align}
which is also the set of layer $\ell$ non-linearities. Let, $\Phi = f_{1:\ell-1}(\mathcal{V})$ be the embedded representation of the neighborhood $\mathcal{V}$ by layer $\ell-1$ of the network. Therefore, approximating the local complexity of $\mathcal{V}$ induced by layer $\ell$, would be equivalent to counting the number of linear regions in,
\begin{equation}
\Phi \cap \partial \Omega  = \bigcup_{i=1}^{D^{(\ell)}} \Phi \cap \mathcal{H}^{(\ell)}_i.
\end{equation}
The local partition inside $\Phi$ results from an arrangement of hyperplanes; therefore the number of regions is of the order $\mathcal{N}^{D^{(\ell-1)}}$ \cite{toth2017handbook}, where
\begin{equation}
\mathcal{N}=|\{ i : i=1,2..D^{(\ell)} \text{   and   } \mathcal{H}_i^{(\ell)} \cap \Phi \neq \emptyset \} |,
\end{equation}
is the number of hyperplanes from layer $\ell$ intersecting $\Phi$. We consider $\mathcal{N}$ as a proxy for the local complexity of any neighborhood $\Phi$. 
To make computation tractable, let,
$\Phi \approx \widehat{\Phi} = conv(f_{1:\ell-1}(\mV))$. 
Therefore, for $\widehat{\Phi}$, any sign changes in layer $\ell$ pre-activations is due to the corresponding neuron hyperplanes intersecting $conv(\mV)$. 
For a single layer, the local complexity (LC) for a sample in the input space can be approximated by the number of neuron hyperplanes that intersect $\bV$ embedded to that layers input space. 
If we consider input space neighborhoods with the same volume, then our approximation method measures the un-normalized density of non-linearity in an input space locality, which we consider a proxy for local complexity. We highlight that this is tied to the VC-dimension of (ReLU) DNN \cite{bartlett2019nearly} where the more regions are present the more expressive the decision boundary can be \cite{montufar2014number}. 
In \cref{fig:complexity-schematic}, we provide a visual explanation of our method for local complexity approximation through a cartoon schematic diagram. To summarize, we consider randomly oriented $P$ dimensional $\ell_1$ norm balls with radius $r$, i.e., cross-polytopes centered on any given data point $x$ as a frame defining the neighborhood.
We therefore follow the steps entailed in \cref{fig:complexity-schematic} in a layerwise fashion, to approximate the local complexity in the prescribed neighborhood for a given layer.   

\begin{figure*}[!htb]
    \begin{minipage}{0.04\textwidth}
        \centering
        \rotatebox[]{90}{{\small{Accuracy}}}
    \end{minipage}%
    \begin{minipage}{.28\textwidth}
    \begin{minipage}{\linewidth}
    \centering
    \includegraphics[width=\linewidth]{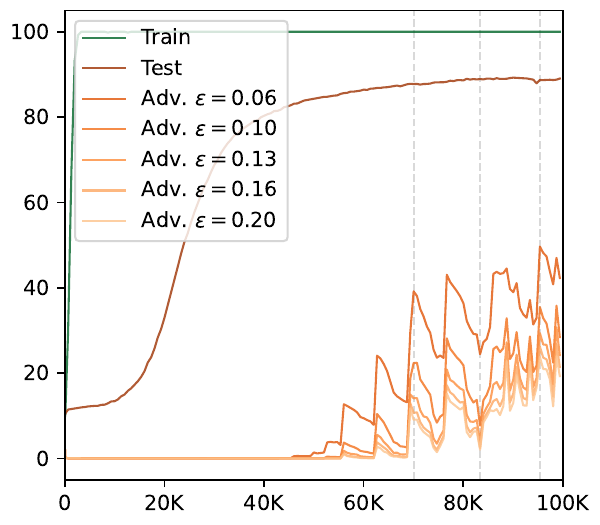}    
    \end{minipage}\\
    \begin{minipage}{\linewidth}
        \centering
        \small{Optimization Steps}
    \end{minipage}%
    \end{minipage}%
    \begin{minipage}{.7\textwidth}
    \begin{minipage}{\linewidth}    
    \centering
    \includegraphics[width=0.33\linewidth]{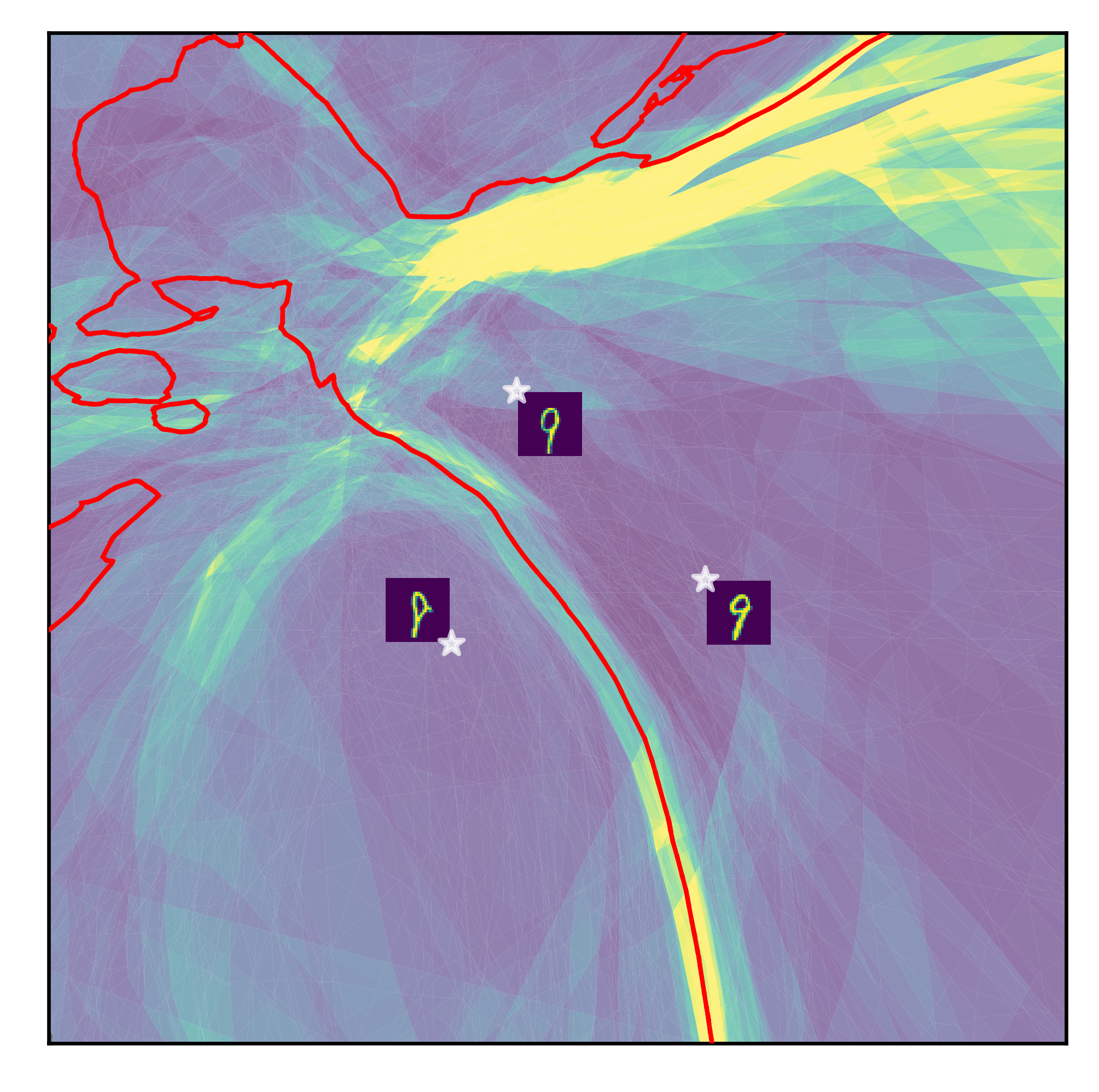}%
    \includegraphics[width=0.33\linewidth]{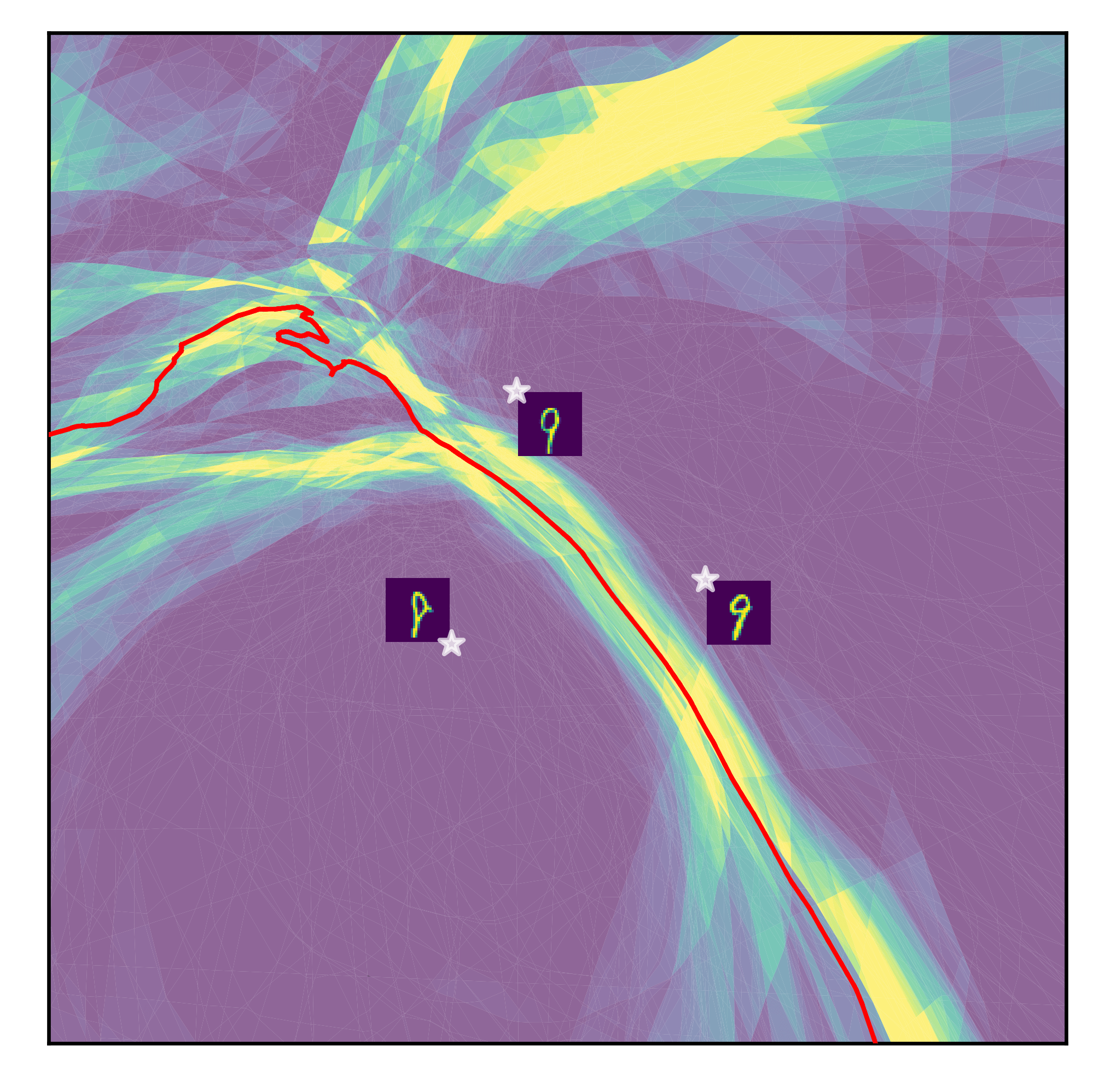}%
    \includegraphics[width=0.33\linewidth]{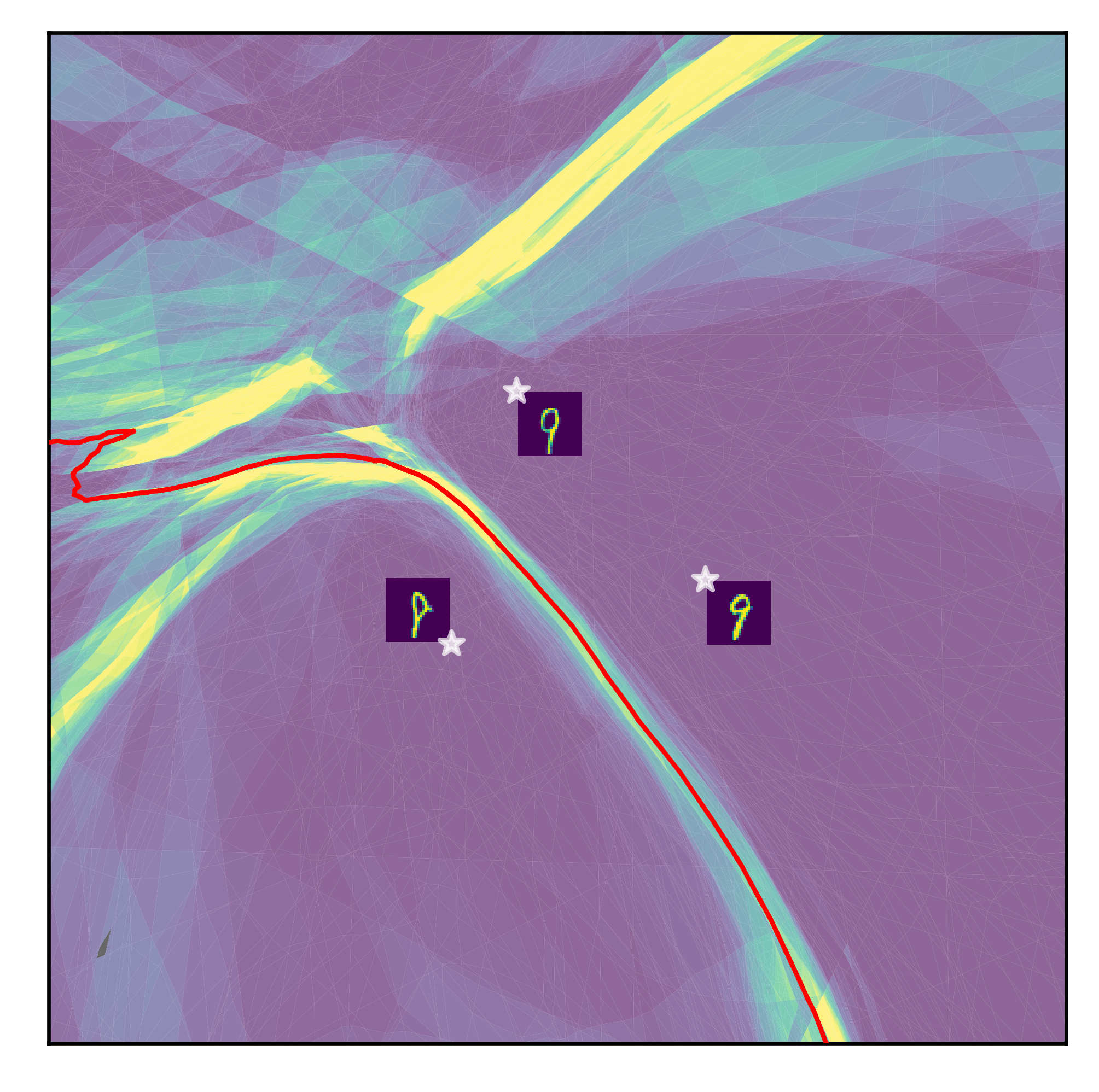}%
    \end{minipage}\\
    \begin{minipage}{\linewidth}
        \centering
        \small{Opt. Step 77035} \hspace{5em} \small{Opt. Step 83375} \hspace{5em} \small{Opt. Step 95381}
    \end{minipage}
    \end{minipage}
    \centering
    \caption{\textbf{Grokking visualized.} We induce grokking by randomly initializing a $4$ depth $200$ width ReLU MLP and scaling the initialized parameters by eight following \cite{liu2022omnigrok}. In the leftmost figure, we can see that the grokking is visible for both the test samples as well as adversarial examples generated using the test set. We see that the network robustness, periodically increases. By visualization the partition and curvature of the function across a 2D slice of the input space \cite{Humayun_2023_CVPR}, we see that the network periodically increases the concentration of non-linearity around its decision boundary, making the boundary sharper at each robustness peak. This occurs even when the network doesn't undergo delayed generalization (\cref{fig:mnist-splinecam}). As the local complexity around the decision boundary increases, the local complexity around data points farther from the decision boundary decreases (\cref{fig:splinecam-grok-inandoutclass}).}
    \label{fig:grokking-splinecam}
\end{figure*}

\begin{figure}[!thb]
\begin{minipage}{.04\linewidth}
    \centering
    \rotatebox[]{90}{{\small{Accuracy}}}
\end{minipage}%
\begin{minipage}{.92\linewidth}
    \centering
    \includegraphics[width=\linewidth]{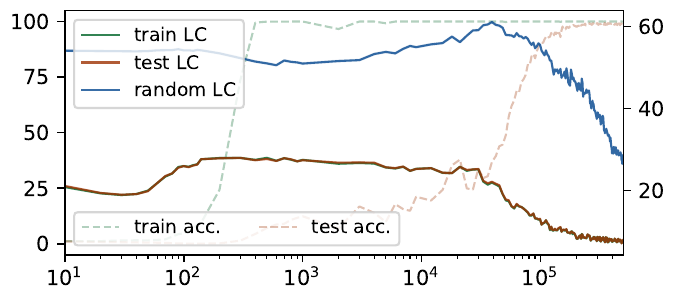}
\end{minipage}%
\begin{minipage}{.04\linewidth}
    \centering
    \rotatebox[]{270}{{\small{Local Complexity}}}
\end{minipage}\\
\begin{minipage}{\linewidth}
    \centering
    \small{Optimization Steps}
\end{minipage}
    \caption{\textbf{Region migration in modular addition.} By measuring the local complexity for the GeLU activated fully connected layers of a Transformer architecture, we see that here as well, region migration occurs during grokking.}
    \label{fig:grokking-transformer}
\end{figure}

\paragraph{Sensitivity of approximation to $P$ and $r$.}

One of the possible limitations of local complexity measure is the deformation of the local neighborhood when its passed through a network from layer to layer, as shown in \cref{fig:complexity-schematic}.
For different radius $r$ of the input space neighborhood $\mV_x$ centered on any arbitrary data point $x$, we compute the graph eccentricity \cite{xu2021comparing} of $\mV_x$ after being embedded by different layers of a CNN. 
We present the results in \cref{fig:main_deformation_cnn} for $1000$ different training data points for a CNN trained on CIFAR10.
The higher the change of eccentricity compared to the input space (index 0), the more likely the neighborhood gets deformed, leading to less reliable approximation.
We see that below a certain radius value, deformation by the CNN is limited and does not exponentially increase. In subsequent experiments however, e.g., \cref{fig:mlp-depth-sweep}, we have observed that the dynamics of local complexity is similar between large and small $r$ neighborhoods.
We present more validation experiments in 
\cref{sec:evaluating-complexity-approximation}.
Our proposed method can also be used to approximate the input space partition formed by a neural network. 
In \cref{fig:2D_lc_partition} we compare the partition approximated via LC computations on a grid, with analytically computed partition via \cite{humayun2023splinecam}. In \cref{fig:shakespeare_ascentpeak} and \cref{fig:shakespeare_aftergrok} we present the input space partition approximated for a GPT architecture
before and after delayed robustness occurs.

\paragraph{Experimental Setup.}
For all experiments we sample $1024$ train test and random points for local complexity (LC) computation, except for the MNIST experiments, where we use $1000$ training points (all of the training set where applicable) and $10000$ test and random points for LC computation. We use $r=0.005$ and $P=25$ unless specified otherwise and except for the ResNet18 experiments with Imagenette where we use $r=10^{-4}$. For training, we use the Adam optimizer and a weight decay of 0 for all the experiments except for the MNIST-MLP experiments where we use a weight decay of $0.01$. Unless specified, we use CNNs with 5 convolutional layers and two linear layers. For the ResNet18 experiments with CIFAR10, we use a pre-activation architecture with width $16$. For the Imagenette experiments, we use the standard torchvision Resnet architecture. For all settings we do not use Batch Normalizaiton, as reasoned in \cref{sec:batchnorm-theorem}.
In all our plots, we denote training accuracy/LC using green, test accuracy/LC using orange and random LC using blue colors. We also color curves for adversarial examples using different shades of orange. All local complexity plots show the $99\%$ confidence interval.

\begin{figure*}[!htb]
    \centering
    \begin{minipage}{.03\linewidth}
    \centering
    \rotatebox[]{90}{{\small{Accuracy}}}
    \end{minipage}%
    \begin{minipage}{0.3\linewidth}
    \centering
    \includegraphics[width=\linewidth]{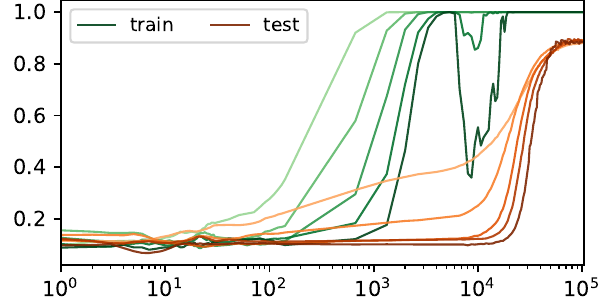}
    \end{minipage}%
    \begin{minipage}{.05\linewidth}
    \centering
    \rotatebox[]{90}{{\small{Local Complexity}}}
    \end{minipage}%
    \begin{minipage}{.6\linewidth}
    \includegraphics[width=0.5\linewidth]{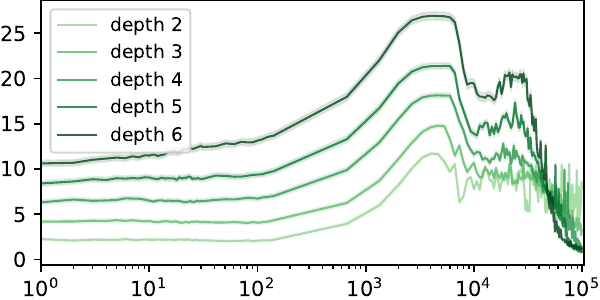}%
    \includegraphics[width=0.5\linewidth]{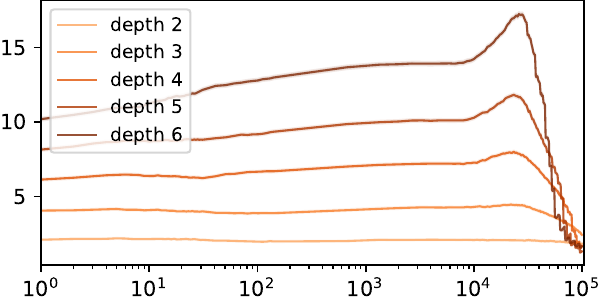}%
    \end{minipage}\\
    \begin{minipage}{\linewidth}
        \centering
        \small{Optimization Steps}
    \end{minipage}
    
    \caption{\textbf{Local complexity across depths.} From left to right, accuracy, local complexity around training and local complexity around test data points, for an MLP trained on MNIST with width $200$ and varying depth. \textit{As depth is increased the max LC during ascent phase becomes larger. We can also see a distinct second peak right before the descent phase.}
    }
    \label{fig:grok_dswp}
\end{figure*}

\section{Local Complexity Training Dynamics and Grokking}
\label{para:splinecam-usage}

\subsection{Emergence of a Robust Partition} We start our exploration of the training dynamics of deep neural networks by formalizing the phases of local complexity observed during training. In all our experiments either involving delayed generalization or robustness, we see three distinct phases in the dynamics of local complexity: 

$\bullet$ \textit{The first descent}, when  the local complexity start by descending after initialization. This phase is subject to the network parameterization as well as initialization, e.g., when grokking is induced in the MLP-MNIST case with scaled initialization, we do not see the first descent (\cref{fig:grokking-init8-mlp-depth-sweep}, \cref{fig:grok_dswp}).

$\bullet$ \textit{The ascent phase}, when the local complexity accumulates around both training and test data points. The ascent phase happens ubiquitously, and the local complexity generally keeps ascending until training interpolation is reached (e.g., \cref{fig:resnet-cifar10-imagenette}, \cref{fig:resnet-cifar10-grok-robustness}). During the ascent phase, the training local complexity may be higher for training data points than for test data points, indicating an accumulation of non-linearities around training data compared to test data \cref{fig:mnist-splinecam}.

\begin{figure}[!bht]
    \centering
    \includegraphics[width=.5\linewidth]{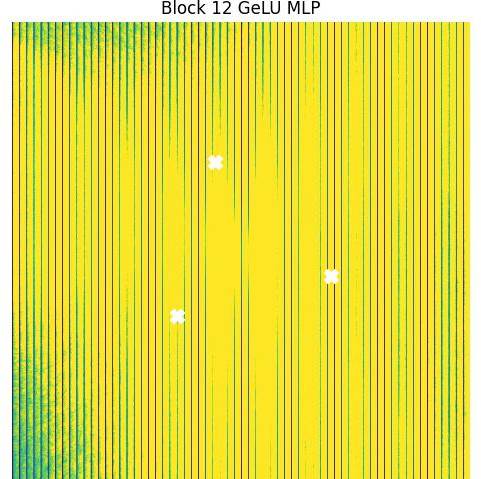}%
    \includegraphics[width=.5\linewidth]{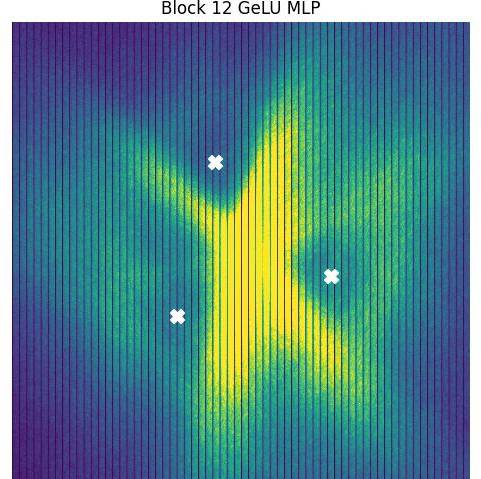}
    \caption{Token embedding space partition formed by the GeLU activated MLP of Block 12 of the GPT model mentioned in \cref{fig:grokking-llm}. The partition is approximated by computing LC on a $512\times512$ grid on a 2D subspace in the token embedding space. Note that after grokking (right), the three random samples used to determine the 2D subspace, has visibly lower local complexity in its immediate neighborhood.}
    \label{fig:layer12_llm_beforeafter}
\end{figure}

$\bullet$ \textit{The second descent phase} or region migration phase, during which the network moves the linear regions or non-linearities away from the training and test data points. Focusing on \cref{fig:mnist-splinecam}-bottom-left and \cref{fig:grokking-init8-mlp-depth-sweep} for the MLP-MNIST setting, one perplexing observation that we make is that the local complexity around random points -- uniformly sampled from the domain of the data -- also decreases during the final descent phase. This would mean that the non-linearities are not randomly moving away from the training data, but systematically reorganizing where we do not have our LC approximation probes. To better understand the phenomenon, we consider a square domain $\mathbb{D}$ that passes through three MNIST training points, and use Splinecam \cite{Humayun_2023_CVPR} to analytically compute the input space partition on $\mathbb{D}$. In short, Splinecam uses the weights of the network to exactly compute the input space representation of each neuron's zero-level set on $\mathbb{D}$ (black lines in \cref{fig:mnist-splinecam}). We present Splinecam visualizations for different optimization steps in \cref{fig:mnist-splinecam}, \cref{fig:grokking-splinecam}, and \cref{fig:splinecam-grok-inandoutclass}. Through these visualizations, we see clear evidence that \textit{during the second descent phases of training, linear regions or the non-linearities of the network, migrate close to the decision boundary creating a robust partition in the input space.} The robust partition contains large linear regions around the training data, as suggested by papers in literature as a precursor for robustness \cite{qin2019adversarial}. Moreover, during region migration, \textit{the network intends to lower the local complexity around training points}, resulting in a decrease in local complexity around training even compared to test data points.  

\begin{figure}[!thb]
\begin{minipage}{0.04\linewidth}
    \centering
    \rotatebox[]{90}{{\small{Local Complexity}}}
\end{minipage}%
\begin{minipage}{0.96\linewidth}
    \centering
    \includegraphics[width=\linewidth]{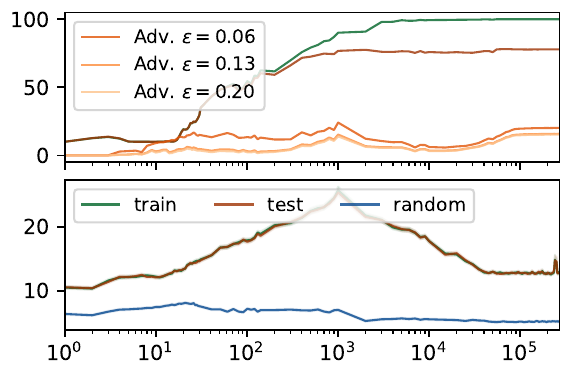}
\end{minipage}\\
\begin{minipage}{\linewidth}
    \centering
    \small{Optimization Steps}
\end{minipage}
    \caption{\textbf{Batch-norm removes grokking.} Training a CNN with an identical setting as in \cref{fig:resnet-cifar10-imagenette}-left, except the CNN now has Batch Normalization layers after every convolution. With the presence of batchnorm, the LC values increase, the initial descent gets removed and most importantly, grokking does not occur for adversarial samples.}
    \label{fig:cnn-cifar10-batchnorm}
\end{figure}

\paragraph{Local complexity as a progress measure.}
While we don't quite understand why the network goes from accumulation to repelling of non-linearities around the training data between the ascent and second descent phases, we see that the second descent always precedes the onset of delayed generalization or delayed robustness. In \cref{fig:grokking-splinecam}-middle and right, we present splinecam visualizations for a network during grokking. The colors denote the norm of the slope parameter $\mA_{\omega}$ for each region $\omega$ computed obtained via SplineCam. We see that while a network groks, the regions start concentrating around the decision boundary where the network has the highest norm. This is intuitive because in such classification settings, an increase of local complexity around the decision boundary allows the function to sharply transition from one class to another. Therefore, therefore the more the non-linearites converge towards the decision boundary, the higher the function norm can be while smoothly transitioning as well. We have provided an animation showing the evolution of partition geometry and emergence of the robust partition during training here\footnote{\href{https://bit.ly/grok-splinecam}{bit.ly/grok-splinecam}}. In the animation, we can see that the partition periodically switches between robust configurations during region migration. As time progresses we see increasing accumulation of the non-linearities around the decision boundary. These results undoubtedly show that the local non-linearity or local complexity dynamics is directly tied to the partition geometry and emergence of delayed generalization/robustness. 

\paragraph{Relationship with Circuits.}
A common theme in mechanistic interpretability, especially when it comes to explaining the grokking phenomenon, is the idea of 'circuit' formation during training \cite{nanda2023progress, varma2023explaining, olah2020zoom}. A circuit is loosely defined as a subgraph of a deep neural network containing neurons (or linear combination of neurons) as nodes, and weights of the network as edges. 
Recall that \cref{eq:CPA} expresses the operation of the network in a region-wise fashion, i.e., for all input vectors $\{x  : x\in \omega\}$, the network performs the same affine operation using parameters $(\mA_{\omega},\vb_{\omega})$ while mapping $x$ to the output. The affine parameters for any given region, are a function of the active neurons in the network as was shown by \citet{Humayun_2023_CVPR} (Lemma 1). Therefore for each region, we necessarily have a circuit or subgraph of the network performing the linear operation. Between two neighboring regions, only one node of the circuit changes. From this perspective, our local complexity measure can be interpreted as a way to measure the density of unique circuits formed in a locality of the input space as well. While in practice this would result in an exponential number of circuits, the emergence of a robust partition show that towards the end of training, the number of unique circuits get drastically reduced. This is especially true for sub-circuits corresponding to deeper layers only. In \cref{fig:layerwise partition}, we show the robust partition in a layerwise fashion. We can see that for deeper layers, there exists large regions, i.e., embedding regions with only one circuit operation through the layer. This result, matches with the intuition provided by \citet{nanda2023progress} on the cleanup phase of circuit formation late in training.



\begin{figure}
\begin{minipage}{0.05\linewidth}
\centering
    \rotatebox[]{90}{{\small{Local Complexity}}}
\end{minipage}%
\begin{minipage}{.95\linewidth}
    \centering
    \includegraphics[width=\linewidth]{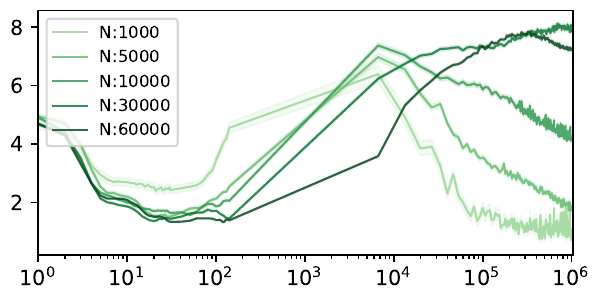}%
\end{minipage}\\
\begin{minipage}{\linewidth}
    \centering
    \hspace{0.05\linewidth}Optimization Steps
\end{minipage}
    \caption{\textbf{Memorization requirement delays grok.} When training an MLP on varying number of randomly labeled MNIST samples, we see that with increase in the number of samples, the local complexity dynamics get delayed, especially the ascent phase gets elongated. This shows that with increased demand for memorization the network takes longer to complete ascent and later undergo region migration.}
    \label{fig:mlp_rand_dataswp}
\end{figure}

\begin{table*}[!thb]
\centering
\caption{Summary of all the experiments showing the relationship between delayed generalization and training/model hyperparameters.}
\label{tab:grokking_summary}
\resizebox{\textwidth}{!}{%
\begin{tabular}{@{}lllll@{}}
\toprule
Training Intervention                       & Dataset          & Model    & Description                                    & Effect on Adv. Grokking              \\ \midrule
Increasing Batchsize                        & CIFAR10          & Resnet18 & Batchsize Increased from 64 to 512             & Expedited (\cref{fig:batch_sweep})                            \\
Increasing Parameters: Depth                & MNIST            & MLP      & Increasing Depth from 2 to 6                   & Expedited (\cref{fig:mlp-depth-sweep})                            \\
Increasing Parameters: Depth                & Shakespeare Text & GPT      & Increasing number of layers from 6 to 12       & Expedited                            \\
Increasing Parameters: Width                & MNIST            & MLP      & Increasing Width from 20 to 2000               & Expedited (\cref{fig:affect-width})                            \\
Increasing Parameters: Width                & CIFAR10          & Resnet18 & Increasing Width from 16 filters to 64 filters & Expedited                             \\
Increasing Regularization                   & MNIST            & MLP      & Weight Decay increased from 0. to 1            & Delayed  (\cref{fig:mlp-weight-decay})                            \\
Increasing Training Data                    & MNIST            & MLP      & Training data increased from 1K to 60K         & Expedited (\cref{fig:mlp-dataswp})                           \\
Increasing Training Data (Randomly Labeled) & MNIST            & MLP      & Training data increased from 1K to 60K         & Delayed (\cref{fig:mlp_rand_dataswp})                              \\
No BatchNorm -\textgreater BatchNorm        & CIFAR10          & Resnet18 & Adding BatchNorm Layer after convolution       & Does not occur/significantly delayed \\
No BatchNorm -\textgreater BatchNorm        & CIFAR10          & CNN      & Adding BatchNorm Layer after convolution       & Does not occur/significantly delayed \\ \bottomrule
\end{tabular}%
}
\end{table*}

\begin{figure}[!htb]
\begin{minipage}{0.04\linewidth}
    \centering
    \rotatebox[]{90}{{\small{Local Complexity}}}
\end{minipage}%
\begin{minipage}{0.96\linewidth}
    \centering
    \includegraphics[width=0.5\linewidth]{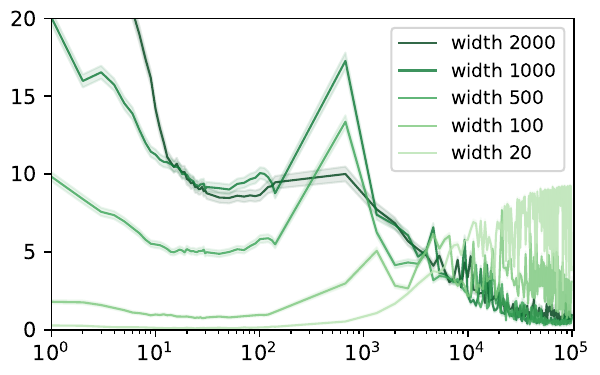}%
    \includegraphics[width=0.5\linewidth]{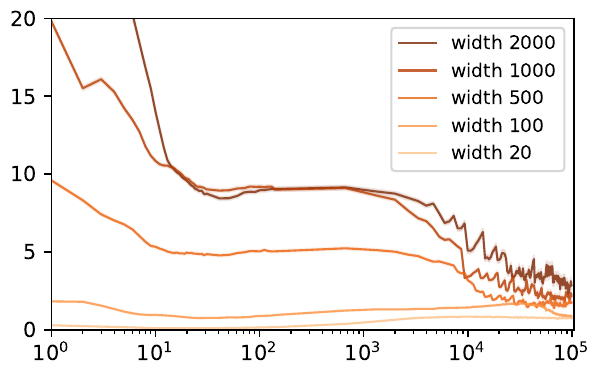}
\end{minipage}\\
\begin{minipage}{\linewidth}
    \centering
    \small{Optimization Steps}
\end{minipage}
    \caption{\textbf{Increasing width hastens region migration.} LC dynamics while training an MLP with \textbf{varying width} on MNIST. For the peak LC achieved around training points during the ascent phase, we see an initial increase and then decrease as the network gets overparameterized. For test and random samples, we see the LC during ascent phase saturating as we increase width.}
    \label{fig:affect-width}
\end{figure}

\section{What Affects the Progress Measure?}

\textbf{Parameterization.} In \cref{fig:affect-width,fig:mlp-depth-sweep,fig:grokking-init8-mlp-depth-sweep}, we see that increasing the number of parameters either by increasing depth, or by increasing width of the network in our MNIST-MLP experiments, hastens region migration, therefore makes grokking happen earlier.

\textbf{Weight Decay} regularizes a neural network by reducing the norm of the network weights, therefore reducing the per region slope norm as well. We train a CNN with depth 5 and width 32 on CIFAR10 with varying weight decay. In \cref{fig:mlp-weight-decay} we present the train, test and  random LC for our experiments for neighborhoods of different radius. Weight decay does not seem to have a monotonic behavior as it both delays and hastens region migration, based on the amount of weight decay.

\textbf{Batch Normalization.} Batch normalization removes grokking. In \cref{sec:batchnorm-theorem}, we show that at each layer $\ell$ of a DN, BN explicitly adapts the partition so that the partition boundaries are as close to the training data as possible. This is confirmed by our experiments in \cref{fig:cnn-cifar10-batchnorm} where we see that grokking adversarial examples ceases to occur compared to the non-batchnorm setting in \cref{fig:resnet-cifar10-imagenette}. BN also removes the first descent, monotonically increasing the local complexity around the data manifold and after a while undergoing a phase change and decreasing. The degree of region migration is reduced during this phase, as can be seen in the higher LC when we use batch normalization. While training a ResNet18 with Batch Norm on Imagenet Full (\cref{fig:batchnorm_imagenet}), we see that the local complexity keeps increasing indefinitely, removing any signs of region migration. 

\paragraph{Activation function} While most of our experiments use ReLU activated networks, in \cref{fig:gelu_mlp} we present results for a GeLU activated MLP, as well as in \cref{fig:grokking-transformer} we present results for a GeLU activated Transformer. For both settings we see similar training dynamics as is observed for ReLU.

\textbf{Effect of Training Data.} We control the training dataset to either induce higher generalization on higher memorization. Recall that in our MNIST experiments, we use $1k$ training samples. We increase the number of samples in our dataset to monitor the effect of grokking \cref{fig:advgrok-mnist-datasweep} and LC \cref{fig:mlp-dataswp}. We see that increasing the size of the dataset hastens grokking. On the other hand we also sweep the dataset size for a random label memorization task \cref{fig:mlp_rand_dataswp}, \cref{fig:mlp-randlbl-depth-sweep}. We see that in this case, increasing dataset size results in more memorization requirement, therefore it delays the region migration phase. 


\section{Conclusions and Limitations}

We have pursued a thorough empirical study of grokking, both on the test dataset and adversarial examples generated using the test dataset. We obtained new observations hinting that grokking is a common phenomenon in deep learning that is not restricted to particular tasks or DNN initialization. Upon this discovery, we delved into DNNs geometry to isolate the root cause of both delayed generalization and robustness which we attributed to the DNN's linear region migration that occurs in the latest phase of training. Again, the observation of such migration of the DNN partition is a new discovery of its own right. We hope that our analysis has provided novel insights into DNNs training dynamics from which grokking naturally emerges. 
While we empirically study the local complexity dynamics, a theoretical justification behind the double descent behavior is lacking.
At a high level, it is clear that the classification function being learned has its curvature concentrated at the decision boundary and approximation theory would normally dictate a free-form spline to therefore concentrate its partition regions around the decision boundary to minimize approximation error. However, it is not clear why that migration occurs so late in the training process, and we hope to study that in future research. We also see empirical evidence of region migration while using Adam as the optimizer. The training dynamics of stochastic gradient descent, as well as sharpness aware minimization \cite{andriushchenko2022towards} can also be studied using our framework. There can be possible connections between region migration and neural collapse \cite{papyan2020prevalence} which are not explored in this paper. The spline viewpoint of deep neural networks may provide strong geometric insights to assist in mechanistic understanding in future works as well. 


\section*{Impact Statement}
This paper presents work whose goal is to advance the field of Machine Learning. One takeaway of this work is that training Deep Neural Networks longer may lead to increased robustness. Training networks especially foundation models for longer may have potential societal consequences in terms of carbon emissions. 

\section*{Acknowledgements}
Humayun and Baraniuk were supported by NSF grants CCF1911094, IIS-1838177, and IIS-1730574; ONR grants N00014-
18-12571, N00014-20-1-2534, and MURI N00014-20-1-2787;
AFOSR grant FA9550-22-1-0060; and a Vannevar Bush Faculty
Fellowship, ONR grant N00014-18-1-2047.

\bibliography{example_paper}

\begin{thebibliography}{37}
\providecommand{\natexlab}[1]{#1}
\providecommand{\url}[1]{\texttt{#1}}
\expandafter\ifx\csname urlstyle\endcsname\relax
  \providecommand{\doi}[1]{doi: #1}\else
  \providecommand{\doi}{doi: \begingroup \urlstyle{rm}\Url}\fi

\bibitem[Andriushchenko \& Flammarion(2022)Andriushchenko and Flammarion]{andriushchenko2022towards}
Andriushchenko, M. and Flammarion, N.
\newblock Towards understanding sharpness-aware minimization.
\newblock In \emph{International Conference on Machine Learning}, pp.\  639--668. PMLR, 2022.

\bibitem[Balestriero \& Baraniuk(2018)Balestriero and Baraniuk]{balestriero2018spline}
Balestriero, R. and Baraniuk, R.
\newblock A spline theory of deep networks.
\newblock In \emph{Proc. ICML}, pp.\  374--383, 2018.

\bibitem[Balestriero \& Baraniuk(2022)Balestriero and Baraniuk]{balestriero2022batch}
Balestriero, R. and Baraniuk, R.~G.
\newblock Batch normalization explained.
\newblock \emph{arXiv preprint arXiv:2209.14778}, 2022.

\bibitem[Balestriero \& LeCun(2023)Balestriero and LeCun]{balestriero2023police}
Balestriero, R. and LeCun, Y.
\newblock Police: Provably optimal linear constraint enforcement for deep neural networks.
\newblock In \emph{ICASSP 2023-2023 IEEE International Conference on Acoustics, Speech and Signal Processing (ICASSP)}, pp.\  1--5. IEEE, 2023.

\bibitem[Barak et~al.(2022)Barak, Edelman, Goel, Kakade, Malach, and Zhang]{barak2022hidden}
Barak, B., Edelman, B., Goel, S., Kakade, S., Malach, E., and Zhang, C.
\newblock Hidden progress in deep learning: Sgd learns parities near the computational limit.
\newblock \emph{Advances in Neural Information Processing Systems}, 35:\penalty0 21750--21764, 2022.

\bibitem[Bartlett et~al.(2019)Bartlett, Harvey, Liaw, and Mehrabian]{bartlett2019nearly}
Bartlett, P.~L., Harvey, N., Liaw, C., and Mehrabian, A.
\newblock Nearly-tight vc-dimension and pseudodimension bounds for piecewise linear neural networks.
\newblock \emph{The Journal of Machine Learning Research}, 20\penalty0 (1):\penalty0 2285--2301, 2019.

\bibitem[Croce \& Hein(2020)Croce and Hein]{croce2020reliable}
Croce, F. and Hein, M.
\newblock Reliable evaluation of adversarial robustness with an ensemble of diverse parameter-free attacks.
\newblock In \emph{International conference on machine learning}, pp.\  2206--2216. PMLR, 2020.

\bibitem[Gamba et~al.(2022)Gamba, Chmielewski-Anders, Sullivan, Azizpour, and Bjorkman]{gamba2022all}
Gamba, M., Chmielewski-Anders, A., Sullivan, J., Azizpour, H., and Bjorkman, M.
\newblock Are all linear regions created equal?
\newblock In \emph{AISTATS}, pp.\  6573--6590, 2022.

\bibitem[Garbin et~al.(2020)Garbin, Zhu, and Marques]{garbin2020dropout}
Garbin, C., Zhu, X., and Marques, O.
\newblock Dropout vs. batch normalization: an empirical study of their impact to deep learning.
\newblock \emph{Multimedia Tools and Applications}, 79:\penalty0 12777--12815, 2020.

\bibitem[Hanin \& Rolnick(2019)Hanin and Rolnick]{hanin2019complexity}
Hanin, B. and Rolnick, D.
\newblock Complexity of linear regions in deep networks.
\newblock \emph{arXiv preprint arXiv:1901.09021}, 2019.

\bibitem[Humayun et~al.(2022)Humayun, Balestriero, and Baraniuk]{humayun2022polarity}
Humayun, A.~I., Balestriero, R., and Baraniuk, R.
\newblock Polarity sampling: Quality and diversity control of pre-trained generative networks via singular values.
\newblock In \emph{CVPR}, pp.\  10641--10650, 2022.

\bibitem[Humayun et~al.(2023{\natexlab{a}})Humayun, Balestriero, Balakrishnan, and Baraniuk]{Humayun_2023_CVPR}
Humayun, A.~I., Balestriero, R., Balakrishnan, G., and Baraniuk, R.~G.
\newblock Splinecam: Exact visualization and characterization of deep network geometry and decision boundaries.
\newblock In \emph{Proceedings of the IEEE/CVF Conference on Computer Vision and Pattern Recognition (CVPR)}, pp.\  3789--3798, June 2023{\natexlab{a}}.

\bibitem[Humayun et~al.(2023{\natexlab{b}})Humayun, Balestriero, Balakrishnan, and Baraniuk]{humayun2023splinecam}
Humayun, A.~I., Balestriero, R., Balakrishnan, G., and Baraniuk, R.~G.
\newblock Splinecam: Exact visualization and characterization of deep network geometry and decision boundaries.
\newblock In \emph{Proceedings of the IEEE/CVF Conference on Computer Vision and Pattern Recognition}, pp.\  3789--3798, 2023{\natexlab{b}}.

\bibitem[Humayun et~al.(2023{\natexlab{c}})Humayun, Casco-Rodriguez, Balestriero, and Baraniuk]{humayun2023provable}
Humayun, A.~I., Casco-Rodriguez, J., Balestriero, R., and Baraniuk, R.
\newblock Provable instance specific robustness via linear constraints.
\newblock In \emph{2nd AdvML Frontiers Workshop at International Conference on Machine Learning 2023}, 2023{\natexlab{c}}.

\bibitem[Ilyas et~al.(2019)Ilyas, Santurkar, Tsipras, Engstrom, Tran, and Madry]{ilyas2019adversarial}
Ilyas, A., Santurkar, S., Tsipras, D., Engstrom, L., Tran, B., and Madry, A.
\newblock Adversarial examples are not bugs, they are features.
\newblock \emph{Advances in neural information processing systems}, 32, 2019.

\bibitem[Ioffe \& Szegedy(2015)Ioffe and Szegedy]{ioffe2015batch}
Ioffe, S. and Szegedy, C.
\newblock Batch normalization: Accelerating deep network training by reducing internal covariate shift.
\newblock \emph{arXiv preprint arXiv:1502.03167}, 2015.

\bibitem[Ji et~al.(2022)Ji, Pascanu, Hjelm, Lakshminarayanan, and Vedaldi]{ji2022test}
Ji, X., Pascanu, R., Hjelm, R.~D., Lakshminarayanan, B., and Vedaldi, A.
\newblock Test sample accuracy scales with training sample density in neural networks.
\newblock In \emph{Conference on Lifelong Learning Agents}, pp.\  629--646. PMLR, 2022.

\bibitem[Kubo et~al.(2019)Kubo, Banno, Manabe, and Minoji]{kubo2019implicit}
Kubo, M., Banno, R., Manabe, H., and Minoji, M.
\newblock Implicit regularization in over-parameterized neural networks.
\newblock \emph{arXiv preprint arXiv:1903.01997}, 2019.

\bibitem[Li et~al.(2022)Li, Jin, Zhong, Hopcroft, and Wang]{li2022robust}
Li, B., Jin, J., Zhong, H., Hopcroft, J., and Wang, L.
\newblock Why robust generalization in deep learning is difficult: Perspective of expressive power.
\newblock \emph{Advances in Neural Information Processing Systems}, 35:\penalty0 4370--4384, 2022.

\bibitem[Liu et~al.(2022)Liu, Michaud, and Tegmark]{liu2022omnigrok}
Liu, Z., Michaud, E.~J., and Tegmark, M.
\newblock Omnigrok: Grokking beyond algorithmic data.
\newblock \emph{arXiv preprint arXiv:2210.01117}, 2022.

\bibitem[Madry et~al.(2017)Madry, Makelov, Schmidt, Tsipras, and Vladu]{madry2017towards}
Madry, A., Makelov, A., Schmidt, L., Tsipras, D., and Vladu, A.
\newblock Towards deep learning models resistant to adversarial attacks.
\newblock \emph{arXiv preprint arXiv:1706.06083}, 2017.

\bibitem[Montufar et~al.(2014)Montufar, Pascanu, Cho, and Bengio]{montufar2014number}
Montufar, G.~F., Pascanu, R., Cho, K., and Bengio, Y.
\newblock On the number of linear regions of deep neural networks.
\newblock In \emph{NeurIPS}, pp.\  2924--2932, 2014.

\bibitem[Nanda et~al.(2023)Nanda, Chan, Lieberum, Smith, and Steinhardt]{nanda2023progress}
Nanda, N., Chan, L., Lieberum, T., Smith, J., and Steinhardt, J.
\newblock Progress measures for grokking via mechanistic interpretability.
\newblock \emph{arXiv preprint arXiv:2301.05217}, 2023.

\bibitem[Novak et~al.(2018)Novak, Bahri, Abolafia, Pennington, and Sohl-Dickstein]{novak2018sensitivity}
Novak, R., Bahri, Y., Abolafia, D.~A., Pennington, J., and Sohl-Dickstein, J.
\newblock Sensitivity and generalization in neural networks: an empirical study.
\newblock \emph{arXiv preprint arXiv:1802.08760}, 2018.

\bibitem[Olah et~al.(2020)Olah, Cammarata, Schubert, Goh, Petrov, and Carter]{olah2020zoom}
Olah, C., Cammarata, N., Schubert, L., Goh, G., Petrov, M., and Carter, S.
\newblock Zoom in: An introduction to circuits.
\newblock \emph{Distill}, 5\penalty0 (3):\penalty0 e00024--001, 2020.

\bibitem[Papyan et~al.(2020)Papyan, Han, and Donoho]{papyan2020prevalence}
Papyan, V., Han, X., and Donoho, D.~L.
\newblock Prevalence of neural collapse during the terminal phase of deep learning training.
\newblock \emph{Proceedings of the National Academy of Sciences}, 117\penalty0 (40):\penalty0 24652--24663, 2020.

\bibitem[Power et~al.(2022)Power, Burda, Edwards, Babuschkin, and Misra]{power2022grokking}
Power, A., Burda, Y., Edwards, H., Babuschkin, I., and Misra, V.
\newblock Grokking: Generalization beyond overfitting on small algorithmic datasets.
\newblock \emph{arXiv preprint arXiv:2201.02177}, 2022.

\bibitem[Qin et~al.(2019)Qin, Martens, Gowal, Krishnan, Dvijotham, Fawzi, De, Stanforth, and Kohli]{qin2019adversarial}
Qin, C., Martens, J., Gowal, S., Krishnan, D., Dvijotham, K., Fawzi, A., De, S., Stanforth, R., and Kohli, P.
\newblock Adversarial robustness through local linearization.
\newblock \emph{Advances in Neural Information Processing Systems}, 32, 2019.

\bibitem[Raghu et~al.(2017)Raghu, Poole, Kleinberg, Ganguli, and Dickstein]{raghu2017expressive}
Raghu, M., Poole, B., Kleinberg, J., Ganguli, S., and Dickstein, J.~S.
\newblock On the expressive power of deep neural networks.
\newblock In \emph{ICML}, pp.\  2847--2854, 2017.

\bibitem[Tan et~al.(2023)Tan, LeJeune, Mason, Javadi, and Baraniuk]{tan2023blessing}
Tan, J., LeJeune, D., Mason, B., Javadi, H., and Baraniuk, R.~G.
\newblock A blessing of dimensionality in membership inference through regularization.
\newblock In \emph{International Conference on Artificial Intelligence and Statistics}, pp.\  10968--10993. PMLR, 2023.

\bibitem[Toth et~al.(2017)Toth, O'Rourke, and Goodman]{toth2017handbook}
Toth, C.~D., O'Rourke, J., and Goodman, J.~E.
\newblock \emph{Handbook of discrete and computational geometry}.
\newblock CRC press, 2017.

\bibitem[Tsipras et~al.(2018)Tsipras, Santurkar, Engstrom, Turner, and Madry]{tsipras2018robustness}
Tsipras, D., Santurkar, S., Engstrom, L., Turner, A., and Madry, A.
\newblock Robustness may be at odds with accuracy.
\newblock \emph{arXiv preprint arXiv:1805.12152}, 2018.

\bibitem[Varma et~al.(2023)Varma, Shah, Kenton, Kram{\'a}r, and Kumar]{varma2023explaining}
Varma, V., Shah, R., Kenton, Z., Kram{\'a}r, J., and Kumar, R.
\newblock Explaining grokking through circuit efficiency.
\newblock \emph{arXiv preprint arXiv:2309.02390}, 2023.

\bibitem[Xu \& Mannor(2012)Xu and Mannor]{xu2012robustness}
Xu, H. and Mannor, S.
\newblock Robustness and generalization.
\newblock \emph{Machine learning}, 86:\penalty0 391--423, 2012.

\bibitem[Xu et~al.(2021)Xu, Ili{\'c}, Ir{\v{s}}i{\v{c}}, Klav{\v{z}}ar, and Li]{xu2021comparing}
Xu, K., Ili{\'c}, A., Ir{\v{s}}i{\v{c}}, V., Klav{\v{z}}ar, S., and Li, H.
\newblock Comparing wiener complexity with eccentric complexity.
\newblock \emph{Discrete Applied Mathematics}, 290:\penalty0 7--16, 2021.

\bibitem[Xu et~al.(2023)Xu, Wang, Frei, Vardi, and Hu]{xu2023benign}
Xu, Z., Wang, Y., Frei, S., Vardi, G., and Hu, W.
\newblock Benign overfitting and grokking in relu networks for xor cluster data.
\newblock \emph{arXiv preprint arXiv:2310.02541}, 2023.

\bibitem[You et~al.(2021)You, Balestriero, Lu, Kou, Shi, Zhang, Wu, Lin, and Baraniuk]{you2021max}
You, H., Balestriero, R., Lu, Z., Kou, Y., Shi, H., Zhang, S., Wu, S., Lin, Y., and Baraniuk, R.
\newblock Max-affine spline insights into deep network pruning.
\newblock \emph{arXiv preprint arXiv:2101.02338}, 2021.

\end{thebibliography}
\bibliographystyle{icml2024}

\newpage
\appendix
\onecolumn
\section{Empirical analysis of our proposed method}
\label{sec:evaluating-complexity-approximation}

Computing the exact number of linear regions or piecewise-linear hyperplane intersections for an deep network with N-dimensional input space neighborhood has combinatorial complexity and therefore is intractable. This is one of the key motivations behind our approximation method. 

\textbf{MLP with zero bias.} To validate our method, we start with a toy experiment with a linear MLP with width $400$, depth $50$, $784$ dimensional input space, initialized with zero bias and random weights. In such a setting all the layerwise hyperplanes intersect the origin at their input space. We compute the LC around the input space origin using our method, for neighborhoods of varying radius $r=\{0.0001,0.001,0.01,0.1,1,10\}$ and dimensionality $P=\{2,10,25,50,100,200\}$. For all the trials, our method recovers all the layerwise hyperplane intersections, even with a neighborhood dimensionality of $P=2$.

\textbf{Non-Zero Bias Random MLP with shifting neighborhood.} For a randomly initialized MLP, we expect to see lower local complexity as we move away from the origin \cite{hanin2019complexity}. For this experiment we take a width $100$ depth $18$ MLP with input dimensionality $d=784$, Leaky-ReLU activation with negative slope $0.01$. We start by computing LC at the origin $[0]^d$, and linearly shift towards the vector $[10]^d$. We see that for all the settings, shifting away from the origin reduces LC. LC gets saturated with increasing $P$, showing that lower dimensional neighborhoods can be good enough for approximating LC. Increasing $r$ on the other hand, increases LC and reduces LC variations between shifts, since the neighborhood becomes larger and LC becomes less local.

\begin{figure}
    \centering
    \includegraphics[width=.35\textwidth]{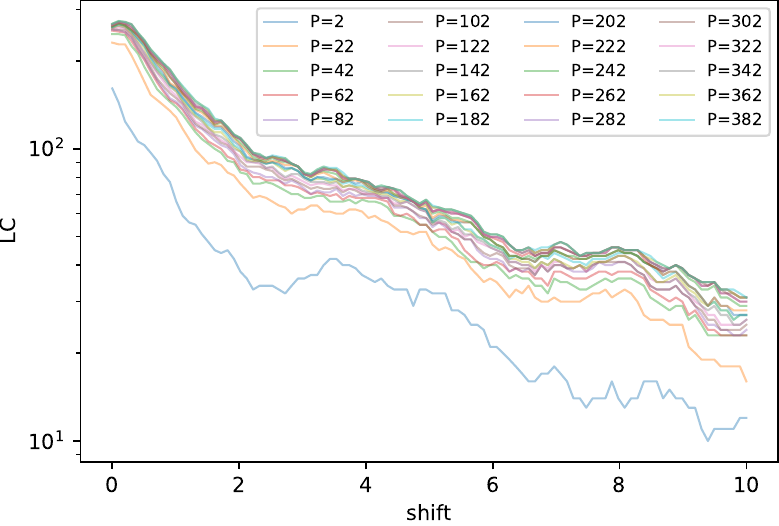}
    \includegraphics[width=.35\textwidth]{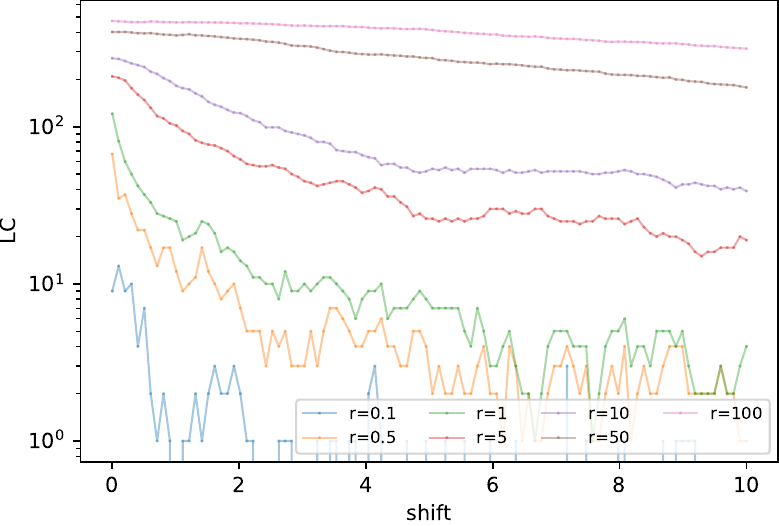}
    \caption{LC for a $P$ dimensional neighborhood with radius $r$ while being shifted from the origin $[0]^d$ to vector $[10]^d$. In \textbf{left}, we vary $P$ with fixed $r=5$ while on \textbf{right} we vary $r$ for fixed $P=20$. We see that for all the settings, shifting away from the origin reduces LC. The increase of LC with the neighborhood dimensionality $P$ gets saturated as we increase $P$, showing that lower dimensional neighborhoods can be good enough for approximating LC. Increasing $r$ on the other hand, increases LC and reduces LC variations between shifts, since the neighborhood becomes larger and LC becomes less local.}
    \label{fig:appendix_shift_randmlp}
\end{figure}

\begin{figure}
    \centering
    \includegraphics[width=.4\linewidth]{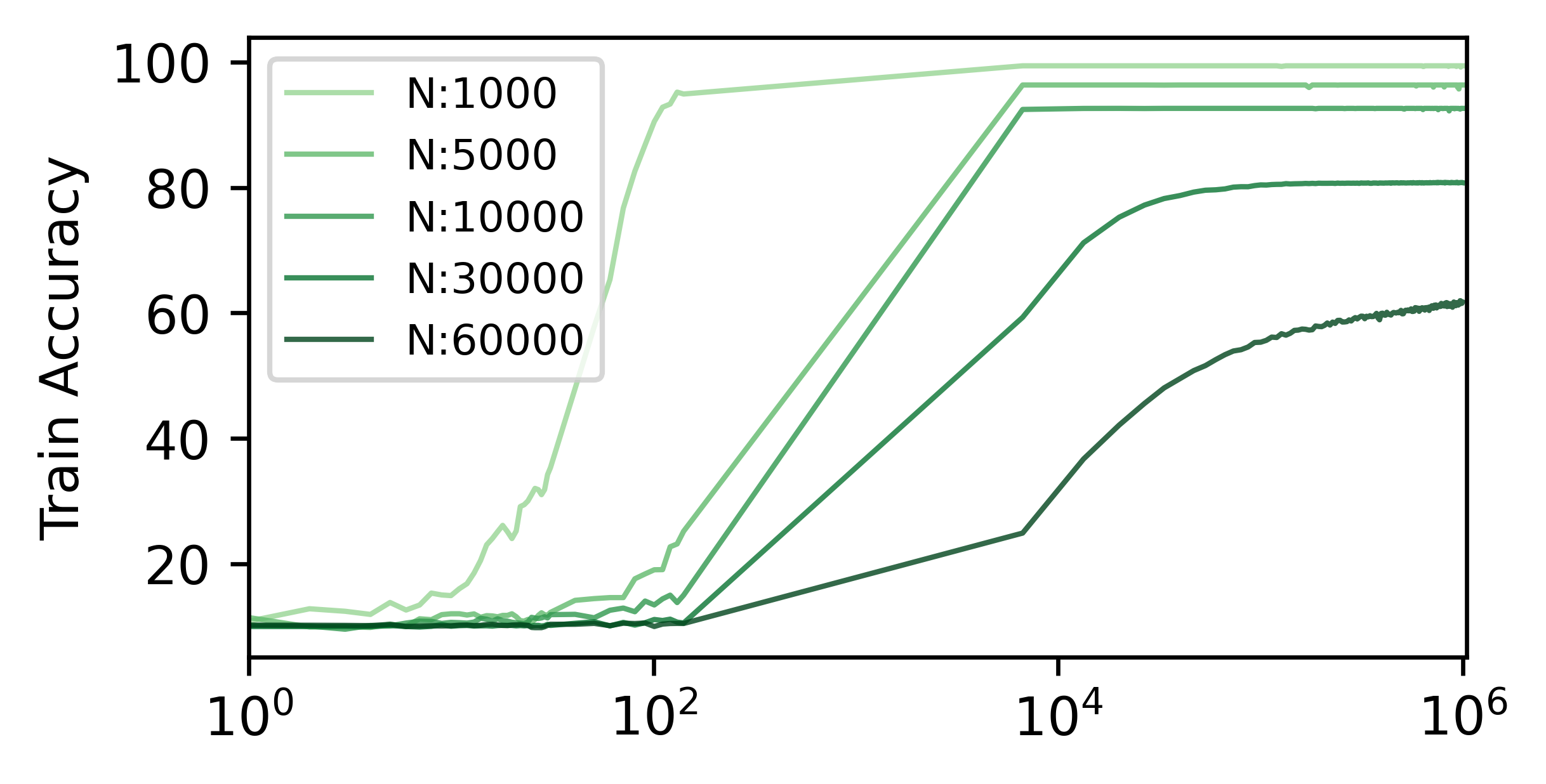}%
  \includegraphics[width=.4\linewidth]{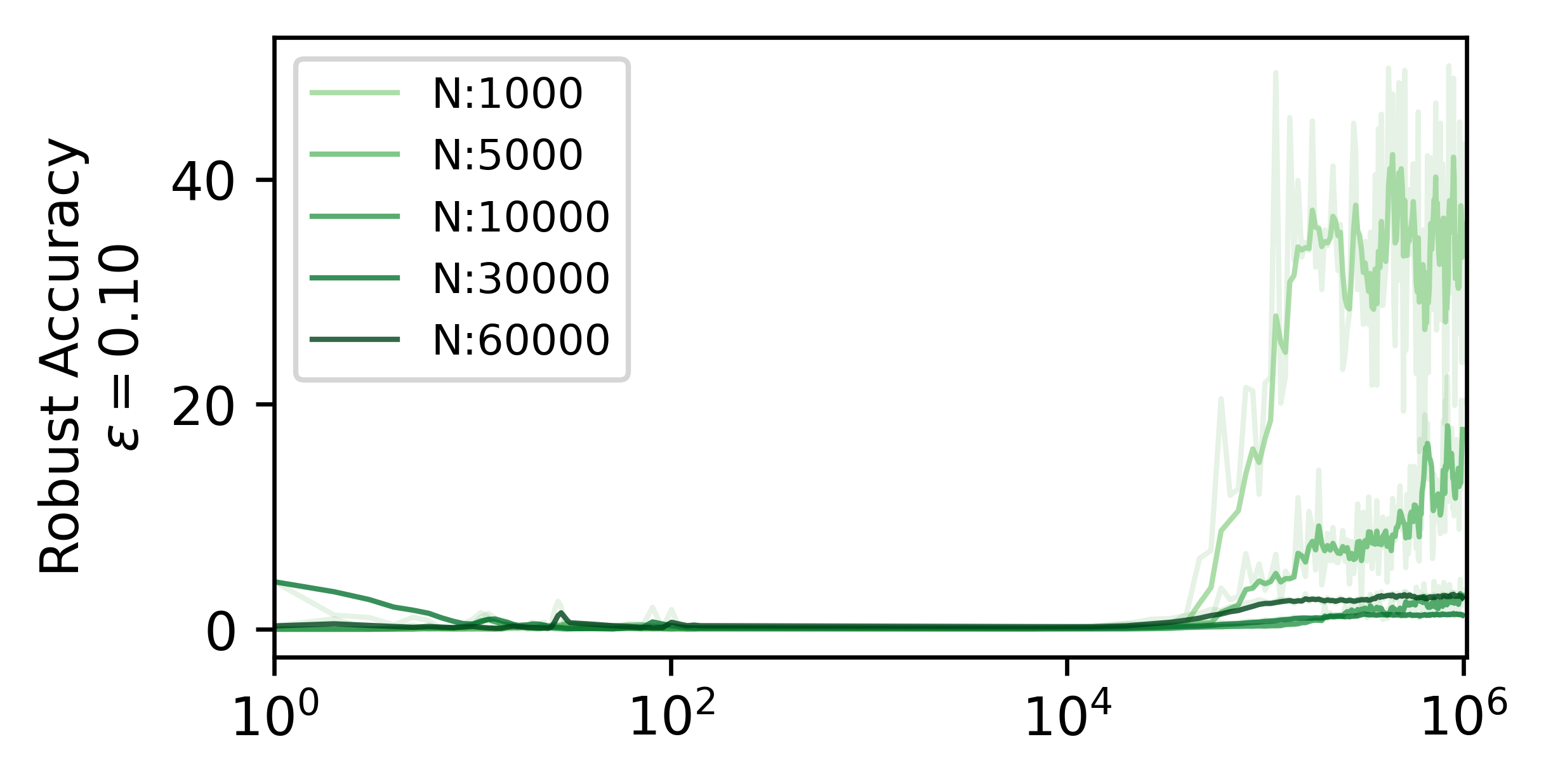}\\
    \begin{minipage}{\linewidth}
        \centering
        \small{Optimization Steps}
    \end{minipage}
    \caption{Training accuracy and robust accuracy for the networks trained on randomly labeled MNIST samples presented in \cref{fig:mlp_rand_dataswp}.}
    \label{fig:rand_mnist_adv}
\end{figure}

\begin{figure}
    \centering
    \includegraphics[width=.3\linewidth]{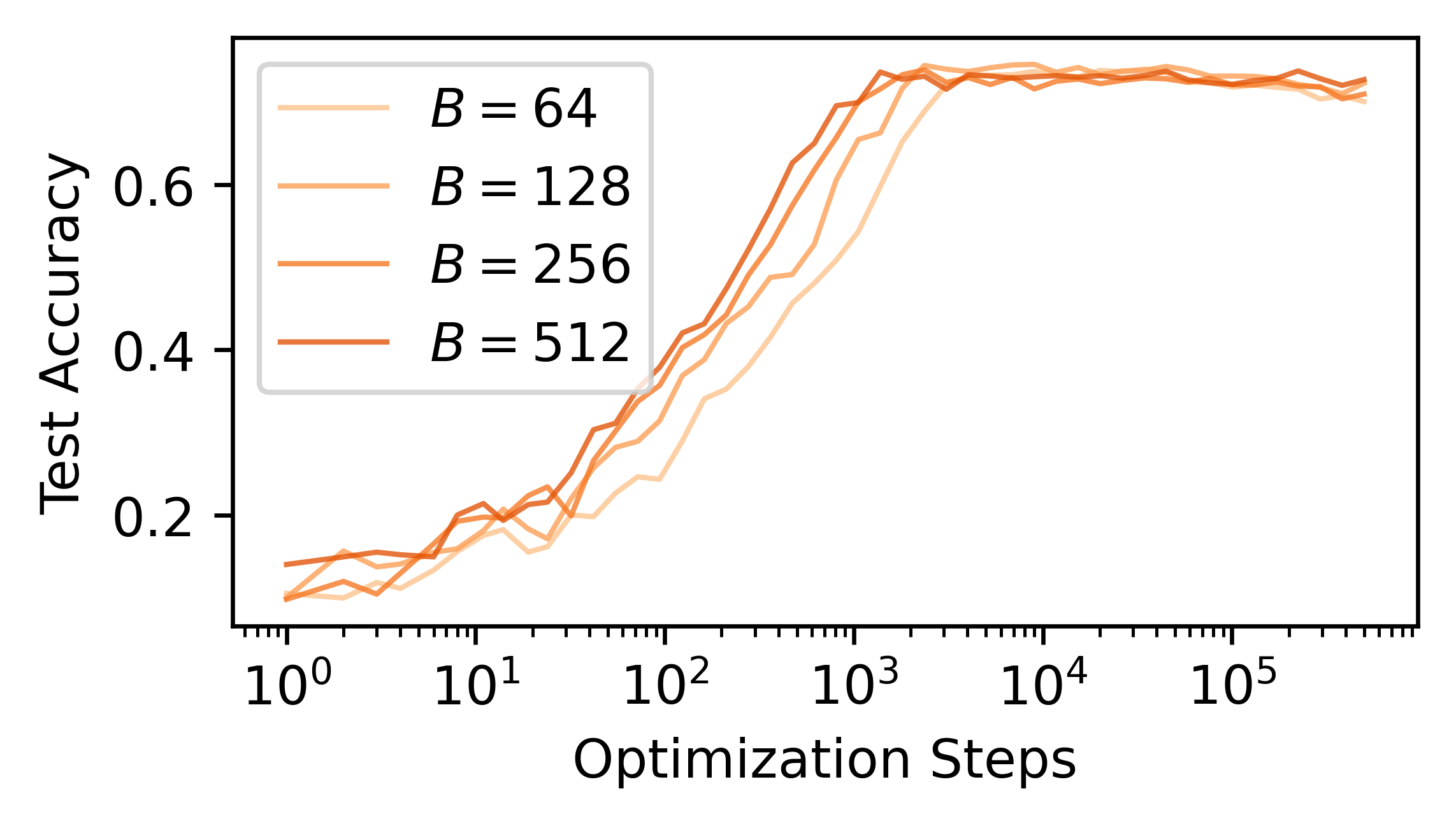}%
    \includegraphics[width=.3\linewidth]{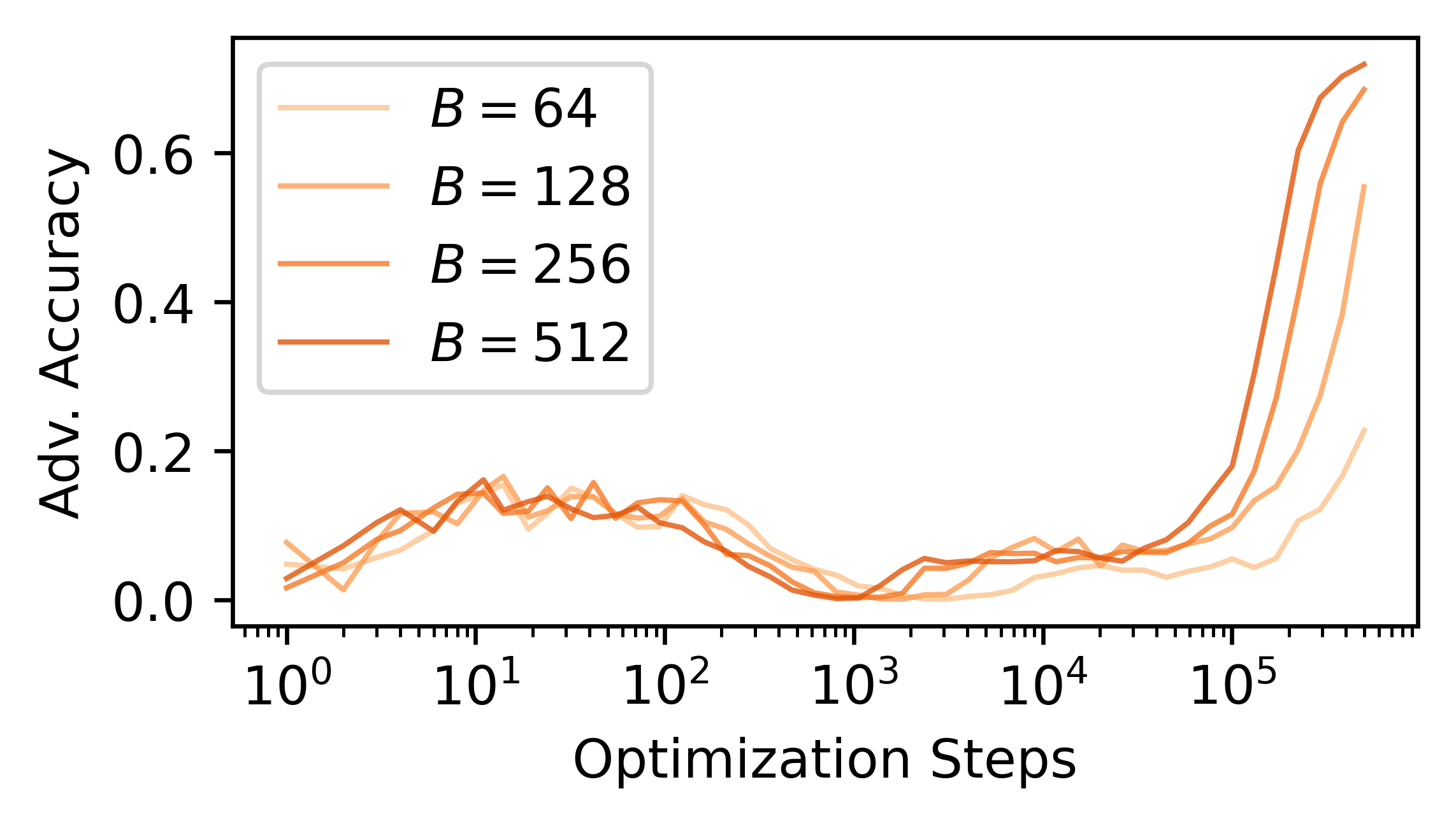}%
    \includegraphics[width=.3\linewidth]{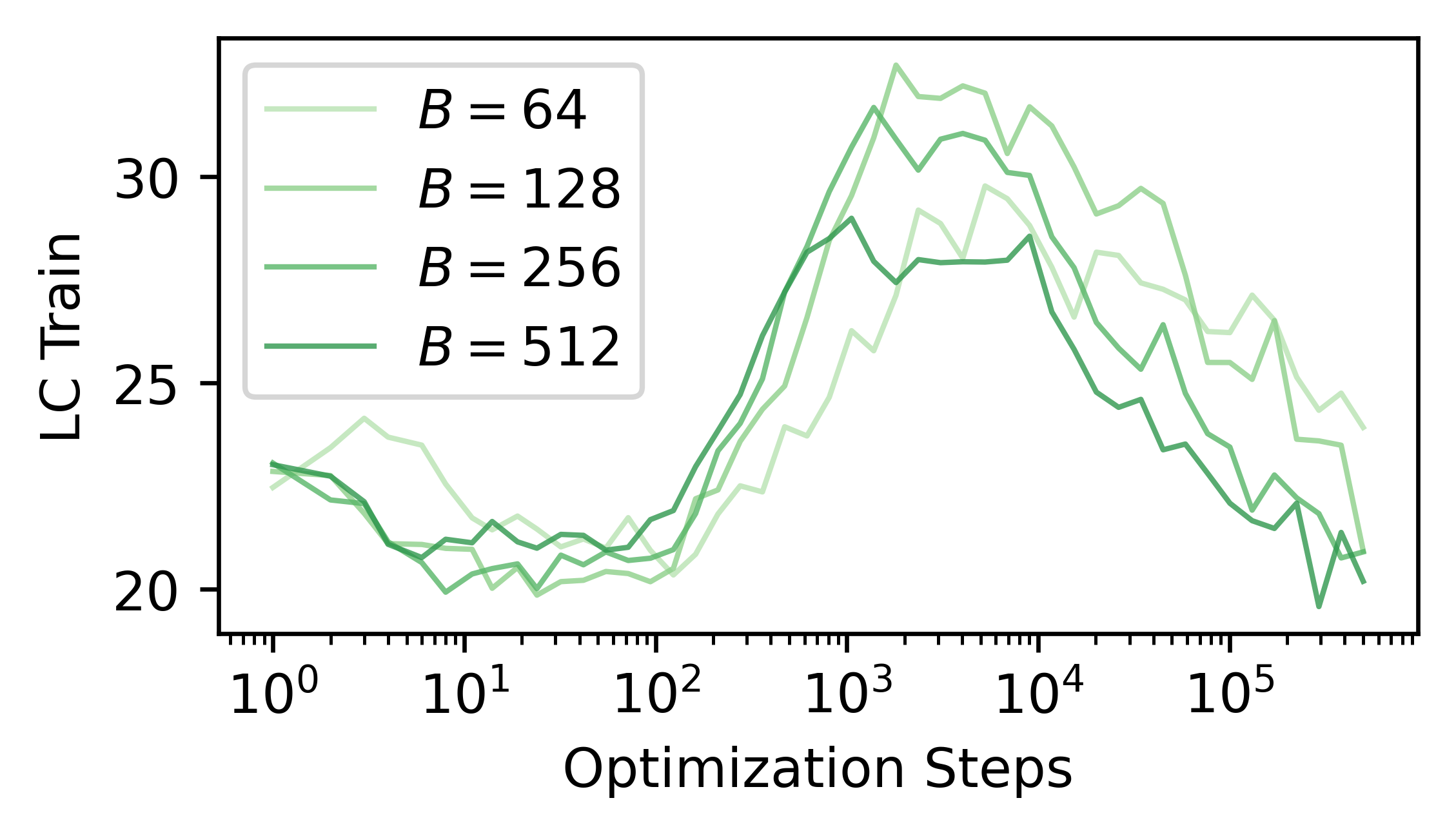}%
    \caption{Increasing the batch-size expedites grokking. This indicates that reduced SGD noise allows region migration to occur earlier in training.}
    \label{fig:batch_sweep}
\end{figure}

\begin{figure}
    \centering
    \includegraphics[width=0.3\linewidth]{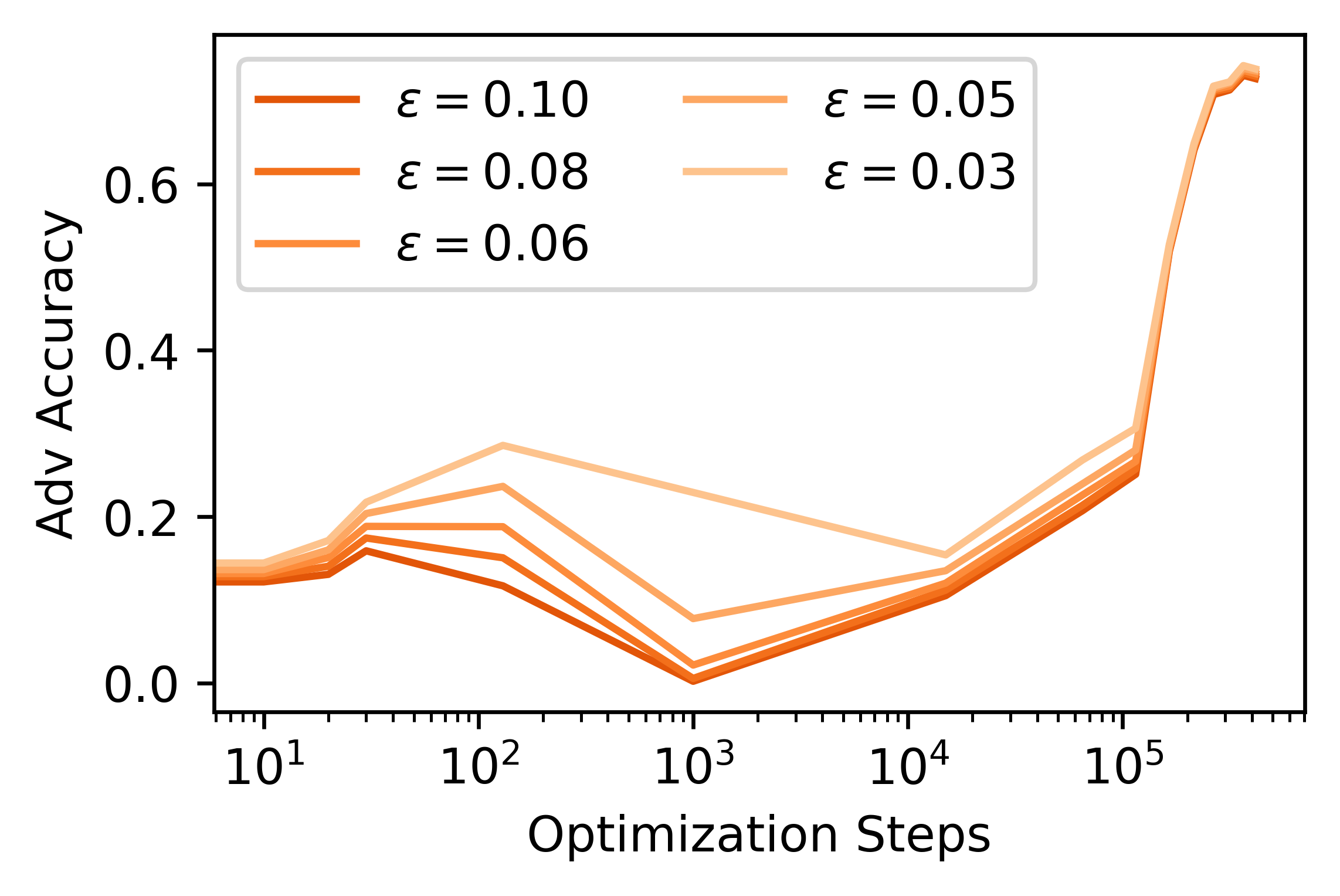}
    \caption{\textbf{Grokking stronger adversarial attacks.} We see that during delayed generalization, the robustness to Auto-Attack \cite{croce2020reliable} also increases. This shows the universality of delayed robustness.}
    \label{fig:apgd_attack}
\end{figure}


\begin{figure}
    \centering
\includegraphics[ width=0.25\linewidth ]{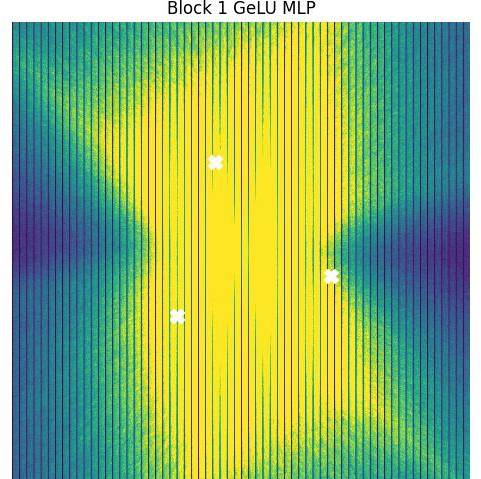}%
\includegraphics[ width=0.25\linewidth ]{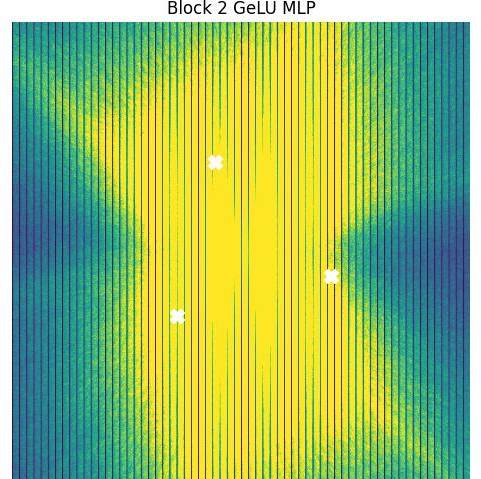}%
\includegraphics[ width=0.25\linewidth ]{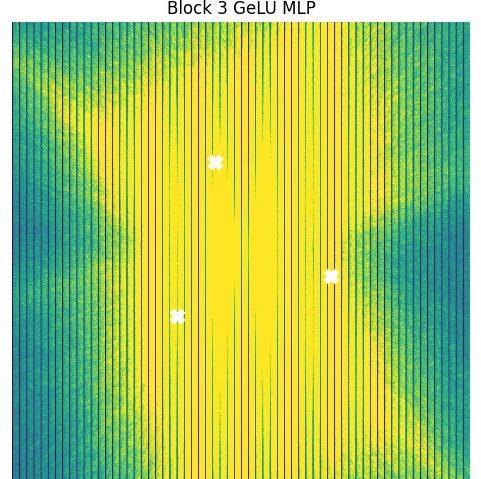}%
\includegraphics[ width=0.25\linewidth ]{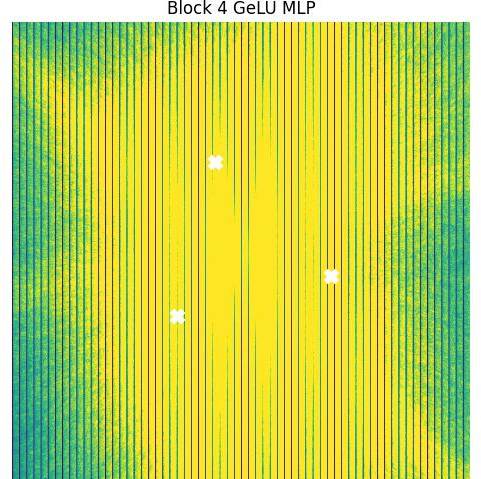}\\
\includegraphics[ width=0.25\linewidth ]{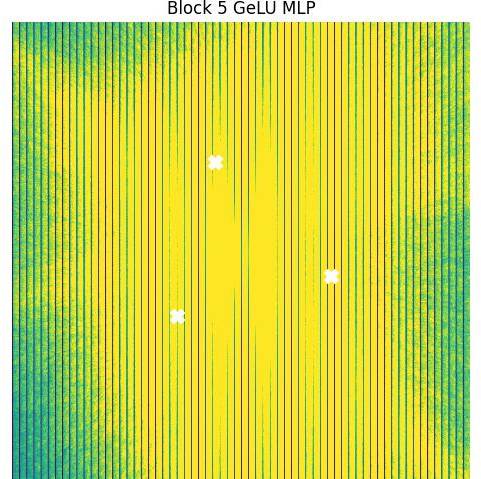}%
\includegraphics[ width=0.25\linewidth ]{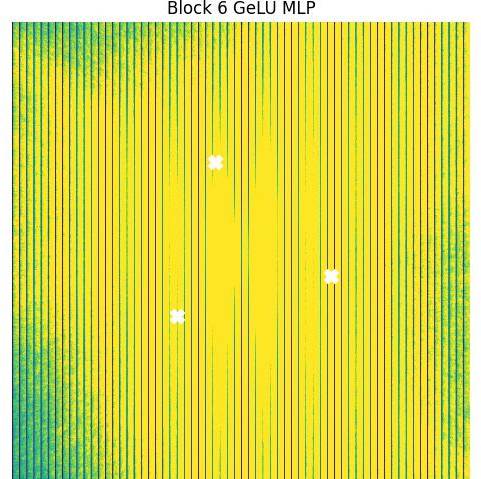}%
\includegraphics[ width=0.25\linewidth ]{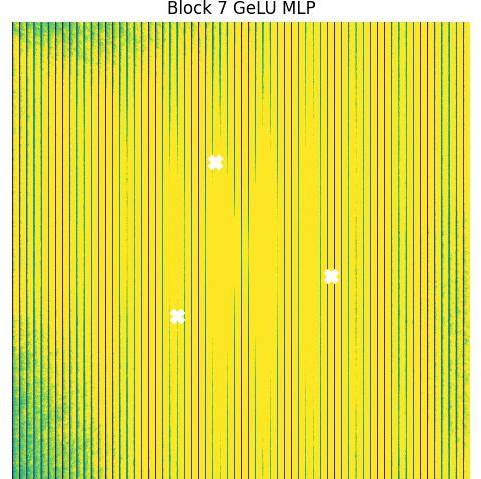}%
\includegraphics[ width=0.25\linewidth ]{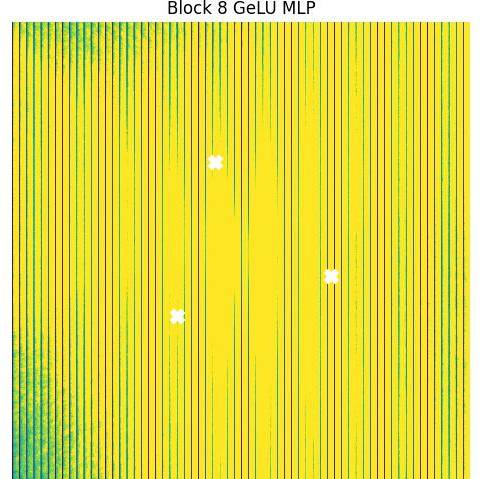}\\
\includegraphics[ width=0.25\linewidth ]{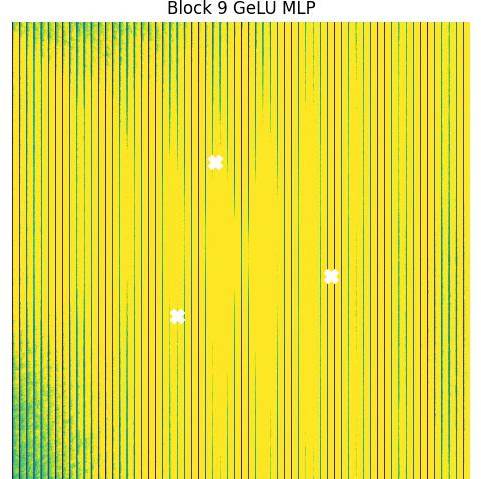}%
\includegraphics[ width=0.25\linewidth ]{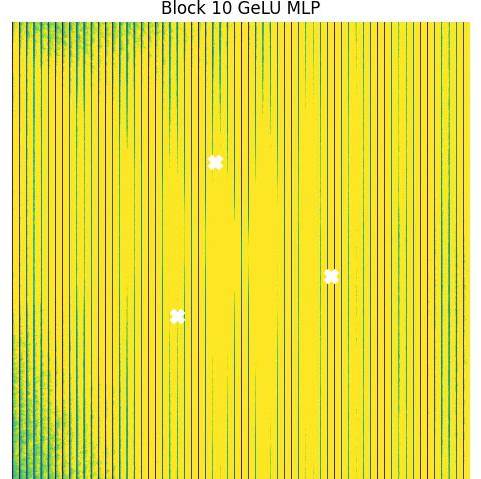}%
\includegraphics[ width=0.25\linewidth ]{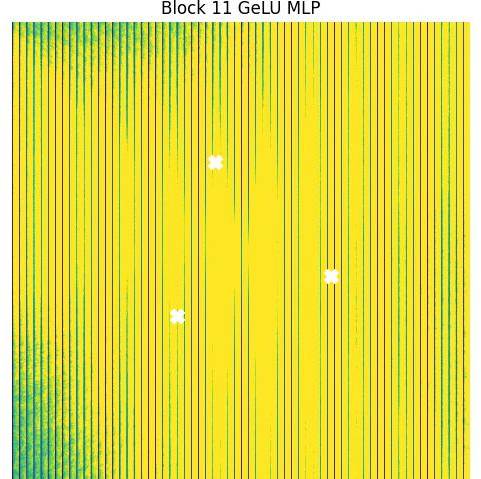}%
\includegraphics[ width=0.25\linewidth ]{figures/llm_grokking/beforeGrokked_h12l12_12_block_gmlp-min.jpg}%
    \caption{Token embedding space LC on a 2D subspace intersecting three random training points. We visualize the LC layerwise for the GeLU activated MLP layers inside each of the 12 blocks of the LLM for which we present training dynamics in \cref{fig:grokking-llm}. The LC is computed after 197 optimization steps, during the peak ascent. We see that LC on this subspace is very high especially close to the data points. LC values are clamped to a maximum of 150.}
    \label{fig:shakespeare_ascentpeak}
\end{figure}

\begin{figure}
    \centering
\includegraphics[ width=0.25\linewidth ]{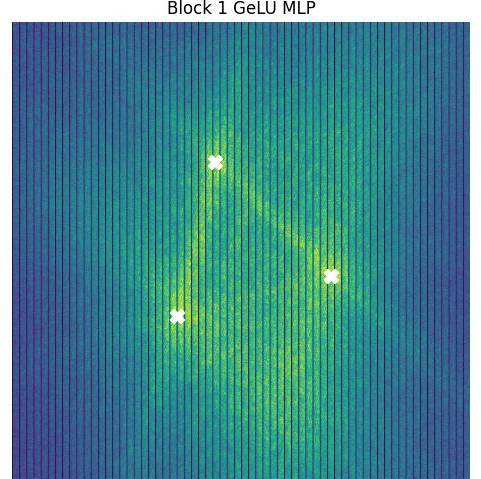}%
\includegraphics[ width=0.25\linewidth ]{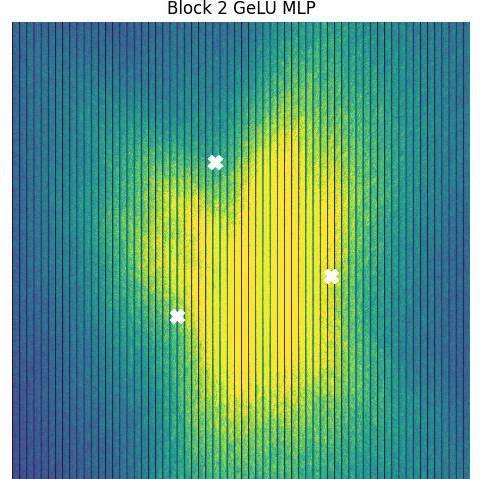}%
\includegraphics[ width=0.25\linewidth ]{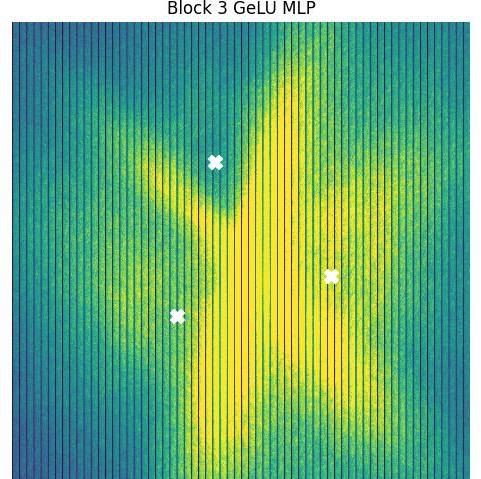}%
\includegraphics[ width=0.25\linewidth ]{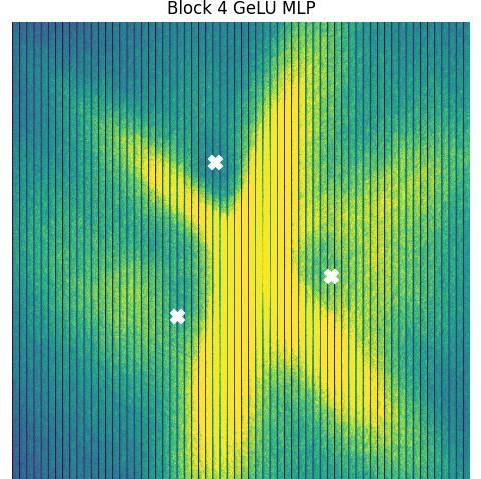}\\
\includegraphics[ width=0.25\linewidth ]{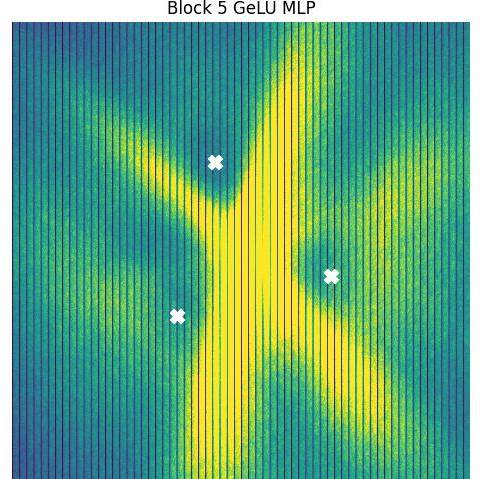}%
\includegraphics[ width=0.25\linewidth ]{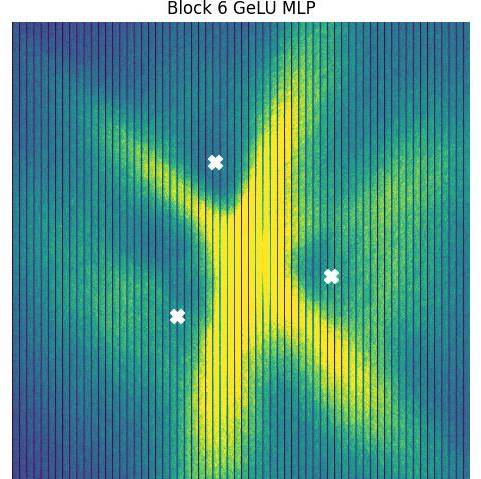}%
\includegraphics[ width=0.25\linewidth ]{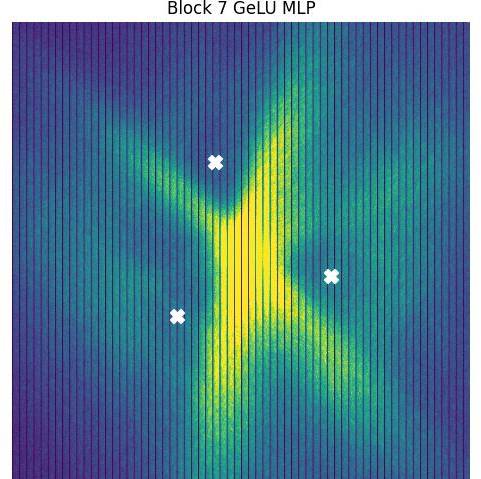}%
\includegraphics[ width=0.25\linewidth ]{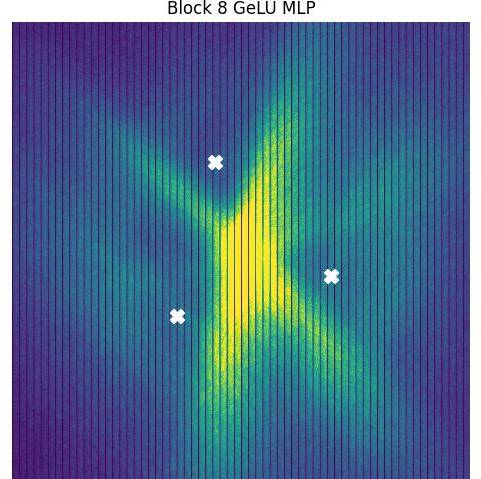}\\
\includegraphics[ width=0.25\linewidth ]{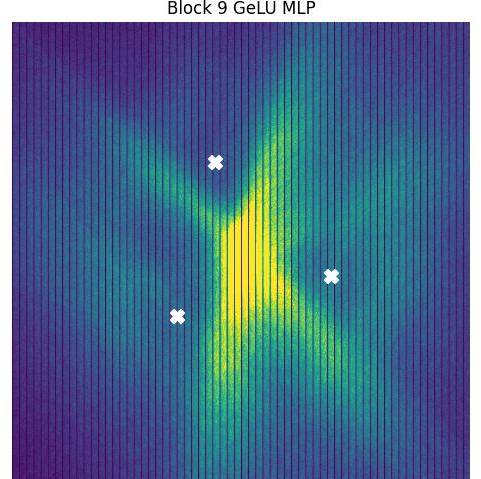}%
\includegraphics[ width=0.25\linewidth ]{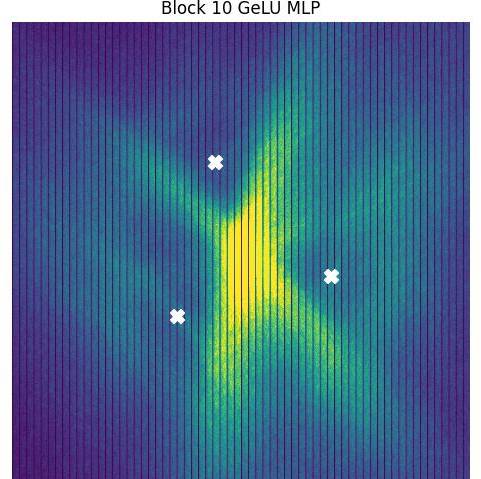}%
\includegraphics[ width=0.25\linewidth ]{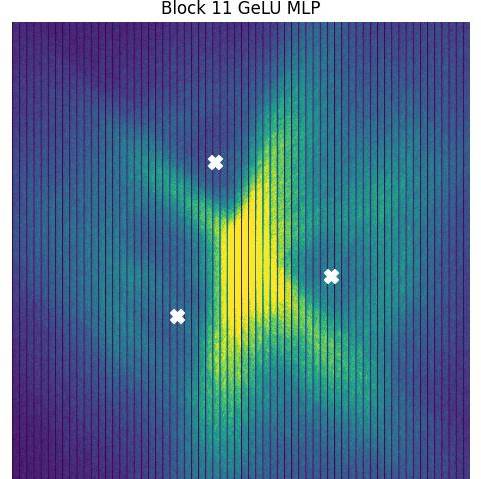}%
\includegraphics[width=0.25\linewidth ]{figures/llm_grokking/Grokked_h12l12_12_block_gmlp-min.jpg}%
    \caption{Token embedding space LC on a 2D subspace intersecting three random training points. We visualize the LC layerwise for the GeLU activated MLP layers inside each of the 12 blocks of the LLM for which we present training dynamics in \cref{fig:grokking-llm}. The LC is computed after 372759 optimization steps, therefore during the second LC descent. We see that LC on this subspace is concentrated away from the training points, especially for the deeper layers. This indicates that region migration occurs in LLMs as well, leading to delayed robustness. LC values are clamped to a maximum of 150.}
    \label{fig:shakespeare_aftergrok}
\end{figure}

\begin{figure}[!b]
    \centering
    \includegraphics[width=.5\linewidth]{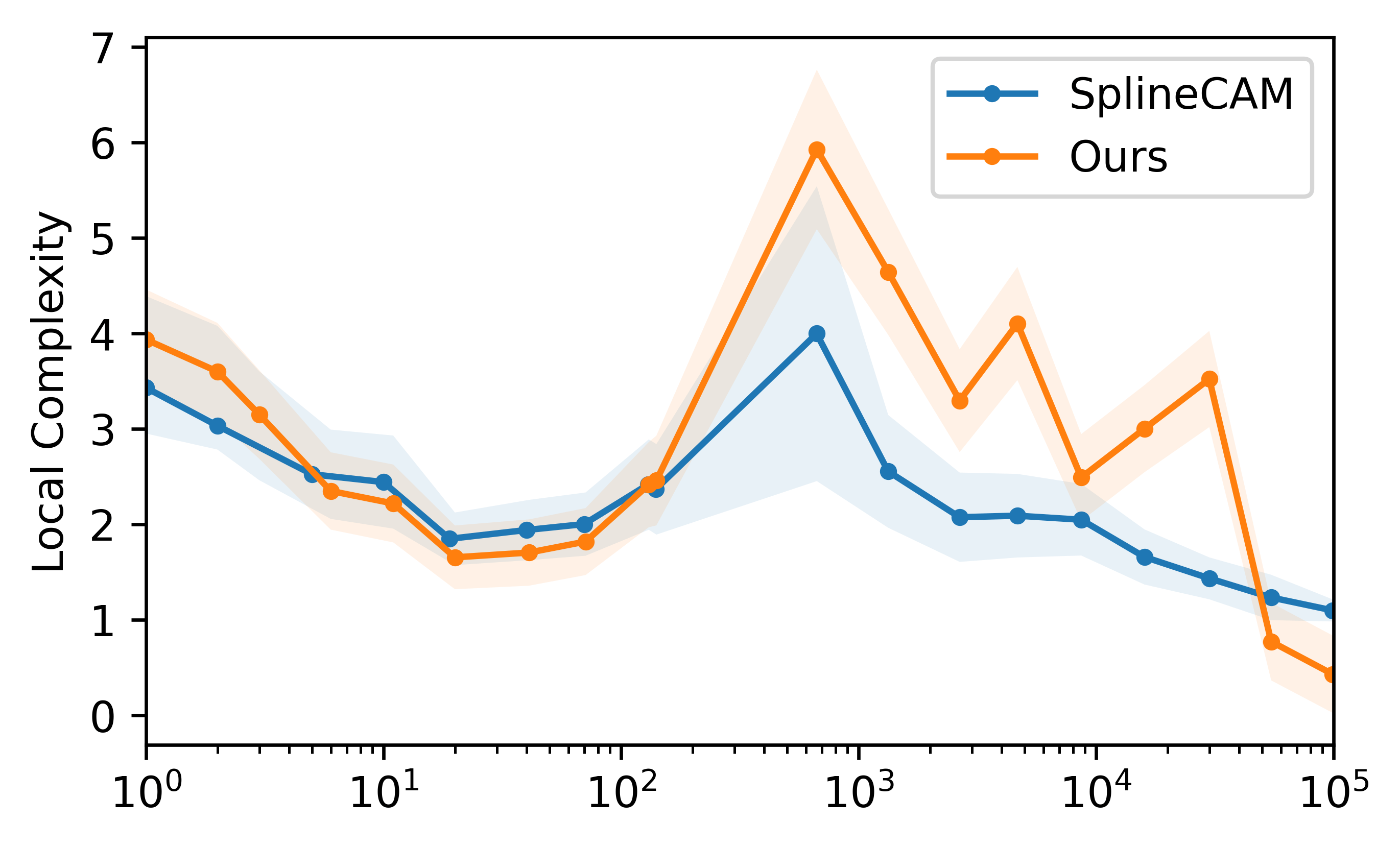}
    \caption{Comparing the local complexity measured in terms of the number of linear regions computed exactly by SplineCAM \cite{Humayun_2023_CVPR} and number of hyperplane cuts by our proposed method. Both methods exhibit the double descent behavior.}
    \label{fig:compare_with_exact}
\end{figure}

\textbf{Trained MLP comparison with SplineCam.} For non-linear MLPs, we compare with the exact computation method Splinecam \cite{Humayun_2023_CVPR}. We take a depth 3 width 200 MLP and train it on MNIST for 100K training steps. For $20$ different training checkpoints, we compute the local complexity in terms of the number of linear regions computed via SplineCam and number of hyperplane intersections via our proposed method. We compute the local complexity for $500$ different training samples. For both our method and SplineCam we consider a radius of $0.001$. For our method, we consider a neighborhood with dimensionality $P=25$. We present the LC trajectories in Fig.~\ref{fig:compare_with_exact}. We can see that for both methods the local complexity follows a similar trend with a double descent behavior. 

\begin{figure}
    \centering
    \includegraphics[width=.35\textwidth]{figures/eccentricity_cifar10_trainpnts_radswp.pdf}
    \includegraphics[width=.35\textwidth]{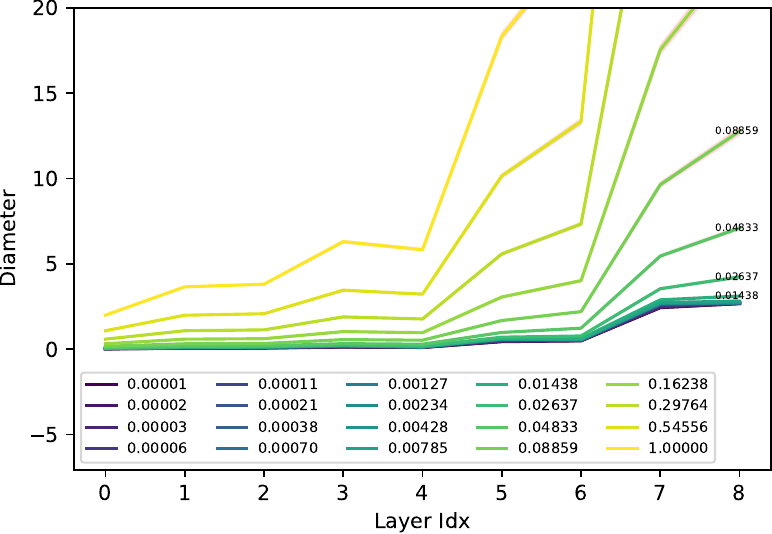}
    \caption{Change of avg. eccentricity and diameter \cite{xu2021comparing} of the input space neighborhood by different layers of a CNN trained on the CIFAR10 dataset. For different sampling radius $r$ of the sampled input space neighborhood $V$, the change of eccentricity and diameter denotes how much deformation the neighborhood undergoes between layers. Here, layer $0$ corresponds to the input space neighborhood. Numbers are averaged over neighborhoods sampled for $1000$ training points from CIFAR10.
    For larger radius the deformation increases with depth exponentially. For $r\leq 0.014$ deformation is lower, indicating that smaller radius neighborhoods are reliable for LC computation on deeper networks. Confidence interval shown in red, is almost imperceptible.}
    \label{fig:appendix_deformation_cnn}
\end{figure}

\textbf{Deformation of neighborhood by deep networks.} As mentioned in \cref{sec:evaluating-complexity-approximation}, we compute the local complexity in a layerwise fashion by embedding a neighborhood $conv(V)$ into the input space for any layer and computing the number of hyperplane intersections with $conv(V^\ell)$, where $V^\ell$ is the embedded vertices at the input space of layer $\ell$. The approximation of local complexity is therefore subject to the deformation induced by each layer to $conv(V)$. To measure deformation by layers $1$ to $\ell-1$, we consider the undirected graph formed by the vertices $V^\ell$ and compute the average eccentricity and diameter of the graphs \cite{xu2021comparing}. Eccentricity for any vertex $v$ of a graph, is denoted by the maximum shortest path distance between $v$ and all the connected vertices in the graph. The diameter is the maximum eccentricity over vertices of a graph. Recall from \cref{sec:evaluating-complexity-approximation} that $conv(V)$ where $ V = \{x \pm rv_p : p = 1...P \}$ for an input space point $x$, is a cross-polytope of dimensionality $P$, where only two vertices are sampled from any of the orthogonal directions $v_p$. Therefore, all vertices share edges with each other except for pairs $\{(x + rv_p, x - rv_p) : p = 1...P \}$. Given such connectivity, we compute the average eccentricity and diameter of neighborhoods $conv(V^\ell)$ around $1000$ training points from CIFAR10 for a trained CNN (Fig.~\ref{fig:appendix_deformation_cnn}). We see that for larger $r$ both of the deformation metrics exponentially increase, where as for $r \leq 0.014$ the deformation is lower and more stable. This shows that for lower $r$ our LC approximation for deeper CNN networks would be better since the neighborhood does not get deformed significantly. 

\begin{figure}
    \centering
    \includegraphics[width=.6\linewidth]{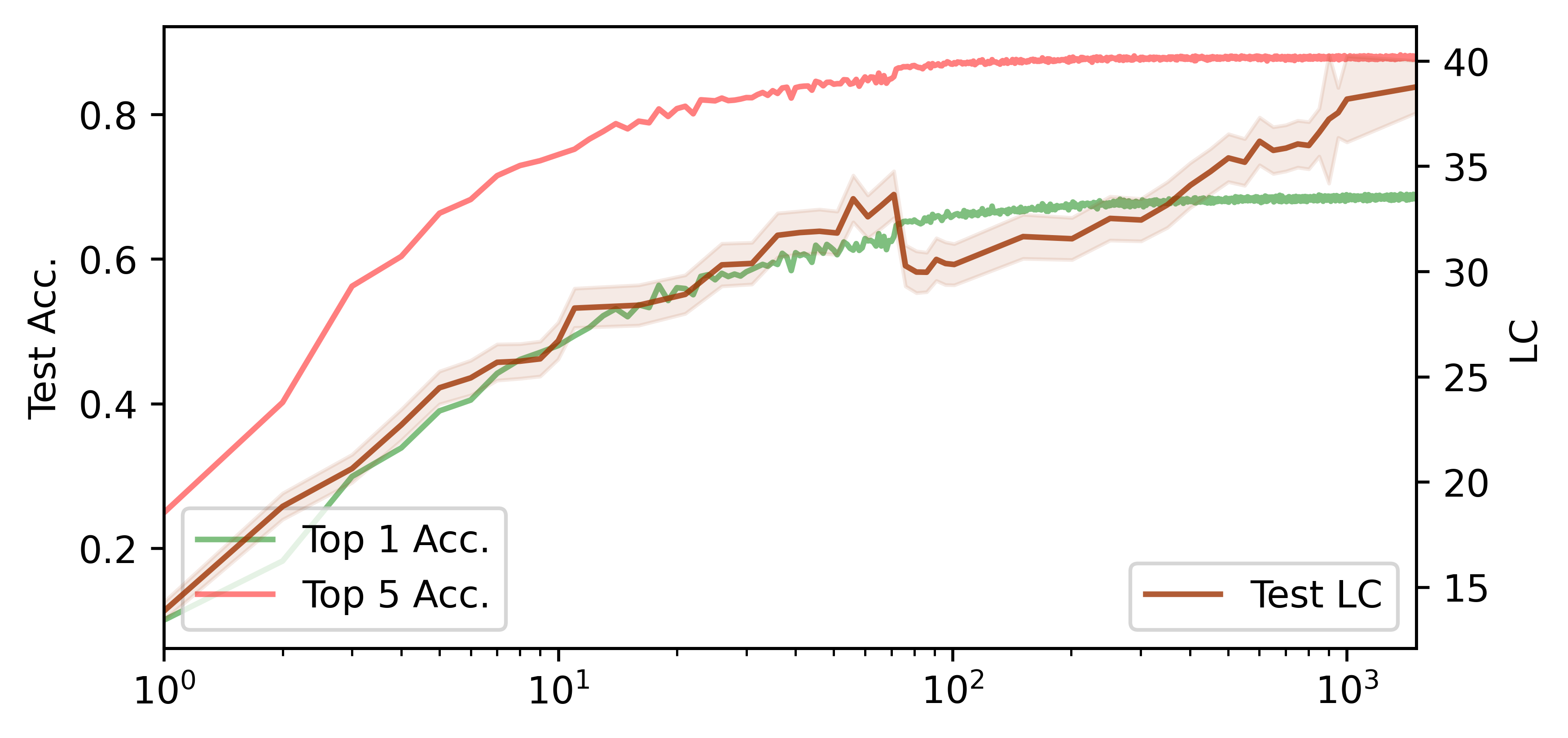}
    \caption{Training a ResNet18 with batchnorm on Imagenet Full. LC is computed only on test points using 1000 test set samples. Computing LC 1000 samples takes approx. 28s on an RTX 8000.}
    \label{fig:batchnorm_imagenet}
\end{figure}

\begin{figure}
    \centering
    \includegraphics[width=.35\textwidth]{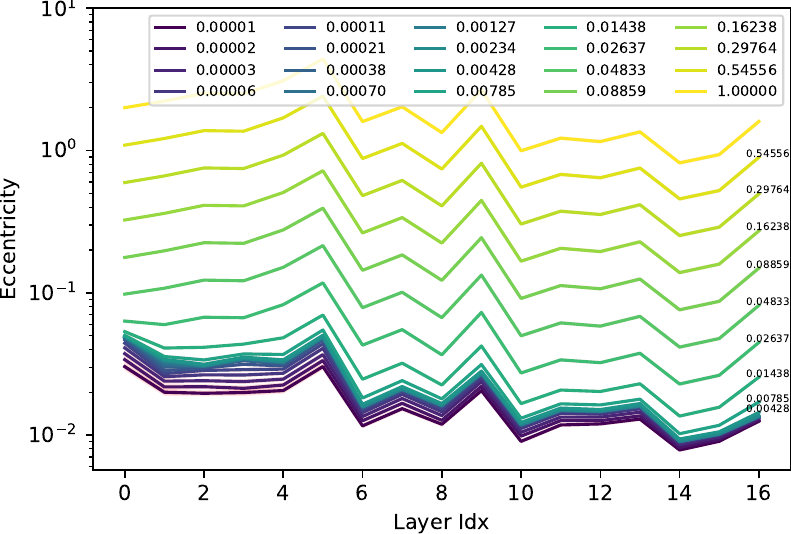}
    \includegraphics[width=.35\textwidth]{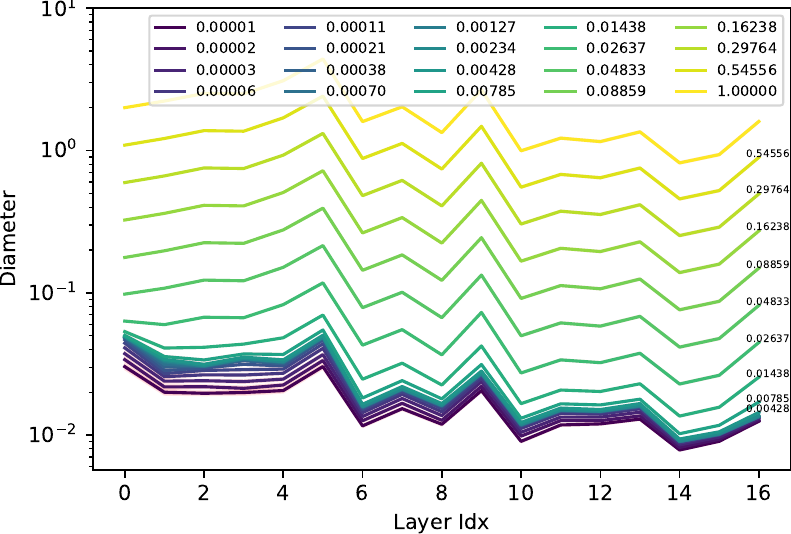}
    \caption{Change of avg. eccentricity and diameter \cite{xu2021comparing} of the input space neighborhood by different layers of a ResNet18 trained on the CIFAR10 dataset, similar to the setting of Fig.~\ref{fig:appendix_deformation_cnn}. Resnet deforms the input neighborhood by reducing the avg. eccentricity and diameter of the neighborhood graphs. For $r\leq 0.014$ deformation is lower, indicating that smaller radius neighborhoods are reliable for LC computation on deeper networks.}
    \label{fig:appendix_deformation_resnet}
\end{figure}

\section{Understanding Batch Normalization and its effect on the partition}
\label{sec:batchnorm-theorem}

Suppose the usual layer mapping is
\begin{equation}
    \vz_{\ell+1}= \va\left(\mW_{\ell}\vz_{\ell}+\vc_{\ell}\right), 
    \quad \ell=0,\dots,L-1
    \label{eq:no_BN2}
\end{equation}
While a host of different DNN architectures have been developed over the past several years, modern, high-performing DNNs nearly universally employ {\em batch normalization} (BN) \cite{ioffe2015batch} to center and normalize the entries of the feature maps using four additional parameters $\mu_\ell,\sigma_\ell,\beta_\ell,\gamma_\ell$.
Define $z_{\ell,k}$ as $k^{\rm th}$ entry of feature map $\vz_\ell$ of length $D_\ell$, 
$\vw_{\ell,k}$ as the $k^{\rm th}$ row of the weight matrix $\mW_\ell$, 
and $\mu_{\ell,k},\sigma_{\ell,k},\beta_{\ell,k},\gamma_{\ell,k}$ as the $k^{\rm th}$ entries of the BN parameter vectors $\mu_\ell,\sigma_\ell,\beta_\ell,\gamma_\ell$, respectively.
Then we can write the BN-equipped layer $\ell$ mapping extending (\ref{eq:no_BN}) as
\begin{equation}
    z_{\ell+1,k}=
    a\left(
    \frac{\left\langle \vw_{\ell,k},\vz_{\ell}\right\rangle-\mu_{\ell,k}}{\sigma_{\ell,k}}
    \: \gamma_{\ell,k} + \beta_{\ell,k}
    \right),k=1,\dots,D_\ell.
\label{eq:BN}
\end{equation}
The parameters $\mu_\ell,\sigma_\ell$ are computed as the element-wise mean and standard deviation of $\mW_\ell \vz_{\ell}$ for each mini-batch during training and for the entire training set during testing.
The parameters $\beta_\ell,\gamma_\ell$ are learned along with $\mW_\ell$ via SGD.\footnote{Note that the DNN bias $\vc_\ell$ from (\ref{eq:no_BN}) has been subsumed into $\mu_\ell$ and $\beta_\ell$.}
For each mini-batch $\sB$ during training, the BN parameters $\mu_{\ell}, \sigma_{\ell}$ are {\em calculated directly} as the mean and standard deviation of the current mini-batch feature maps $\mathcal{B}_\ell$
\begin{align}
&\mu_{\ell} 
\leftarrow
\frac{1}{|\sB_\ell|}\sum_{\vz_\ell \in \sB_\ell}  \mW_{\ell}\vz_\ell,
&\sigma_{\ell} 
\leftarrow
\sqrt{\frac{1}{|\mathcal{B}_\ell|}\sum_{\vz_\ell \in \sB_\ell}\big( \mW_{\ell}\vz_\ell-\mu_{\ell} \big)^2},
\label{eq:bnupdate}
\end{align}
where the right-hand side square is taken element-wise.
After SGD learning is complete, a final fixed ``test time'' mean $\overline{\mu}_{\ell}$ and standard deviation $\overline{\sigma}_{\ell}$ are computed using the above formulae over all of the training data,\footnote{or more commonly as an exponential moving average of the training mini-batch values.} i.e., with $\sB_\ell=\sX_\ell$.

The Euclidean distance from a point $\vv$ in layer $\ell$'s input space to the layer's $k^{\rm th}$ hyperplane $\sH_{\ell,k}$  is easily calculated as 
\begin{align}
    d(\vv,\sH_{\ell,k}) = \frac{\left | \langle \vw_{\ell,k},\vv\rangle - \mu_{\ell,k}\right | }{\|\vw_{\ell,k}\|_2}
    \label{eq:distance}
\end{align}
as long as $\|\vw_{\ell,k}\| >0$.

Then, the average squared distance between $\sH_{\ell,k}$ and a collection of points $\sV$ in layer $\ell$'s input space is given by
\begin{align}
\L_k(\mu_{\ell,k},\sV) = \frac{1}{|\sV|}\sum_{\vv \in \sV}
d\left(\vv,\sH_{\ell,k}\right)^2= \frac{\sigma_{\ell,k}^2}{\|\vw_{\ell,k}\|_2^2},
\label{eq:optimization-k}
\end{align}

\section{What affects the robust partition? \textit{Reprise}}

\begin{figure}
\begin{minipage}{0.03\linewidth}
\centering
    \rotatebox[]{90}{{\small{Local Complexity}}}
\end{minipage}%
\begin{minipage}{.94\textwidth}
    \centering
    \includegraphics[width=0.2\linewidth]{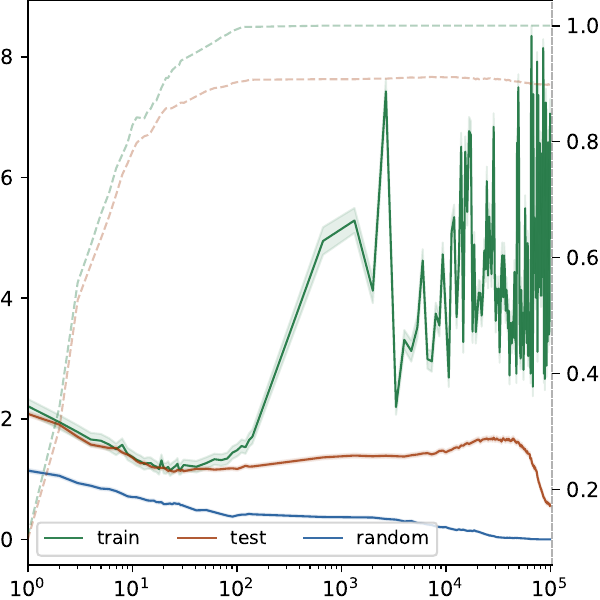}%
    \includegraphics[width=0.2\linewidth]{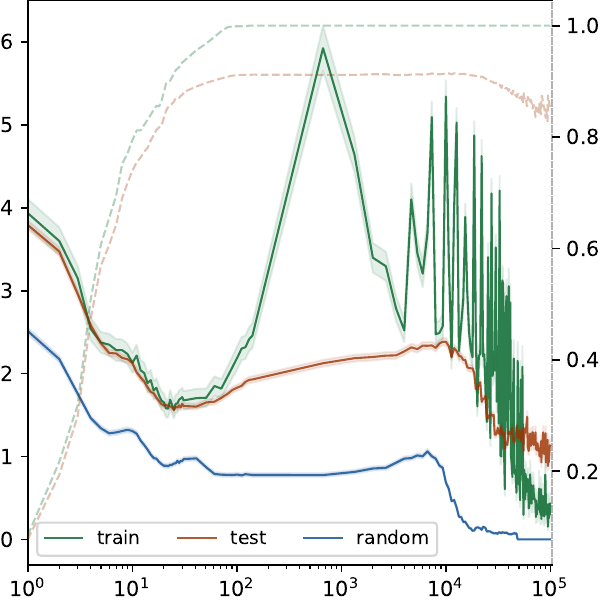}%
    \includegraphics[width=0.2\linewidth]{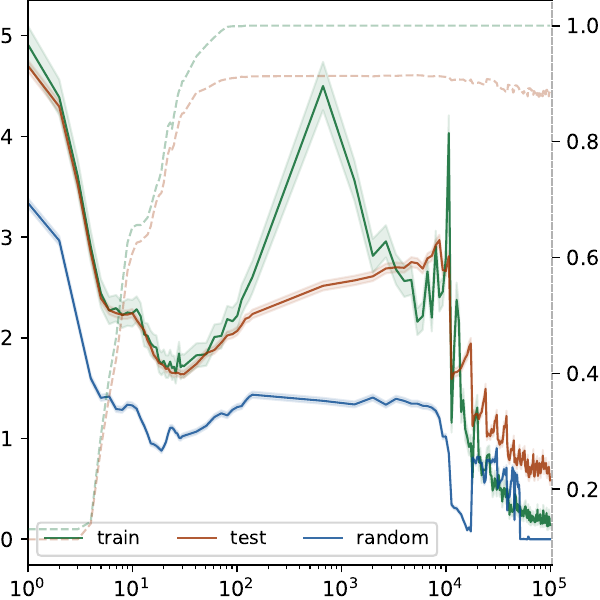}%
    \includegraphics[width=0.2\linewidth]{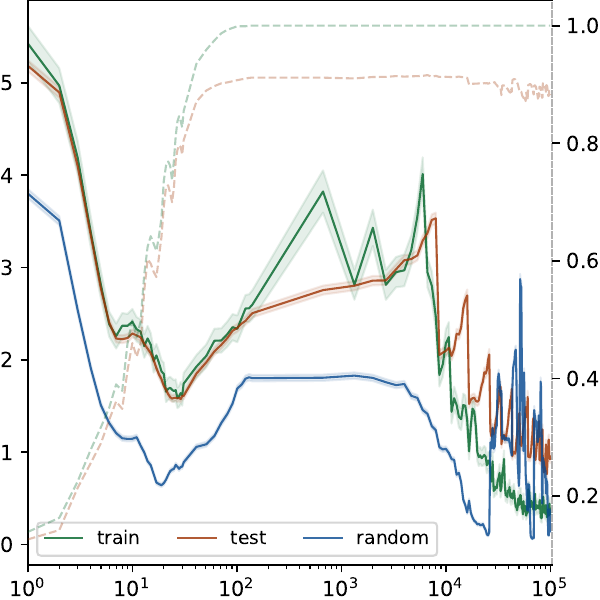}%
    \includegraphics[width=0.2\linewidth]{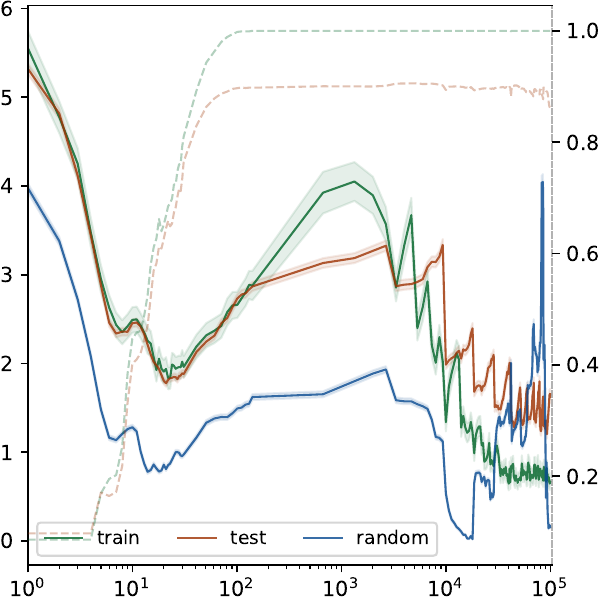}%
\end{minipage}%
\begin{minipage}{0.03\linewidth}
\centering
    \rotatebox[]{270}{{\small{Accuracy}}}
\end{minipage}\\
\begin{minipage}{\linewidth}
    \centering
    \vspace{1em}
    Optimization Steps
\end{minipage}
    \caption{MLP with width 200 and varying depth being trained on $1000$ samples from MNIST. Increasing the depth of the network decreases the sharpness of the LC peak during ascend phase. Deeper networks also tend to have a sharper decline in the training LC during region migration.}
    \label{fig:mlp-depth-sweep}
\end{figure}

\textbf{Depth.} In \cref{fig:mlp-depth-sweep} we plot LC during training on MNIST for Fully Connected Deep Networks with depth in $\{2,3,4,5\}$ and width $200$. In each plot, we show both LC as well as train-test accuracy. For all the depths, the accuracy on both the train and test sets peak during the first descent phase. 
During the ascent phase, we see that the train LC has a sharp ascent while the test and random LC do not. 

The difference as well as the sharpness of the ascent is reduced when increasing the depth of the network. This is visible for both fine and coarse $r$ scales. For the shallowest network, we can see a second descent in the coarser scale but not in the finer $r$ scale. This indicates that for the shallow network some regions closer to the training samples are retained during later stages of training. One thing to note is that during the ascent and second descent phase, there is a clear distinction between the train and test LC. \textit{This is indicative of membership inference fragility especially during latter phases of training.} It has previously been observed in membership inference literature \cite{tan2023blessing}, where early stopping has been used as a regularizer for membership inference. We believe the LC dynamics can shed a new light towards membership inference and the role of network complexity/capacity.

In \cref{fig:cnn-cifar10-batchnorm}, we plot the local complexity during training for CNNs trained on CIFAR10 with varying depths with and without batch normalization. The CNN architecture comprises of only convolutional layers except for one fully connected layer before output. Therefore when computing LC, we only take into account the convolutional layers in the network. Contrary to the MNIST experiments, we see that in this setting, the train-test LC are almost indistinguishable throughout training. We can see that the network train and test accuracy peaks during the ascent phase and is sustained during the second descent. It can also be noticed that increasing depth increases the max LC during the ascent phase for CNNs which is contrary to what we saw for fully connected networks on MNIST. The increase of density during ascent is all over the data manifold, contrasting to just the training samples for fully connected networks. 

In Appendix, we present layerwise visualization of the LC dynamics. We see that shallow layers have sharper peak during ascent phase, with distinct difference between train and test. For deeper layers however, the train vs test LC difference is negligible. 

\textbf{Width.} In \cref{fig:affect-width} we present results for a fully connected DNN with depth $3$ and width $\{20,100,500,1000,2000\}$. Networks with smaller width start from a low LC at initialization compared to networks that are wider. Therefore for small width networks the initial descent becomes imperceptible. We see that as we increase width from $20$ to $1000$ the ascent phase starts earlier as well as reaches a higher maximum LC. However overparameterizing the network by increasing the width further to $2000$, reduces the max LC during ascent, therefore reducing the crowding of neurons near training samples. \textit{This is a possible indication of how overparameterization performs implicit regularization \cite{kubo2019implicit}, by reducing non-linearity or local complexity concentration around training samples.}

\begin{figure}
\begin{minipage}{\linewidth}
    \centering
    \hspace{3em} Training Samples \hspace{8em} Test Samples \hspace{10em} Random Samples
\end{minipage}\\
\begin{minipage}{0.03\linewidth}
\centering
    \rotatebox[]{90}{{\small{Local Complexity}}}
\end{minipage}%
\begin{minipage}{.97\textwidth}
    \centering
    \includegraphics[width=0.33\linewidth]{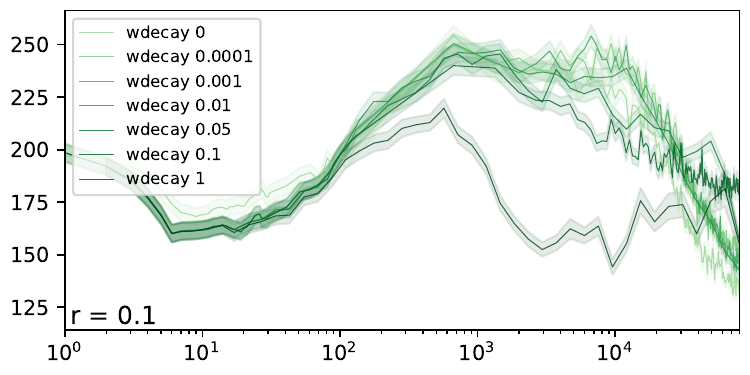}%
    \includegraphics[width=0.33\linewidth]{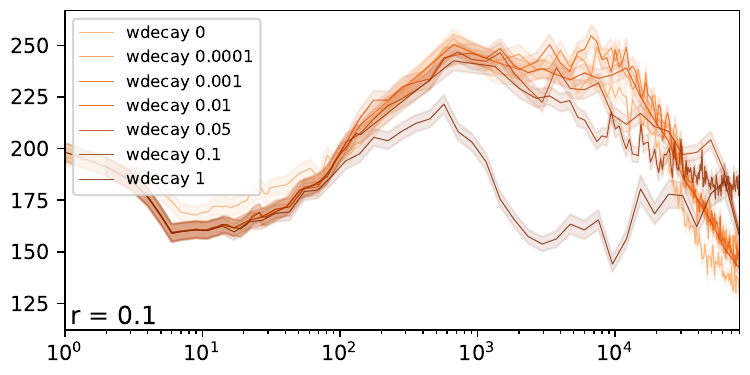}%
    \includegraphics[width=0.33\linewidth]{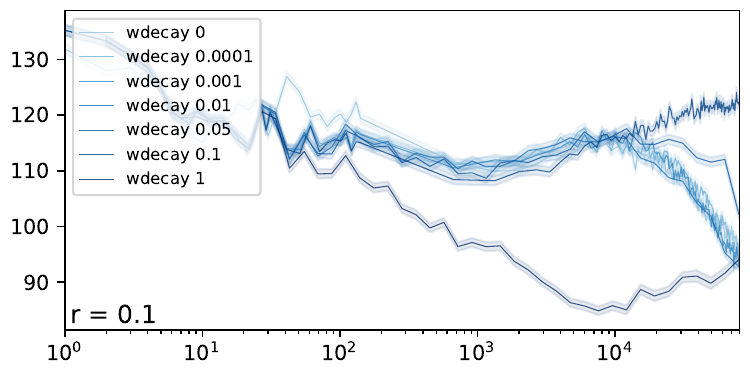}\\
    \includegraphics[width=0.33\linewidth]{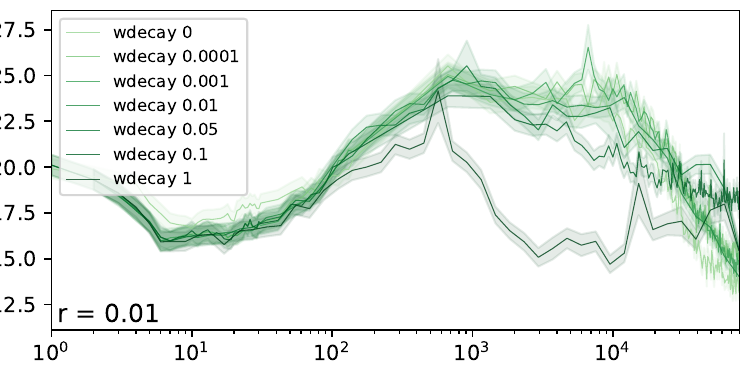}%
    \includegraphics[width=0.33\linewidth]{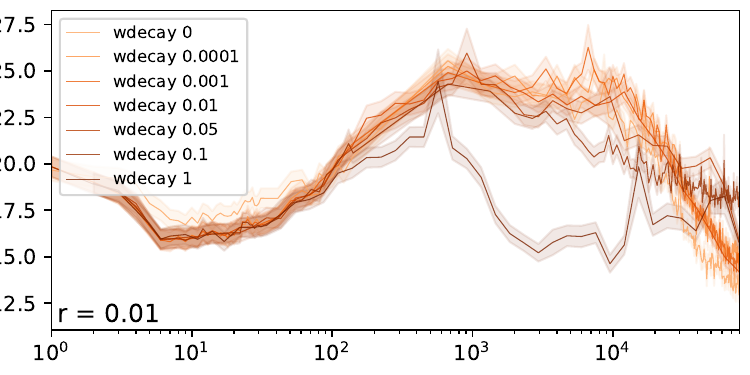}%
    \includegraphics[width=0.33\linewidth]{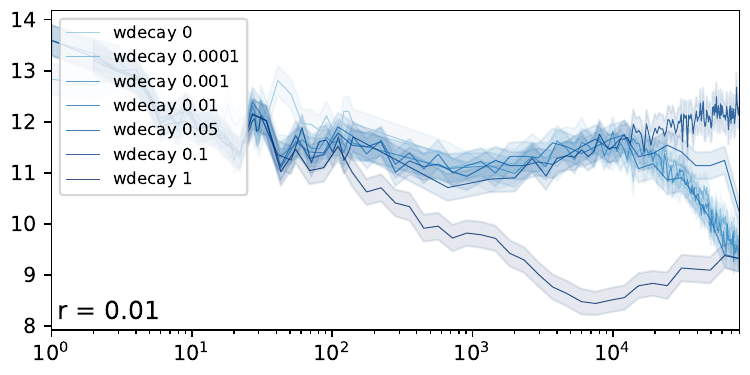}\\
    \includegraphics[width=0.33\linewidth]{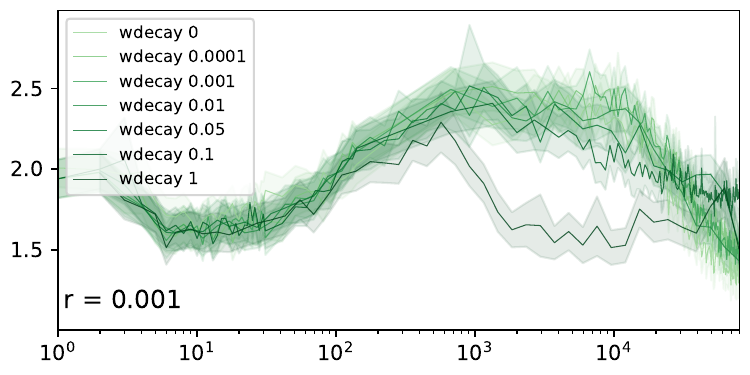}%
    \includegraphics[width=0.33\linewidth]{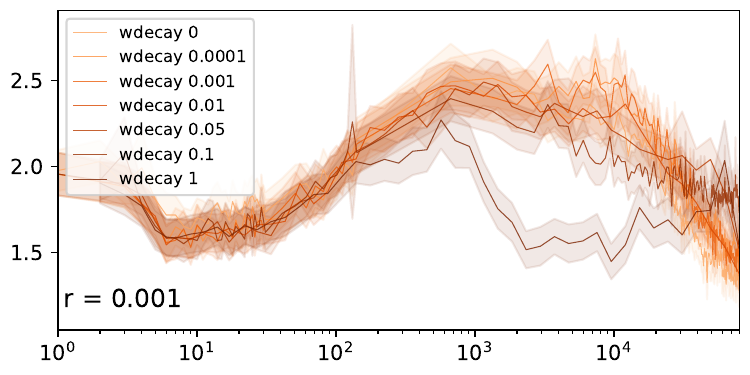}%
    \includegraphics[width=0.33\linewidth]{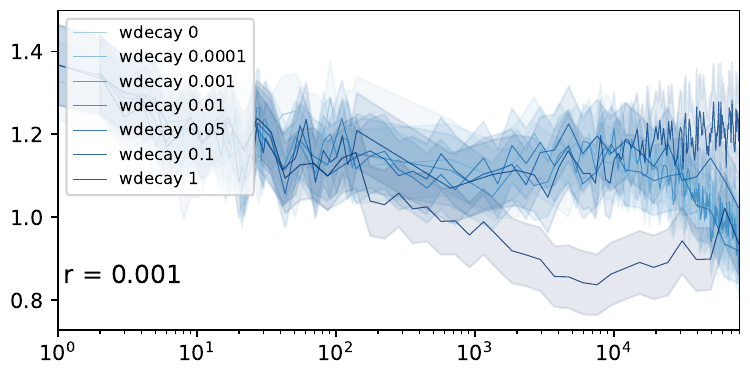}\\
\end{minipage}\\
\begin{minipage}{\linewidth}
    \centering
    \vspace{1em}
    Optimization Steps
\end{minipage}
    \caption{Local complexity dynamics training an MLP on MNIST with weight decay}
    \label{fig:mlp-weight-decay}
\end{figure}

\textbf{Weight Decay} regularizes a neural network by reducing the norm of the network weights, therefore reducing the per region slope norm as well. We train a CNN with depth 5 and width 32 and varying weight decay. In Fig.~\ref{fig:mlp-weight-decay} we present the train and random LC for our experiments. We can see that increasing weight decay also delays or removes the second descent in training LC. Moreover, strong weight decay also reduces the duration of ascent phase, as well as reduces the peak LC during ascent. This is dissimilar from BN, which removes the second descent but increases LC overall. 

\textbf{Batch Normalization}. It has previously been shown that Batch normalization (BN) regularizes training by dynamically updating the normalization parameters for every mini-batch, therefore increasing the noise in training \cite{garbin2020dropout}. In fact, we recall that BN replaces the per-layer mapping from \cref{eq:no_BN} by centering and scaling the layer's pre-activation and adding back the learnable bias $\vb^{(\ell)}$. The centering and scaling statistics are computed for each mini-batch.
After learning is complete, a final fixed ``test time'' mean $\overline{\mu}^{(\ell)}$ and standard deviation $\overline{\sigma}^{(\ell)}$ are computed using the training data. Of key interest to our observation is a result tying BN to the position in the input space of the partition region from \cite{balestriero2022batch}. In particular, it was proved that at each layer $\ell$ of a DN, BN explicitly adapts the partition so that the partition boundaries are as close to the training data as possible. This is confirmed by our experiments in Fig.~\ref{fig:cnn-cifar10-batchnorm} 
we present results for CNN
trained on CIFAR10, with and without BN. 

\section{Extra Figures}

\begin{figure}
\begin{minipage}{0.2\linewidth}
\centering
\small{Layer 1}\\
\includegraphics[width=\linewidth]{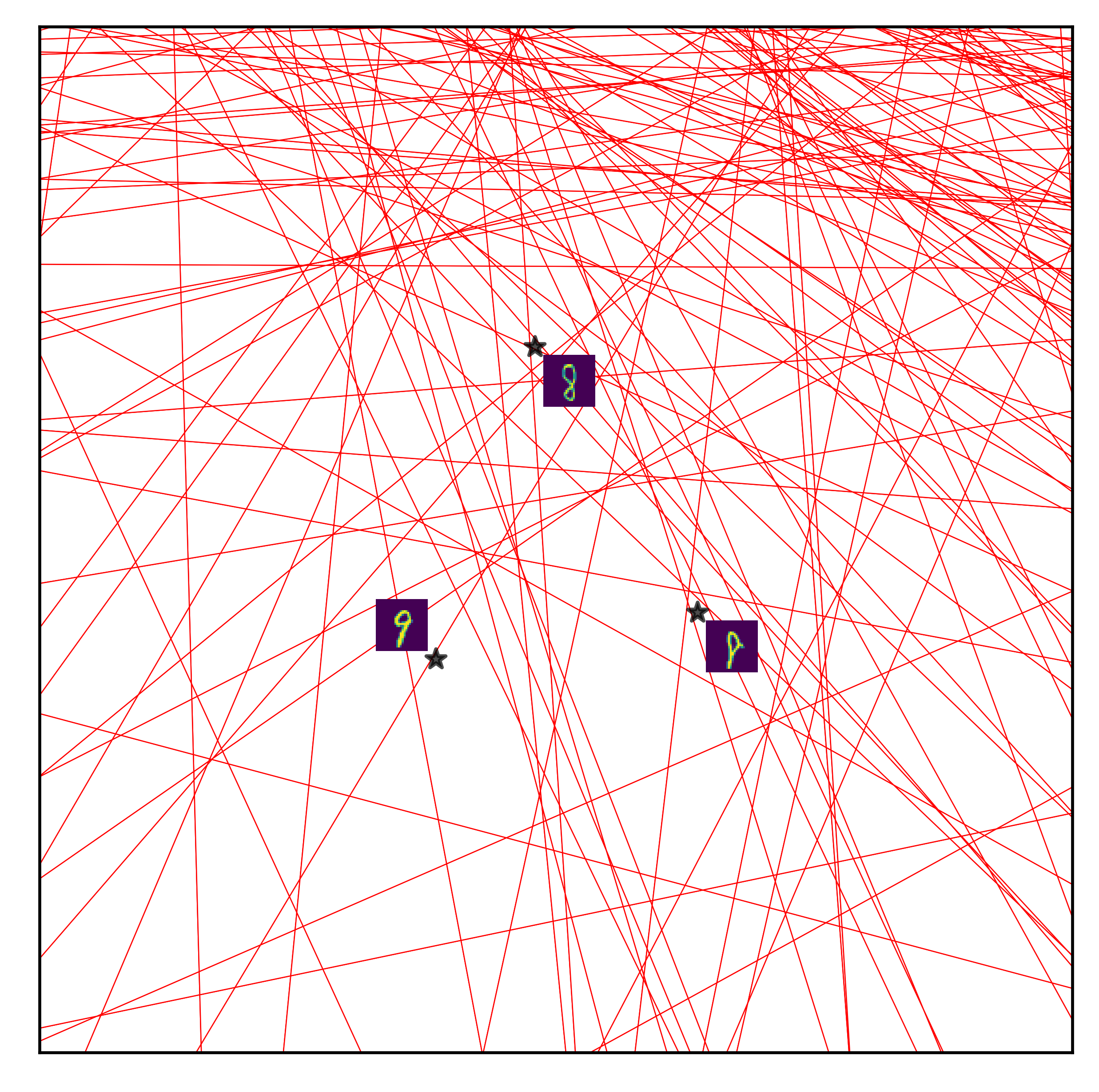}%
\end{minipage}%
\begin{minipage}{0.2\linewidth}
\centering
\small{Layer 2}\\
\includegraphics[width=\linewidth]{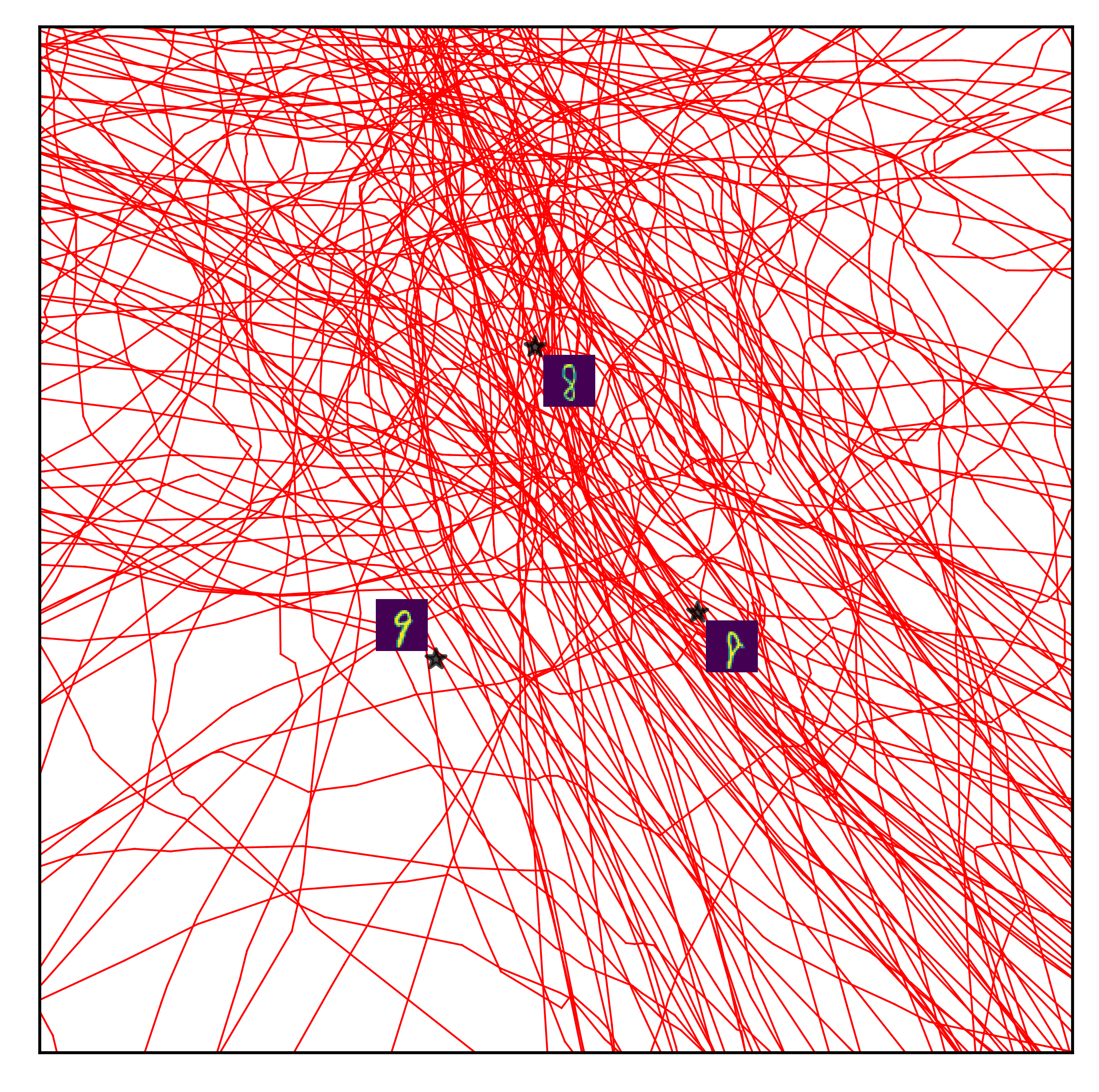}%
\end{minipage}%
\begin{minipage}{0.2\linewidth}
\centering
\small{Layer 3}\\
\includegraphics[width=\linewidth]{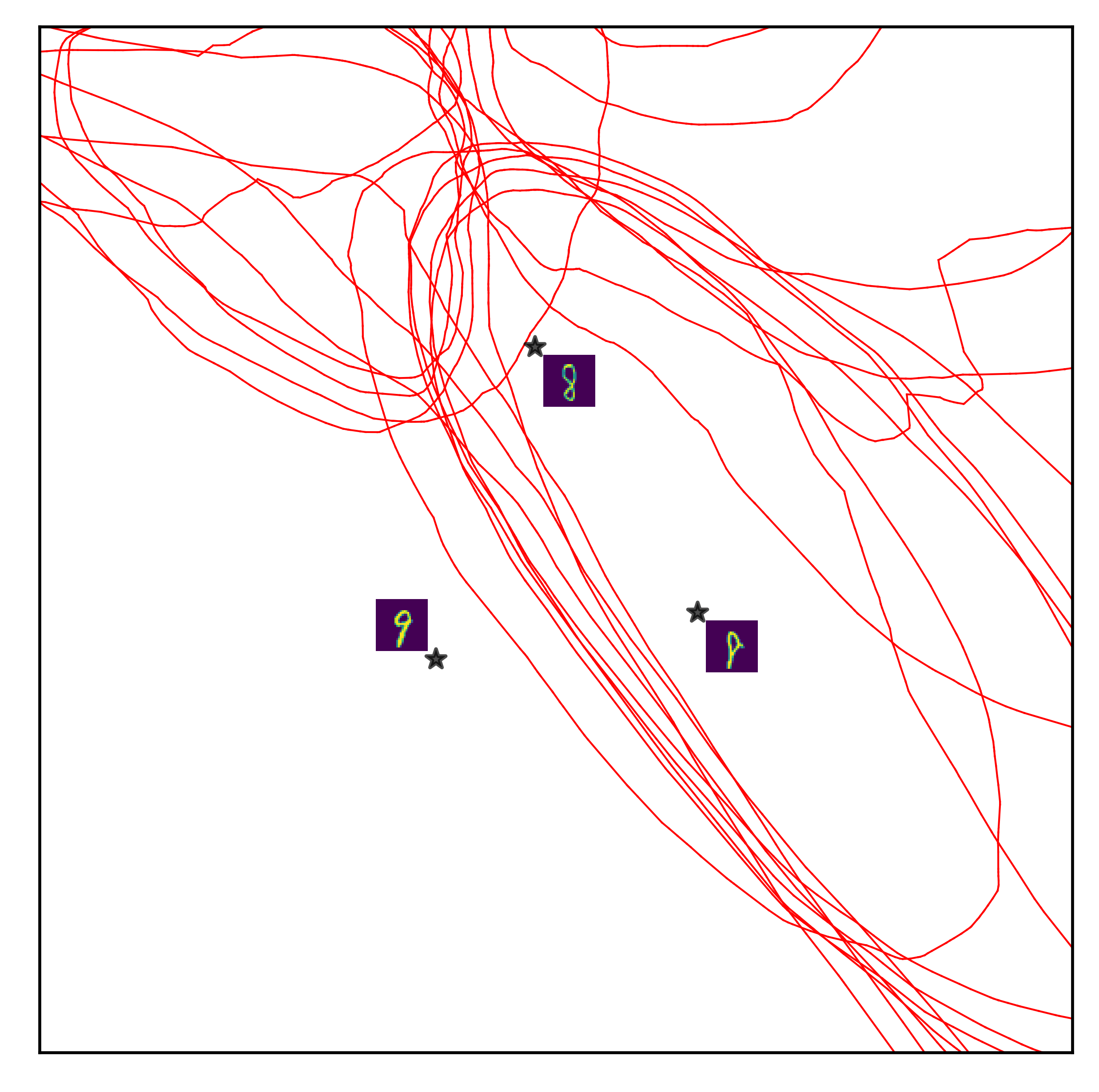}%
\end{minipage}%
\begin{minipage}{0.2\linewidth}
\centering
\small{Layer 4}\\
\includegraphics[width=\linewidth]{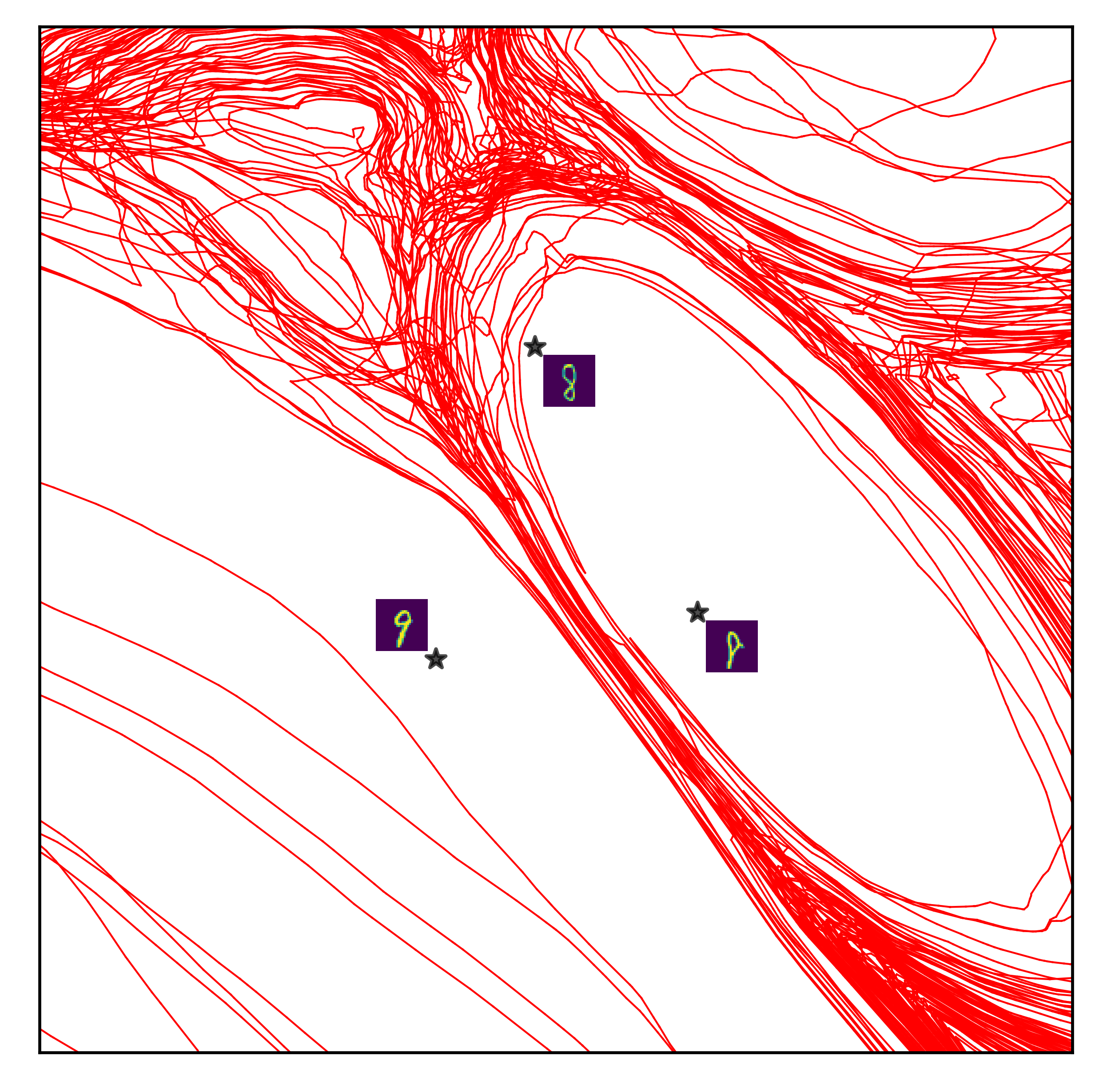}%
\end{minipage}%
\begin{minipage}{0.2\linewidth}
\centering
\small{Layer 5}\\
\includegraphics[width=\linewidth]{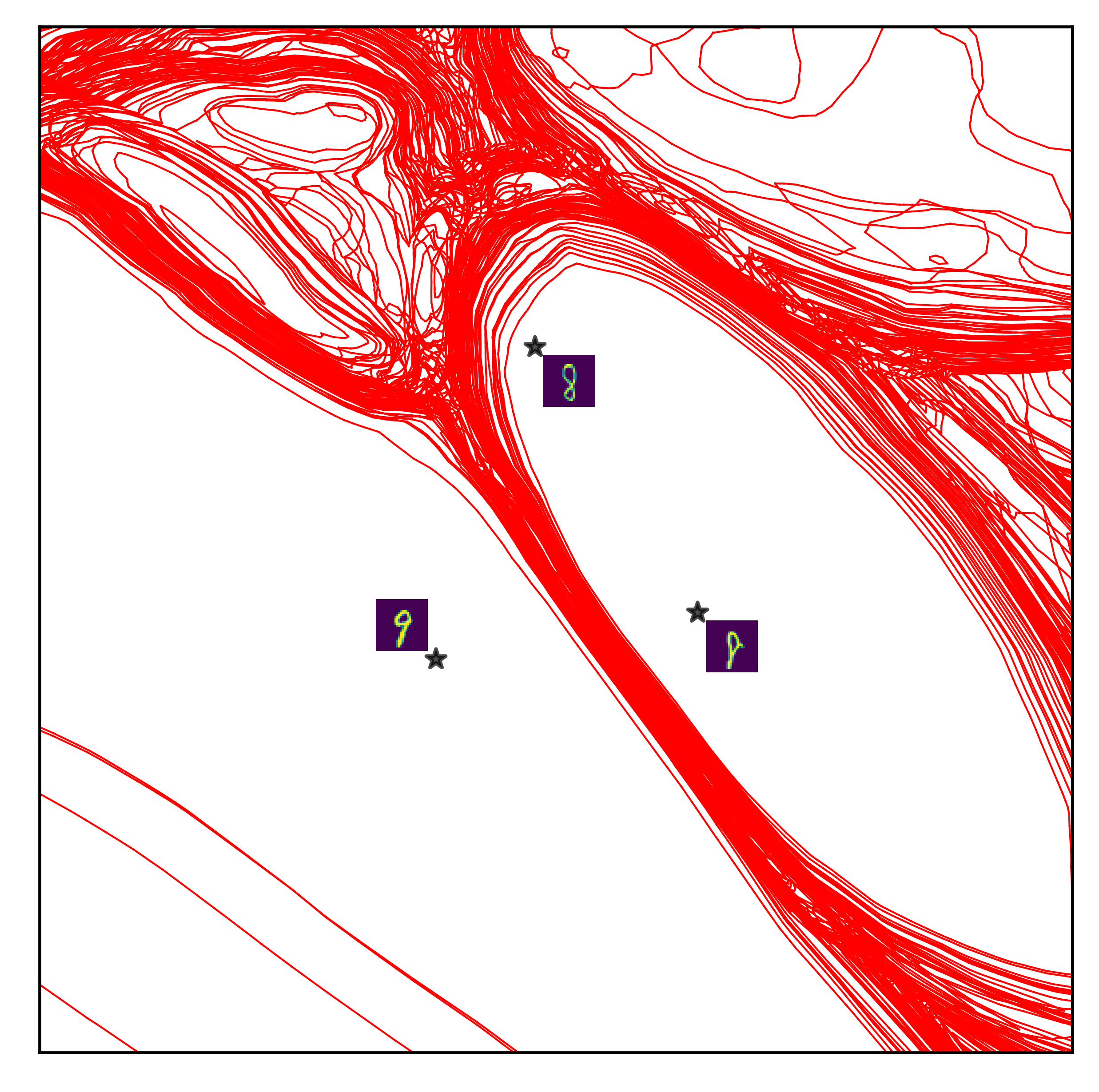}%
\end{minipage}
    \caption{Layerwise visualization of the input space partition for a 2D domain passing through a training set triad, after robust partition formation. The partition is visualized for an MLP with depth 6 and width 200, trained on $1000$ samples from MNIST, similar to the setting described in \cref{fig:mnist-splinecam}. We see that deeper layer neurons partake more in the formation of the robust partition, compared to shallower layers. This is due to the fact that deeper layer neurons can be more localized in the input space due to the non-linearity induced by preceding layers.}
    \label{fig:layerwise partition}
\end{figure}

\begin{figure}
\begin{minipage}{0.19\linewidth}
\centering
\small{Input Partition}\\
\includegraphics[width=\linewidth]{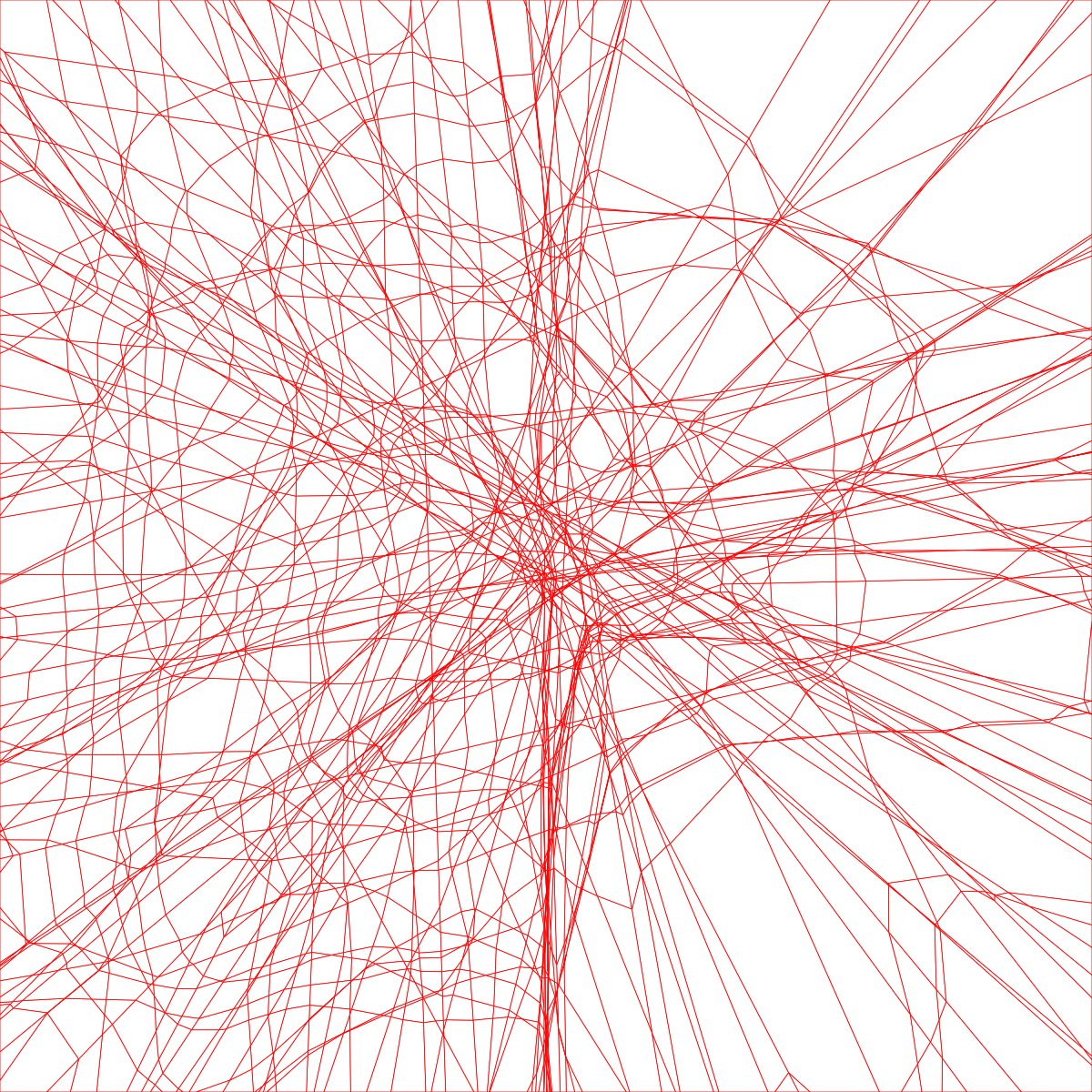}%
\end{minipage}%
\begin{minipage}{0.01\linewidth}
\centering
\hspace{\linewidth}
\end{minipage}%
\begin{minipage}{0.2\linewidth}
\centering
\small{LC, $r=0.05$}\\
\includegraphics[width=\linewidth]{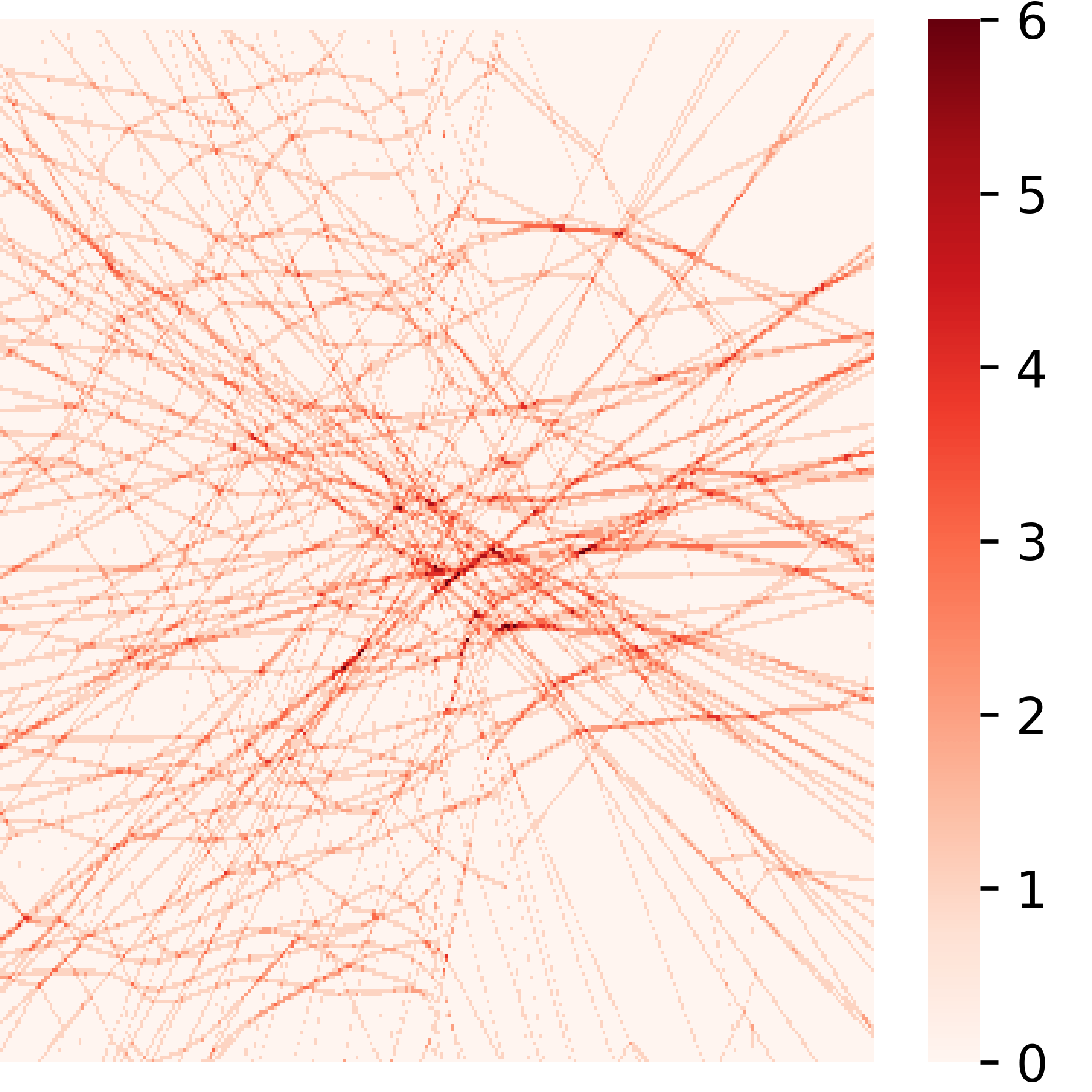}%
\end{minipage}%
\begin{minipage}{0.2\linewidth}
\centering
\small{LC, $r=0.1$}\\
\includegraphics[width=\linewidth]{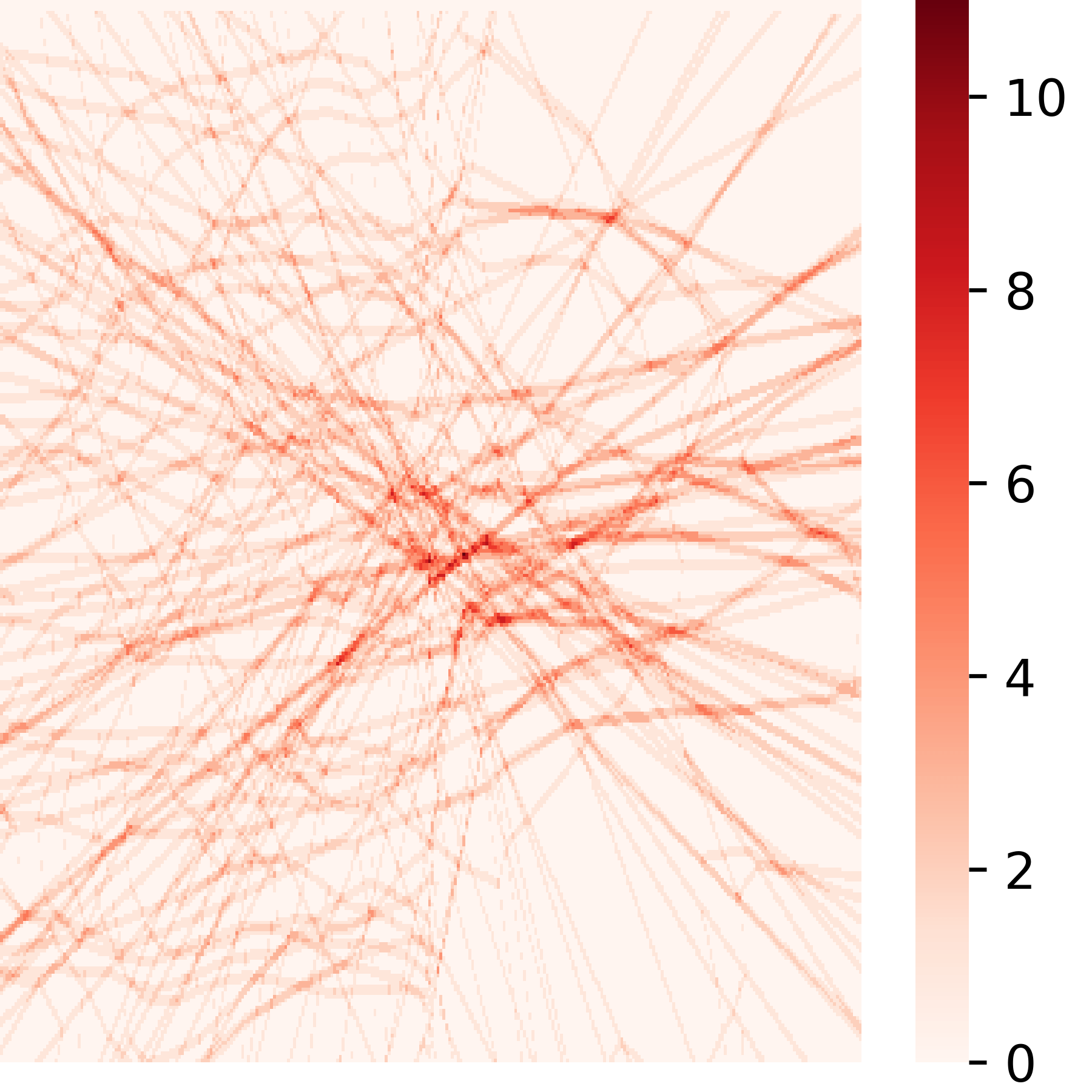}%
\end{minipage}%
\begin{minipage}{0.2\linewidth}
\centering
\small{LC, $r=0.5$}\\
\includegraphics[width=\linewidth]{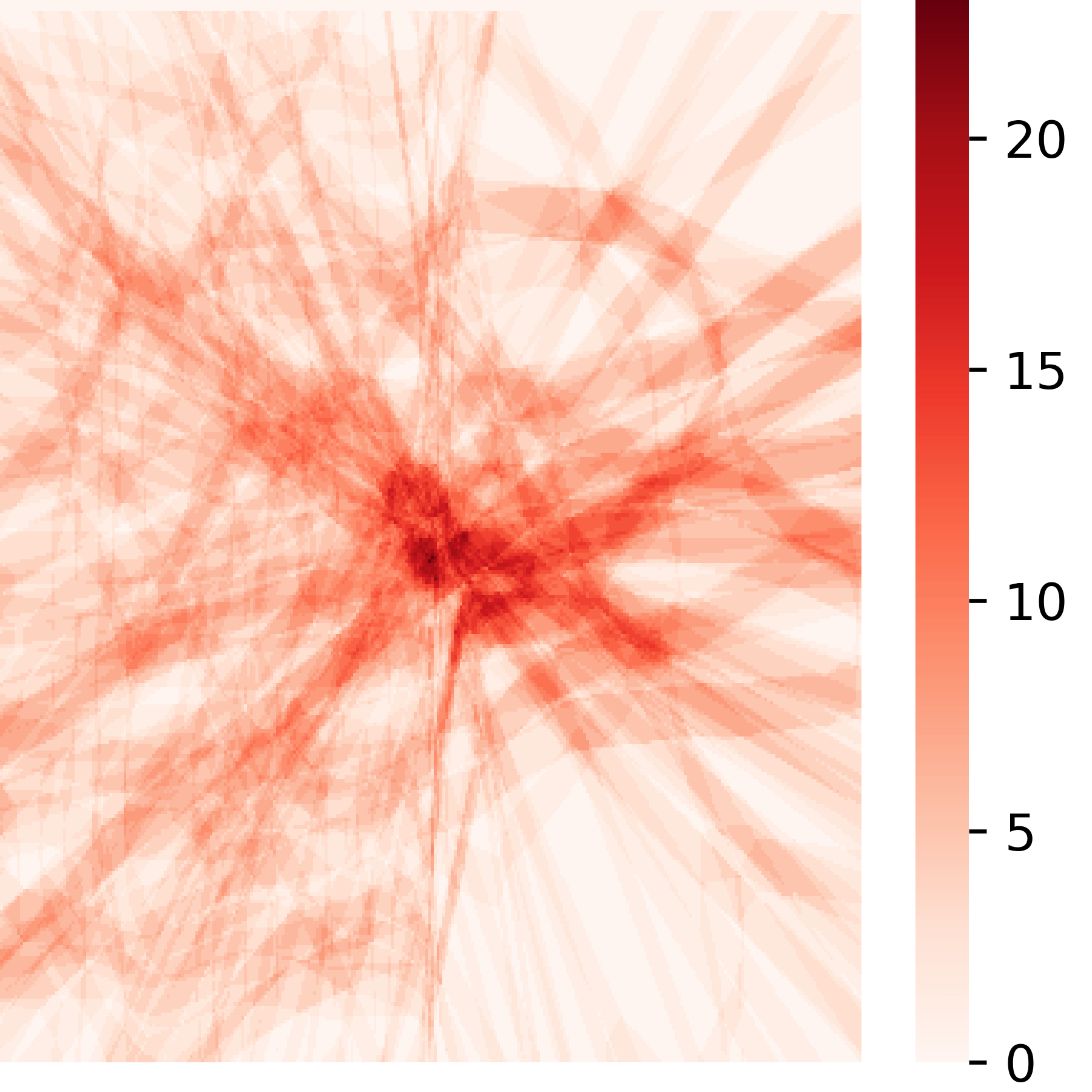}%
\end{minipage}%
\begin{minipage}{0.2\linewidth}
\centering
\small{LC, $r=1$}\\
\includegraphics[width=\linewidth]{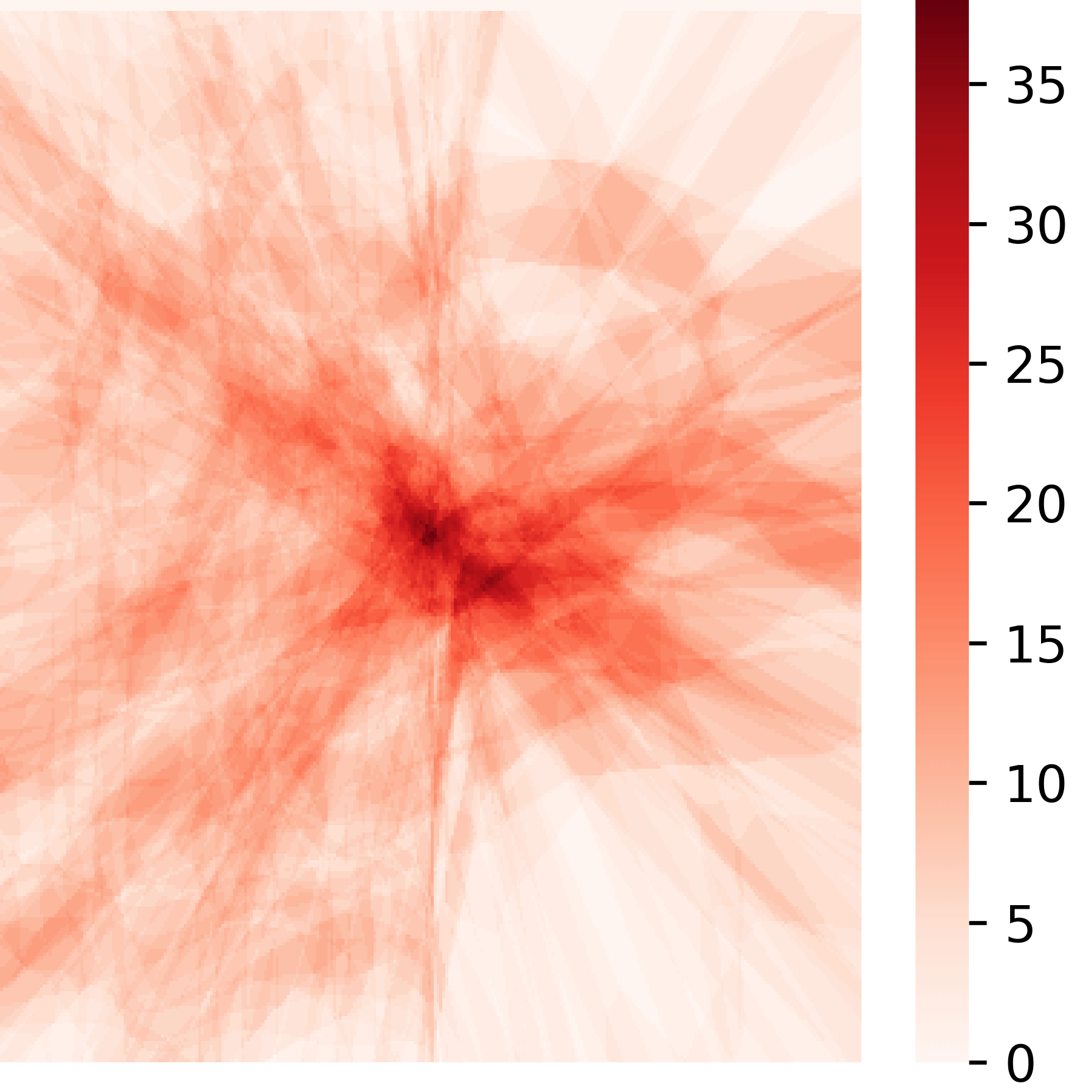}%
\end{minipage}
    \caption{Input space partition computed analytically via SplineCam \cite{humayun2023splinecam} for the 2D toy setting presented in \cref{fig:2dexact} (left). Regions are colored by white and knots  are colored by red. The partition is computed for the input space domain $[-10,10]^2$, induced by an MLP of depth $5$ and width $30$. We take a meshgrid of $300\times300$ points over the input domain, and measure the local complexity at each point with radius, $r \in \{0.05,0.1,0.5,1\}$ (rest). We see that our proposed method can locate the non-linearities for small $r$. As $r$ is increased our method provides a coarser estimate of the local density of non-linearities, i.e., number of non-linearities intersecting the a fixed volume defined by the local neighborhood. 
    }
    \label{fig:2D_lc_partition}
\end{figure}


\begin{figure}
    \centering
    \includegraphics[width=.5\linewidth]{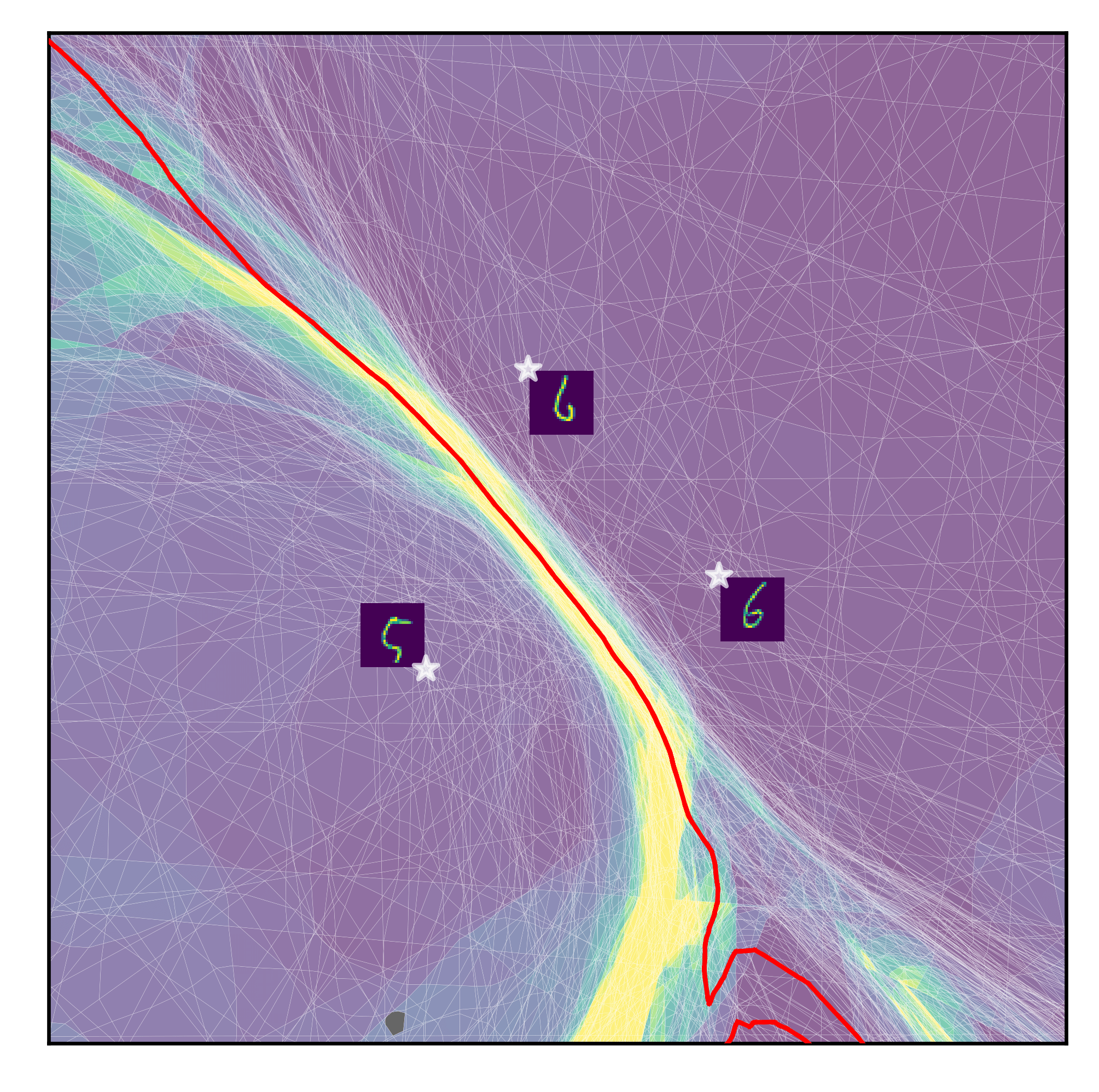}%
    \includegraphics[width=.5\linewidth]{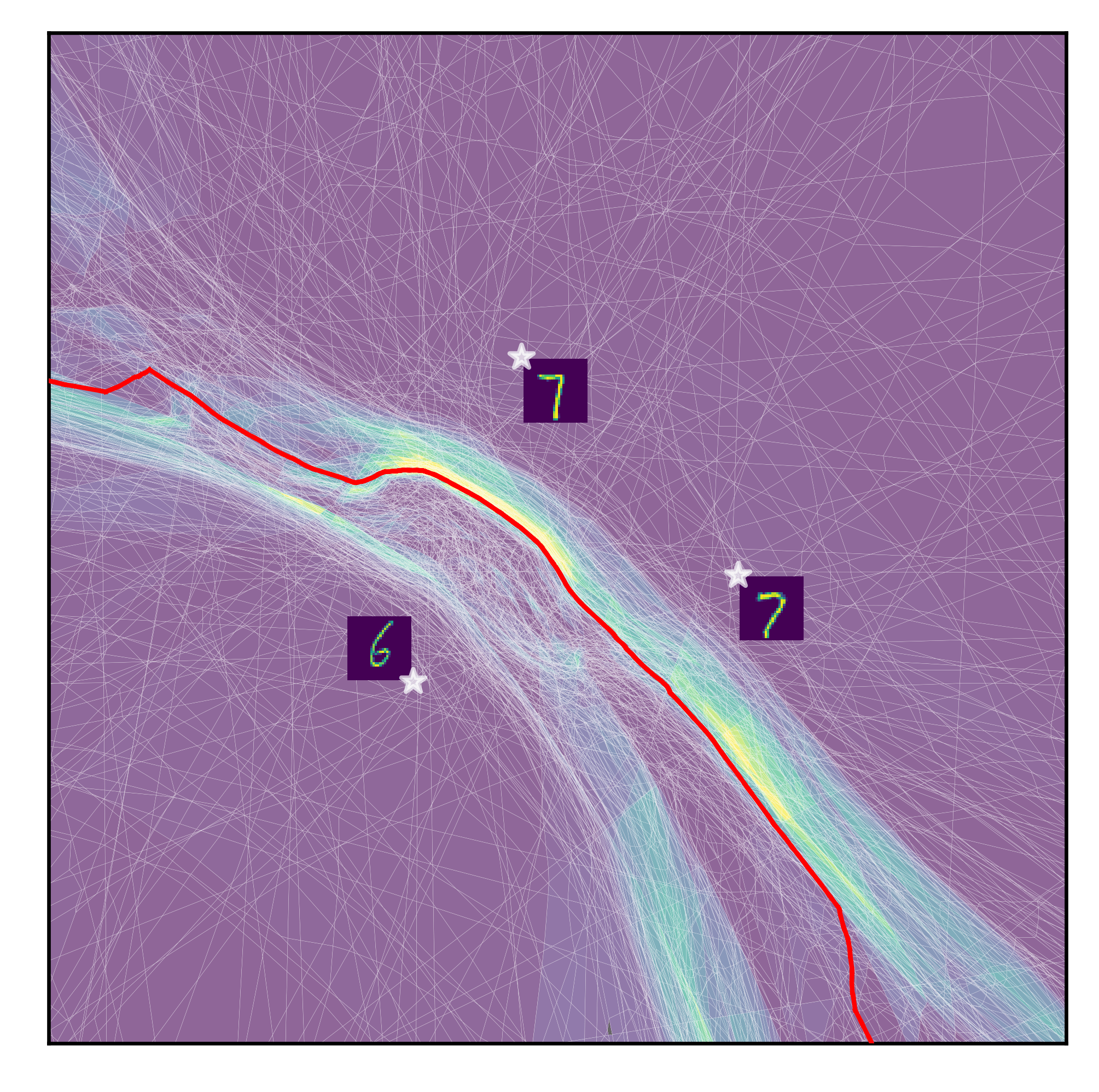}\\
    \includegraphics[width=.5\linewidth]{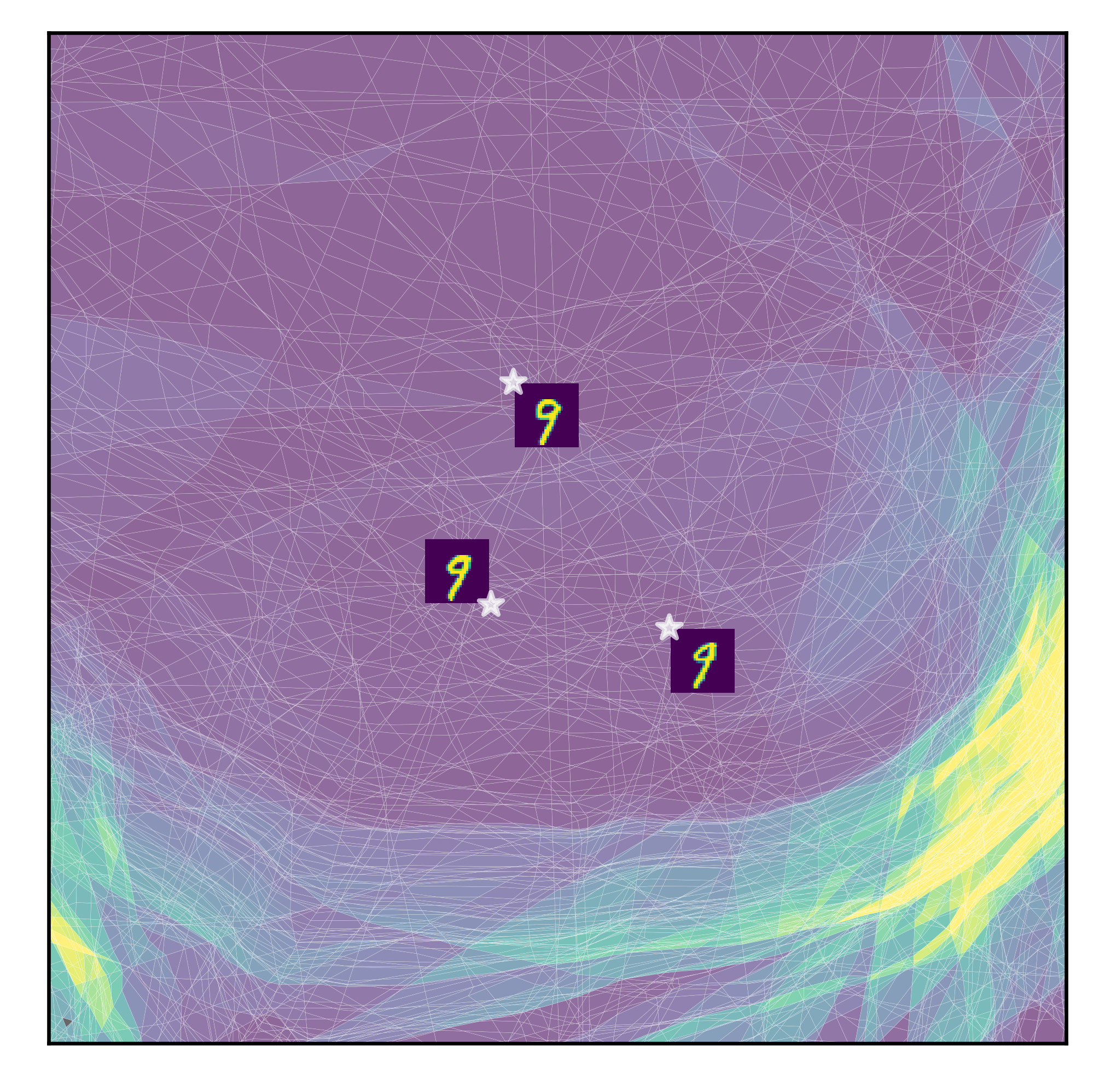}%
    \includegraphics[width=.5\linewidth]{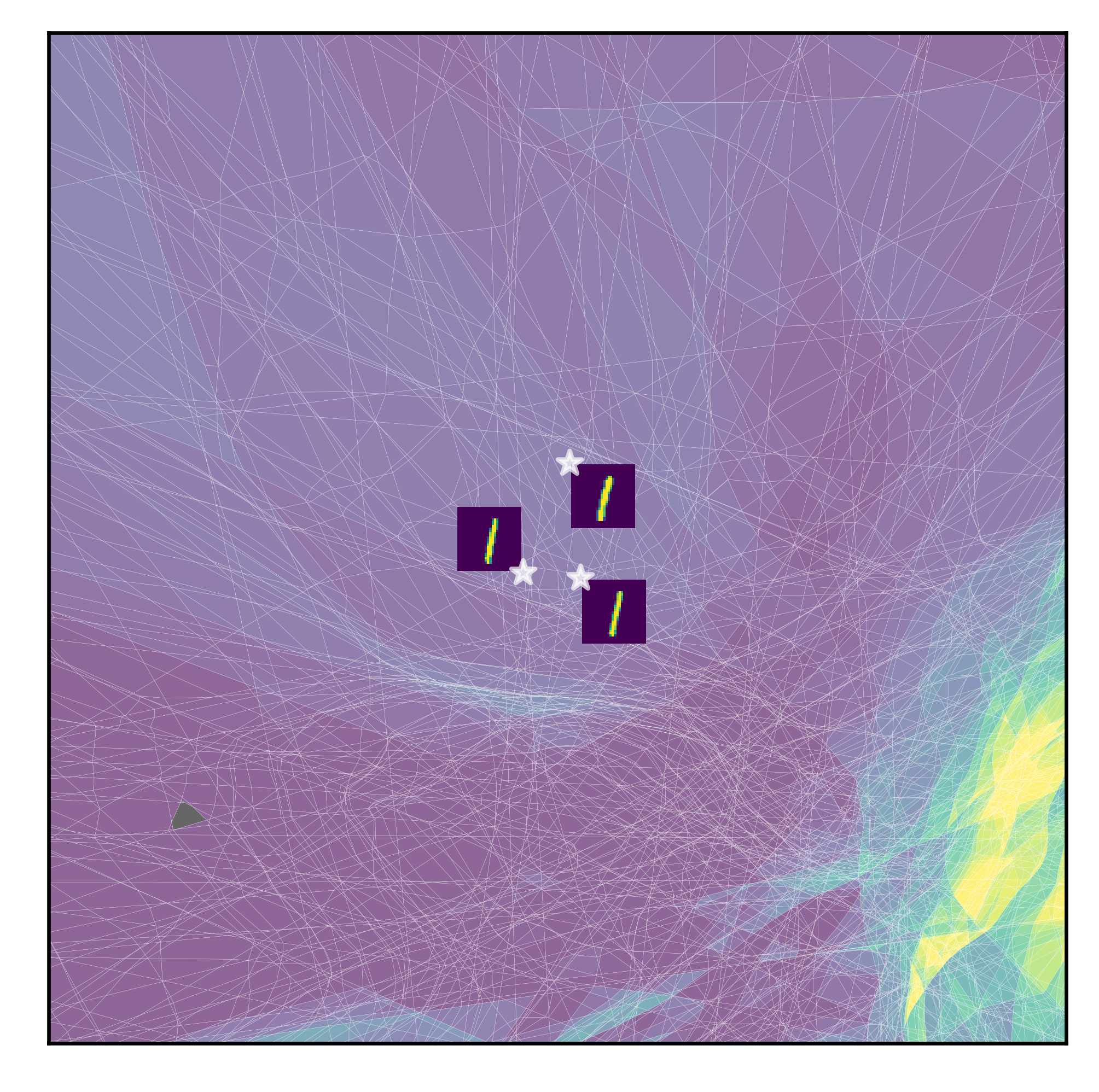}%
    \caption{Partition visualization for 2D domains localized around the decision boundary (top) and away from the decision boundary (bottom) for the grokking setup presented in \cref{fig:grokking-splinecam}. All the plots are show for the optimization step $95381$. Number of regions in the partition for top-right, top-left, bottom-right, and bottom-left are 123156, 88362, 33273, and 32018 respectively. The domain used for all of the plots has the same area/volume. Therefore, close to the decision boundary, the region density is much higher compared to away from the decision boundary. This is evidence of region migration happening during the latter phases of training.}
    \label{fig:splinecam-grok-inandoutclass}
\end{figure}

\begin{figure}
\begin{minipage}{0.03\linewidth}
\centering
    \rotatebox[]{90}{{\small{Accuracy}}}
\end{minipage}%
\begin{minipage}{.94\textwidth}
    \centering
    \includegraphics[width=0.2\linewidth]{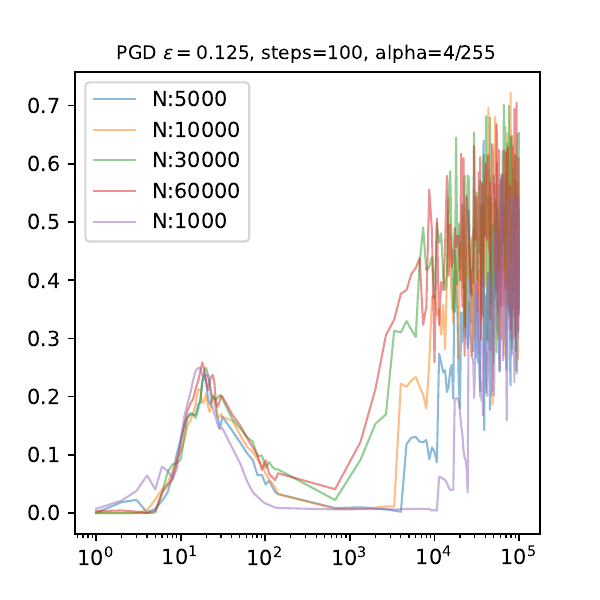}%
    \includegraphics[width=0.2\linewidth]{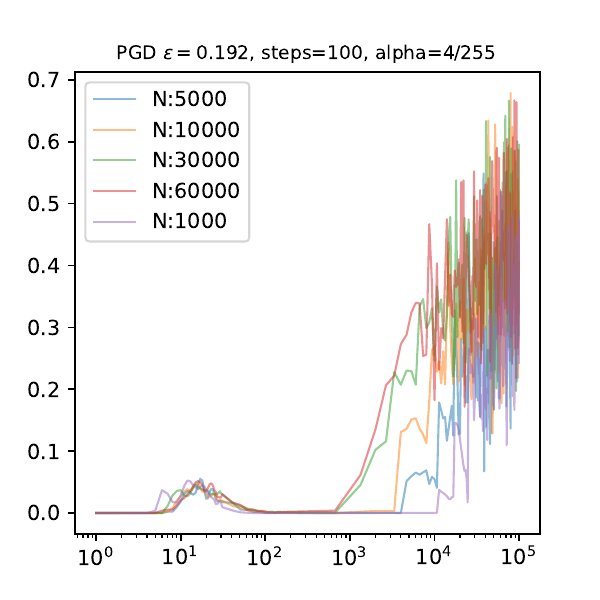}%
    \includegraphics[width=0.2\linewidth]{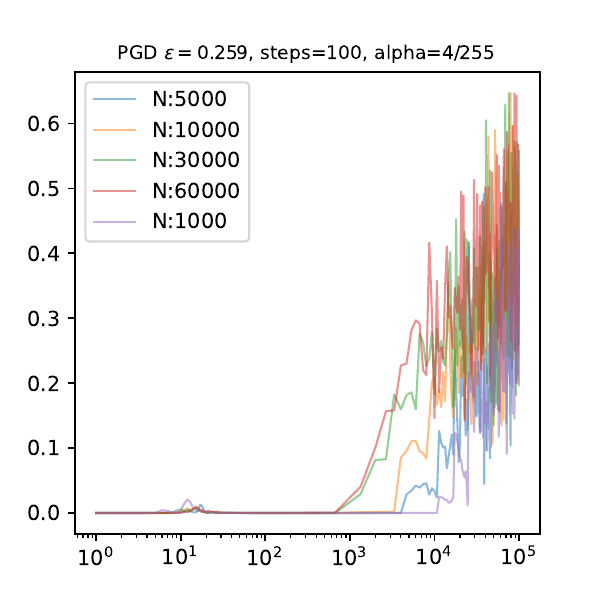}%
    \includegraphics[width=0.2\linewidth]{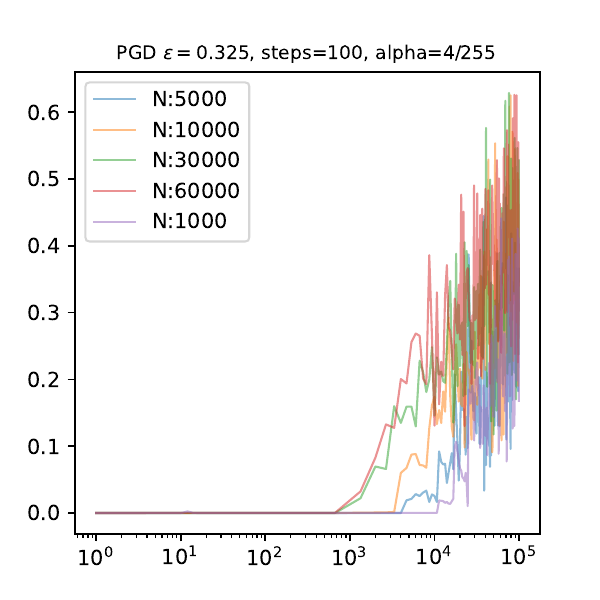}%
    \includegraphics[width=0.2\linewidth]{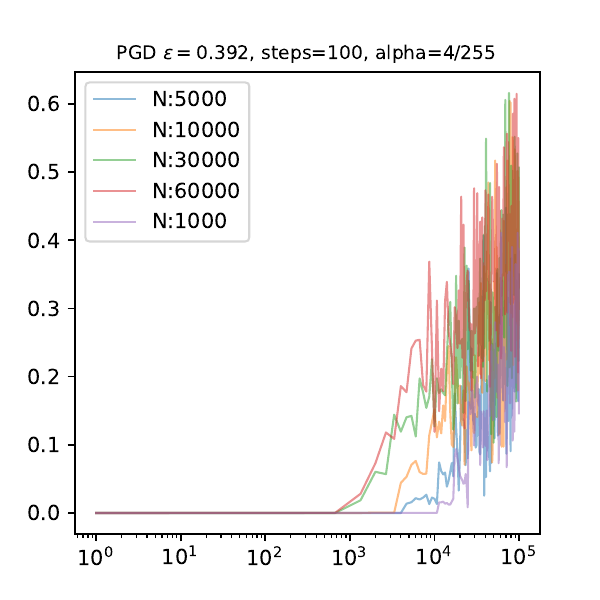}%
\end{minipage}\\
\begin{minipage}{\linewidth}
    \centering
    \vspace{1em}
    Optimization Steps
\end{minipage}
    \caption{For an MLP with depth 4 and width 200, we train with varying training set sizes and evaluate the adversarial performance after each training iteration. We see that with increasing dataset size, the network groks earlier in time, as can be visible in the adversarial grokking curves for all the different epsilon values.}
    \label{fig:advgrok-mnist-datasweep}
\end{figure}

\begin{figure}
\begin{minipage}{0.03\linewidth}
\centering
    \rotatebox[]{90}{{\small{Local Complexity}}}
\end{minipage}%
\begin{minipage}{.94\textwidth}
    \centering
    \includegraphics[width=0.2\linewidth]{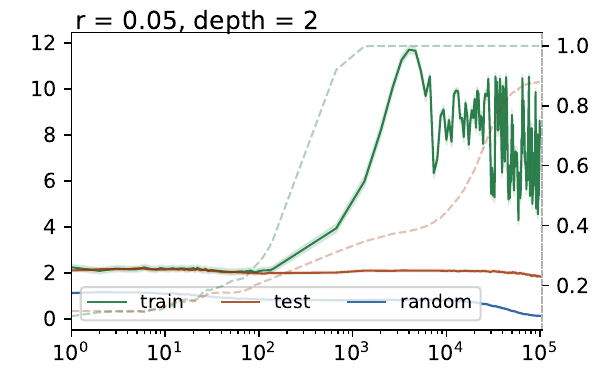}%
    \includegraphics[width=0.2\linewidth]{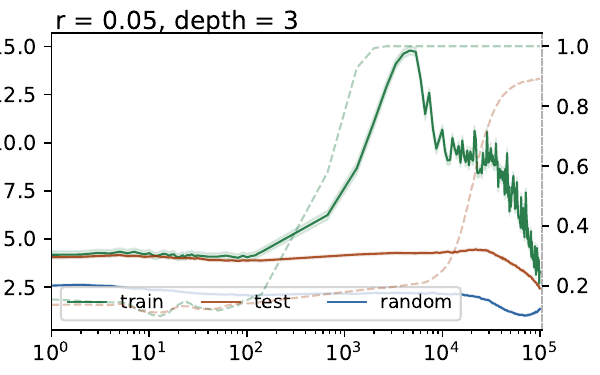}%
    \includegraphics[width=0.2\linewidth]{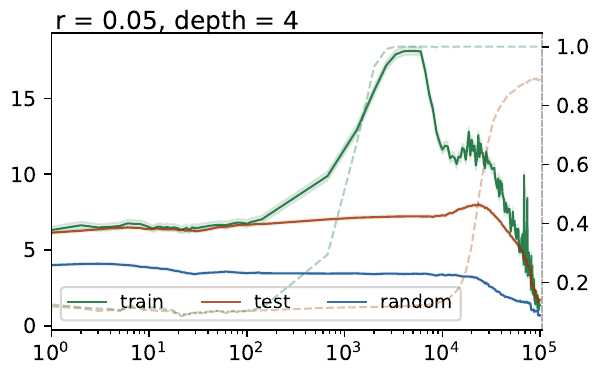}%
    \includegraphics[width=0.2\linewidth]{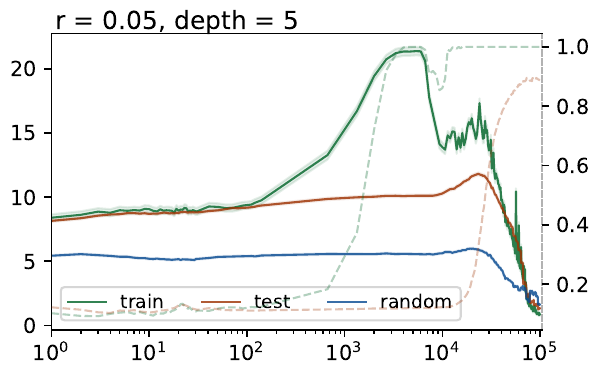}%
    \includegraphics[width=0.2\linewidth]{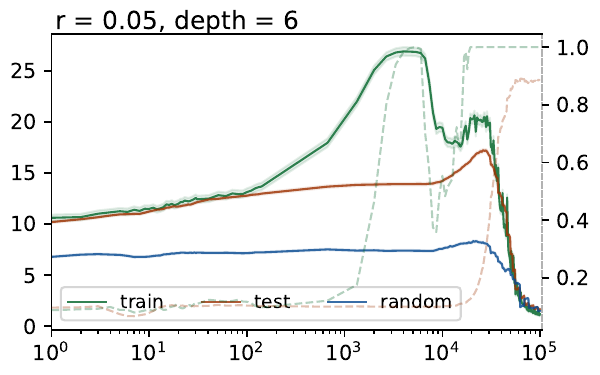}\\
    \includegraphics[width=0.2\linewidth]{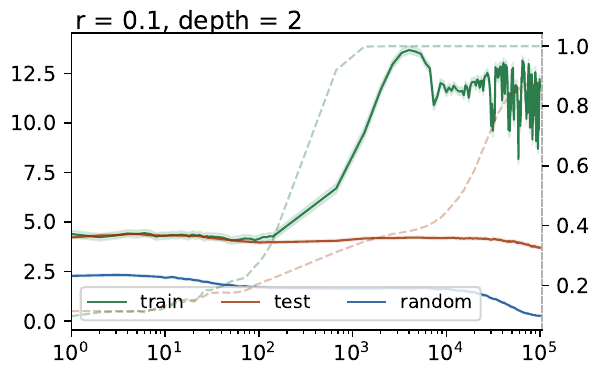}%
    \includegraphics[width=0.2\linewidth]{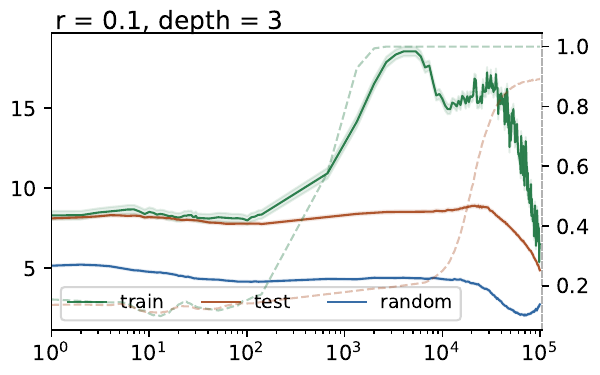}%
    \includegraphics[width=0.2\linewidth]{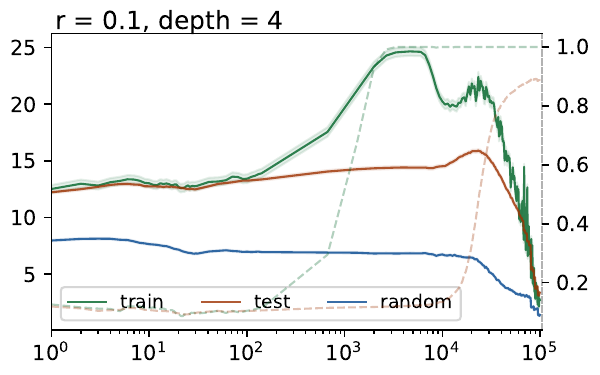}%
    \includegraphics[width=0.2\linewidth]{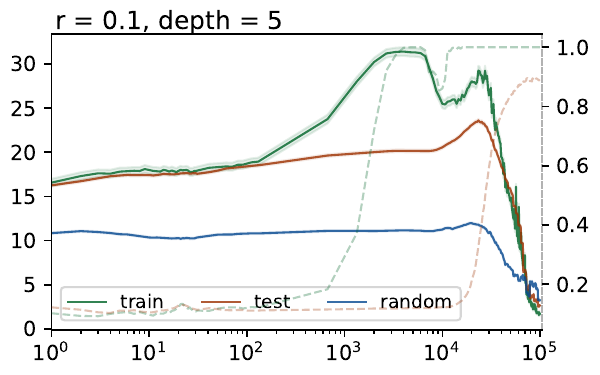}%
    \includegraphics[width=0.2\linewidth]{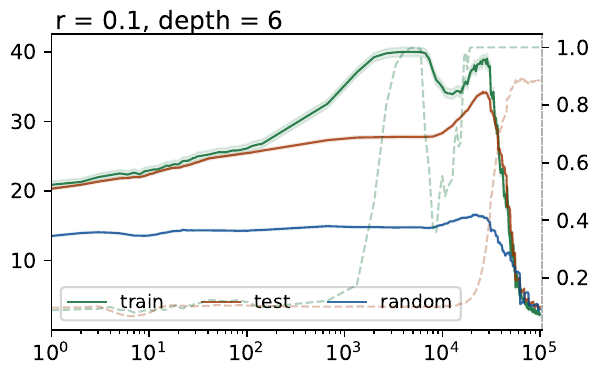}\\
    \includegraphics[width=0.2\linewidth]{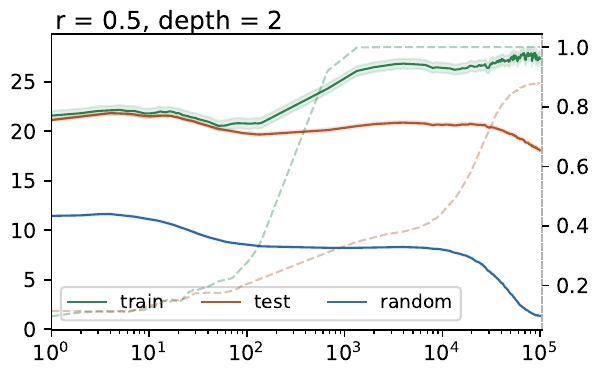}%
    \includegraphics[width=0.2\linewidth]{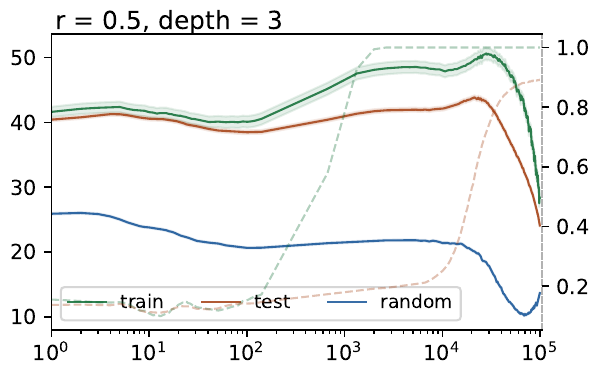}%
    \includegraphics[width=0.2\linewidth]{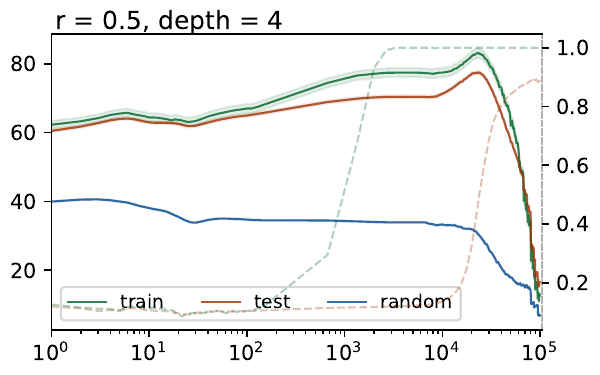}%
    \includegraphics[width=0.2\linewidth]{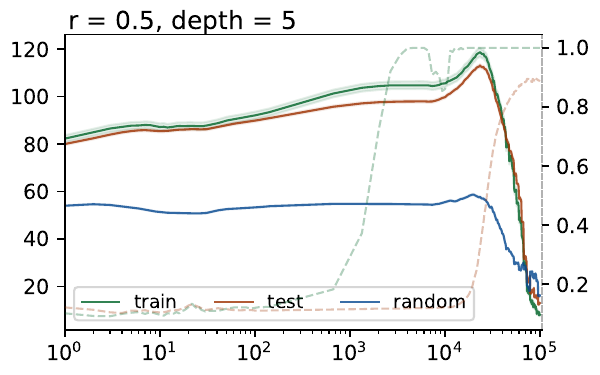}%
    \includegraphics[width=0.2\linewidth]{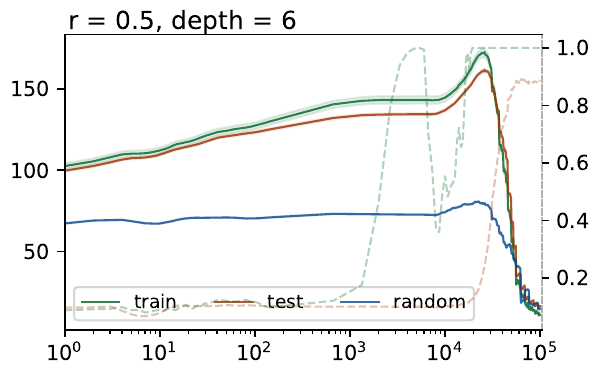}\\
    \includegraphics[width=0.2\linewidth]{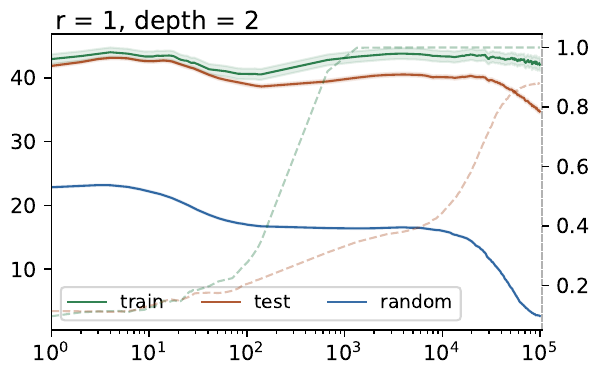}%
    \includegraphics[width=0.2\linewidth]{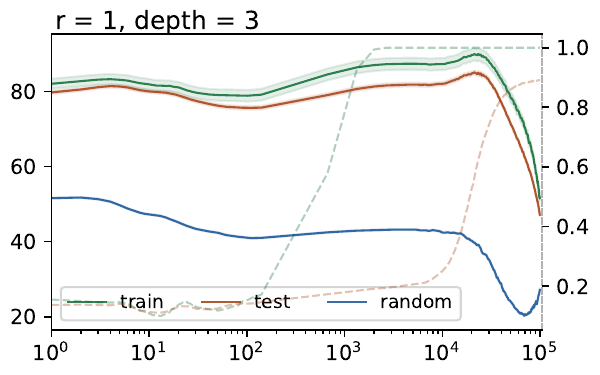}%
    \includegraphics[width=0.2\linewidth]{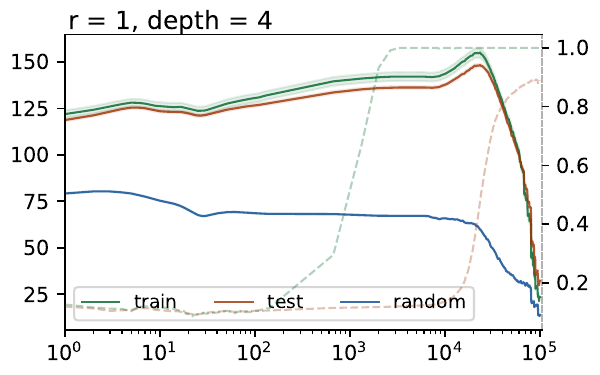}%
    \includegraphics[width=0.2\linewidth]{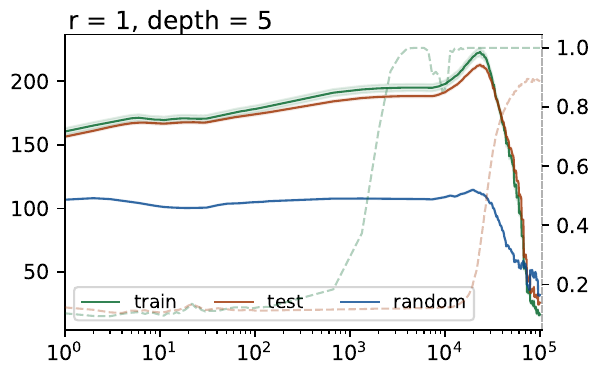}%
    \includegraphics[width=0.2\linewidth]{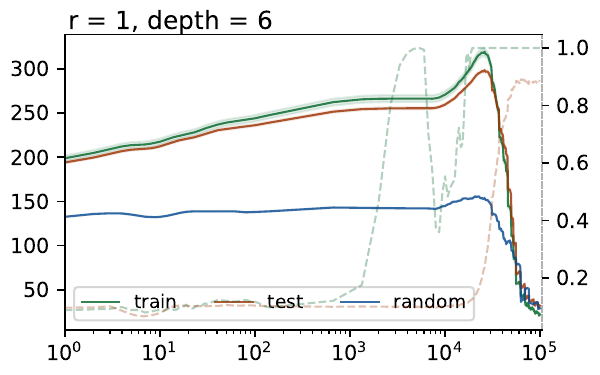}\\  \includegraphics[width=0.2\linewidth]{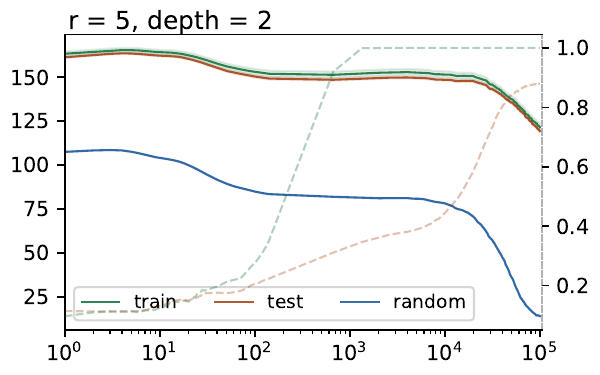}%
    \includegraphics[width=0.2\linewidth]{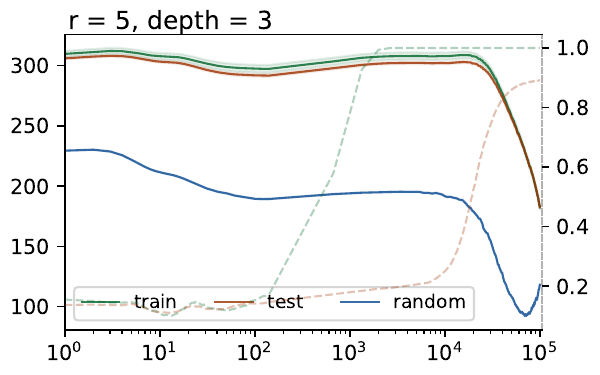}%
    \includegraphics[width=0.2\linewidth]{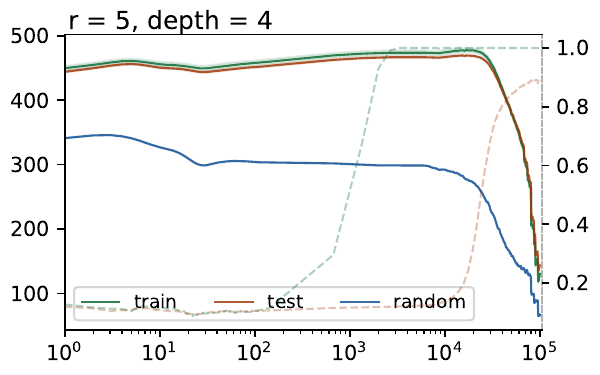}%
    \includegraphics[width=0.2\linewidth]{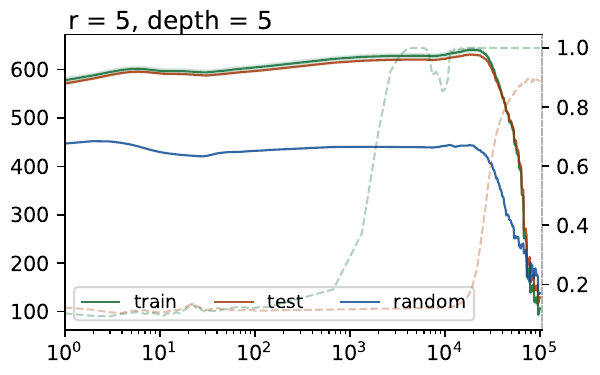}%
    \includegraphics[width=0.2\linewidth]{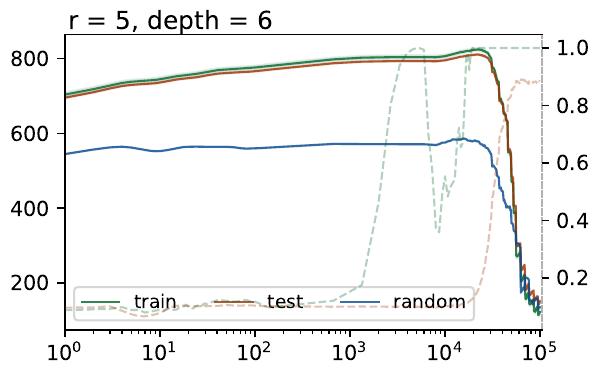}%
\end{minipage}%
\begin{minipage}{0.03\linewidth}
\centering
    \rotatebox[]{270}{{\small{Accuracy}}}
\end{minipage}\\
\begin{minipage}{\linewidth}
    \centering
    \vspace{1em}
    Optimization Steps
\end{minipage}
    \caption{Training a 200 width MLP on MNIST with initialization scaling of 8 and varying depths. Along the row, we consider larger and larger radius neighborhoods for local complexity approximation.}
    \label{fig:grokking-init8-mlp-depth-sweep}
\end{figure}

\begin{figure}
\begin{minipage}{\linewidth}
    \centering
    \hspace{3em} Training Samples \hspace{8em} Test Samples \hspace{10em} Random Samples
\end{minipage}\\
\begin{minipage}{0.03\linewidth}
\centering
    \rotatebox[]{90}{{\small{Local Complexity}}}
\end{minipage}%
\begin{minipage}{.97\textwidth}
    \centering
    \includegraphics[width=0.33\linewidth]{figures/mlp_rand_datswp_train_r0.05.pdf}%
    \includegraphics[width=0.33\linewidth]{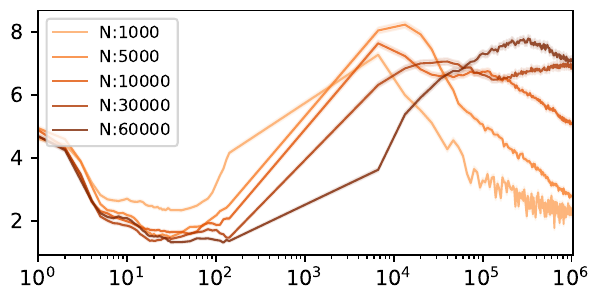}%
    \includegraphics[width=0.33\linewidth]{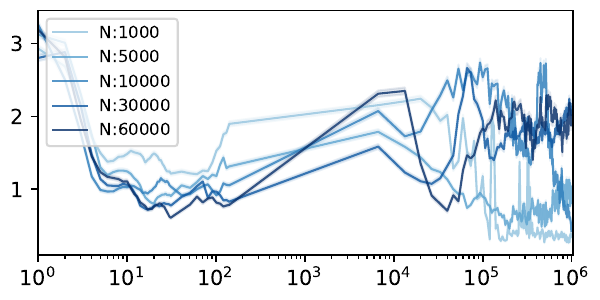}%
\end{minipage}\\
\begin{minipage}{\linewidth}
    \centering
    \vspace{1em}
    Optimization Steps
\end{minipage}
    \caption{\textbf{Increasing the volume of randomly labeled training data.} Continued from \cref{fig:mlp_rand_dataswp}. Increasing the number of randomly labeled training samples delays the ascent phase of the LC training dynamics for both training and test samples. For random samples the behavior is not affected as much.}
    \label{fig:mlp_rand_dataswp_all}
\end{figure}

\begin{figure}
\begin{minipage}{\linewidth}
    \centering
    \hspace{3em} Training Samples \hspace{8em} Test Samples \hspace{10em} Random Samples
\end{minipage}\\
\begin{minipage}{0.03\linewidth}
\centering
    \rotatebox[]{90}{{\small{Local Complexity}}}
\end{minipage}%
\begin{minipage}{.97\textwidth}
    \centering
    \includegraphics[width=0.33\linewidth]{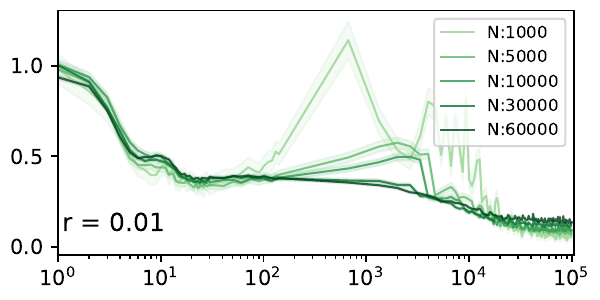}%
    \includegraphics[width=0.33\linewidth]{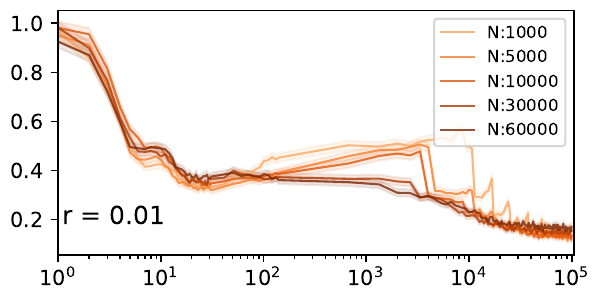}%
    \includegraphics[width=0.33\linewidth]{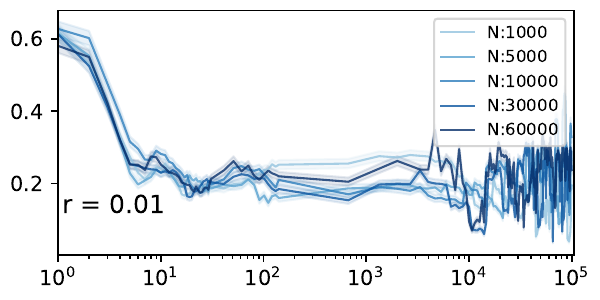}\\
    \includegraphics[width=0.33\linewidth]{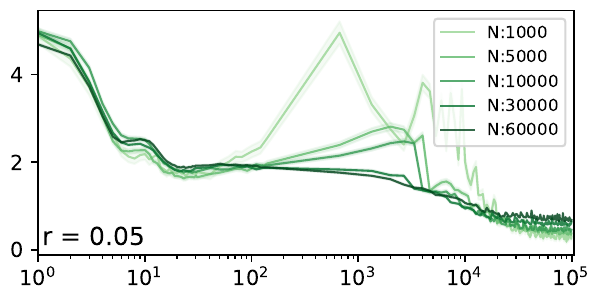}%
    \includegraphics[width=0.33\linewidth]{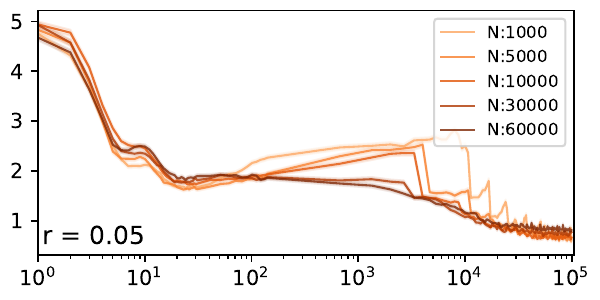}%
    \includegraphics[width=0.33\linewidth]{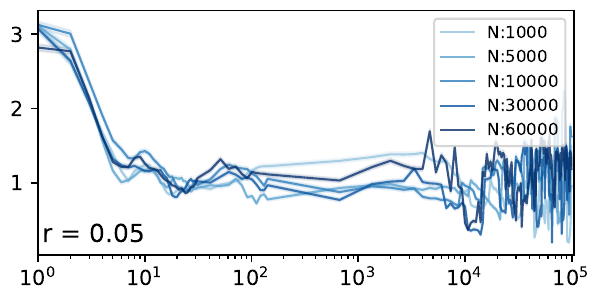}\\
    \includegraphics[width=0.33\linewidth]{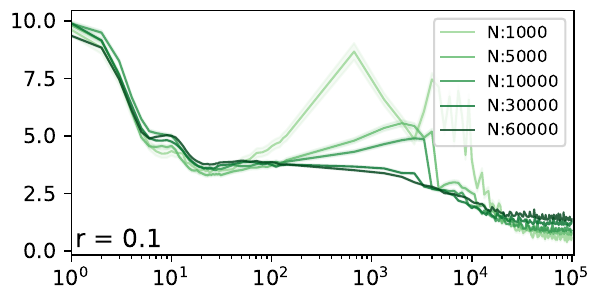}%
    \includegraphics[width=0.33\linewidth]{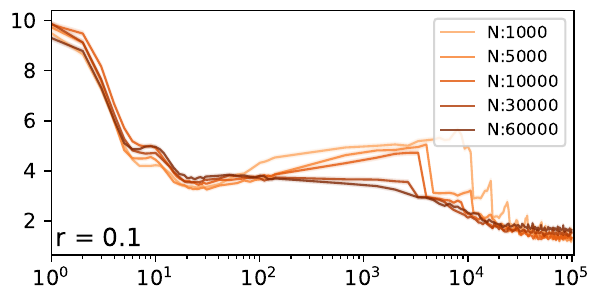}%
    \includegraphics[width=0.33\linewidth]{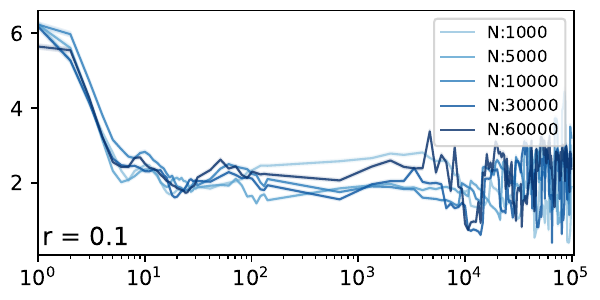}\\
    \includegraphics[width=0.33\linewidth]{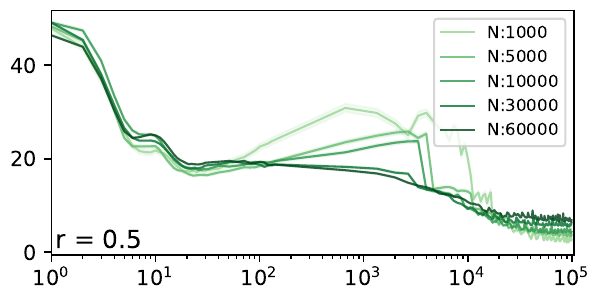}%
    \includegraphics[width=0.33\linewidth]{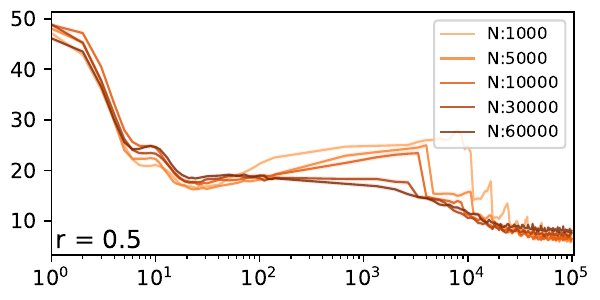}%
    \includegraphics[width=0.33\linewidth]{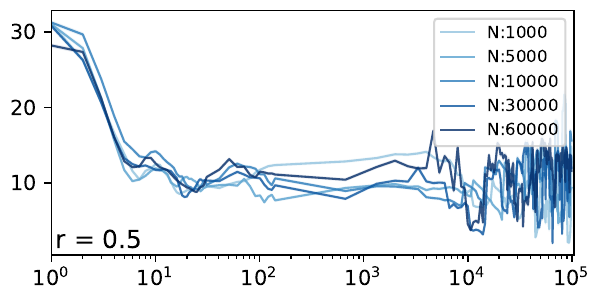}\\
\end{minipage}\\
\begin{minipage}{\linewidth}
    \centering
    \vspace{1em}
    Optimization Steps
\end{minipage}
    \caption{\textbf{Increasing training data size expedites region migration.} Local complexity dynamics training an MLP on MNIST with weight decay. Robustness plots presented in \cref{fig:advgrok-mnist-datasweep}.}
    \label{fig:mlp-dataswp}
\end{figure}

\begin{figure}
    \centering
    \includegraphics[width=.4\linewidth]{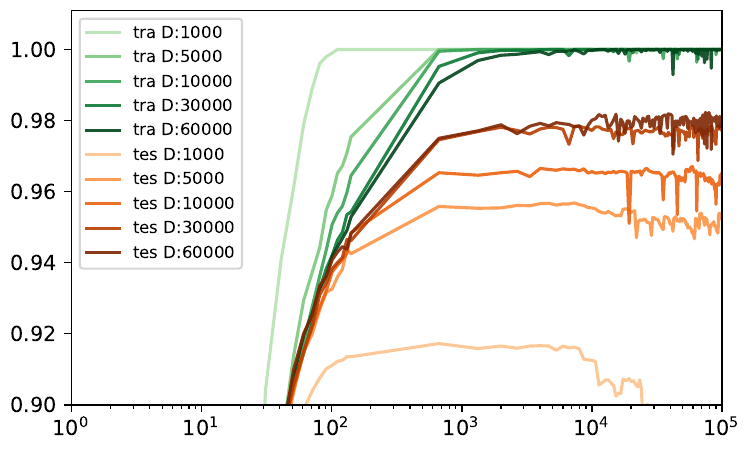}
    \caption{Training and Test accuracy for the different datset sizes presented in \cref{fig:mlp-dataswp}.}
    \label{fig:mlp-dataswp-accuracies}
\end{figure}

\begin{figure}[!b]
\begin{minipage}{0.03\linewidth}
\centering
    \rotatebox[]{90}{{\small{Accuracy}}}
\end{minipage}
\begin{minipage}{0.92\linewidth}
    \centering
    \includegraphics[width=.33\linewidth]{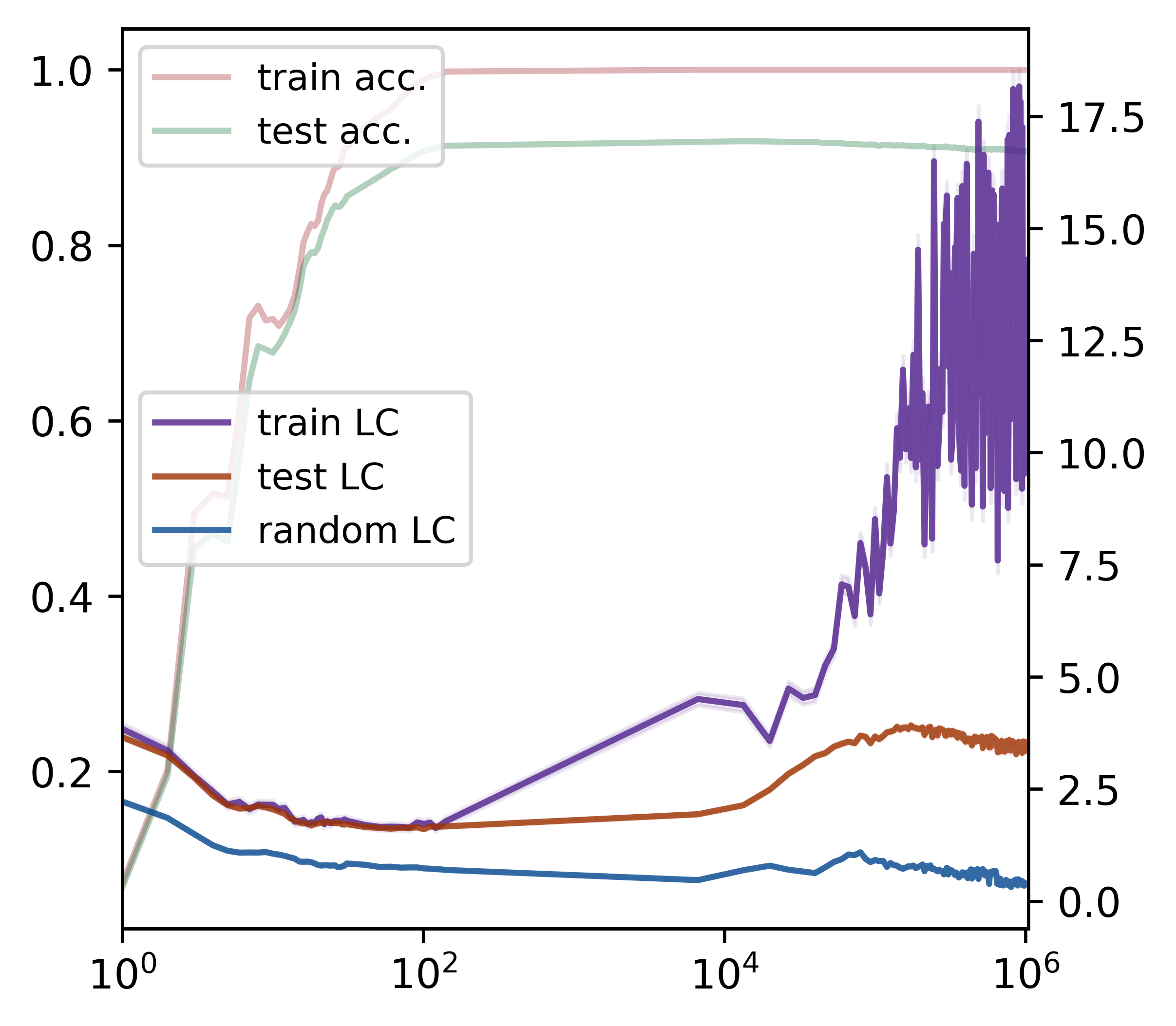}%
    \includegraphics[width=.33\linewidth]{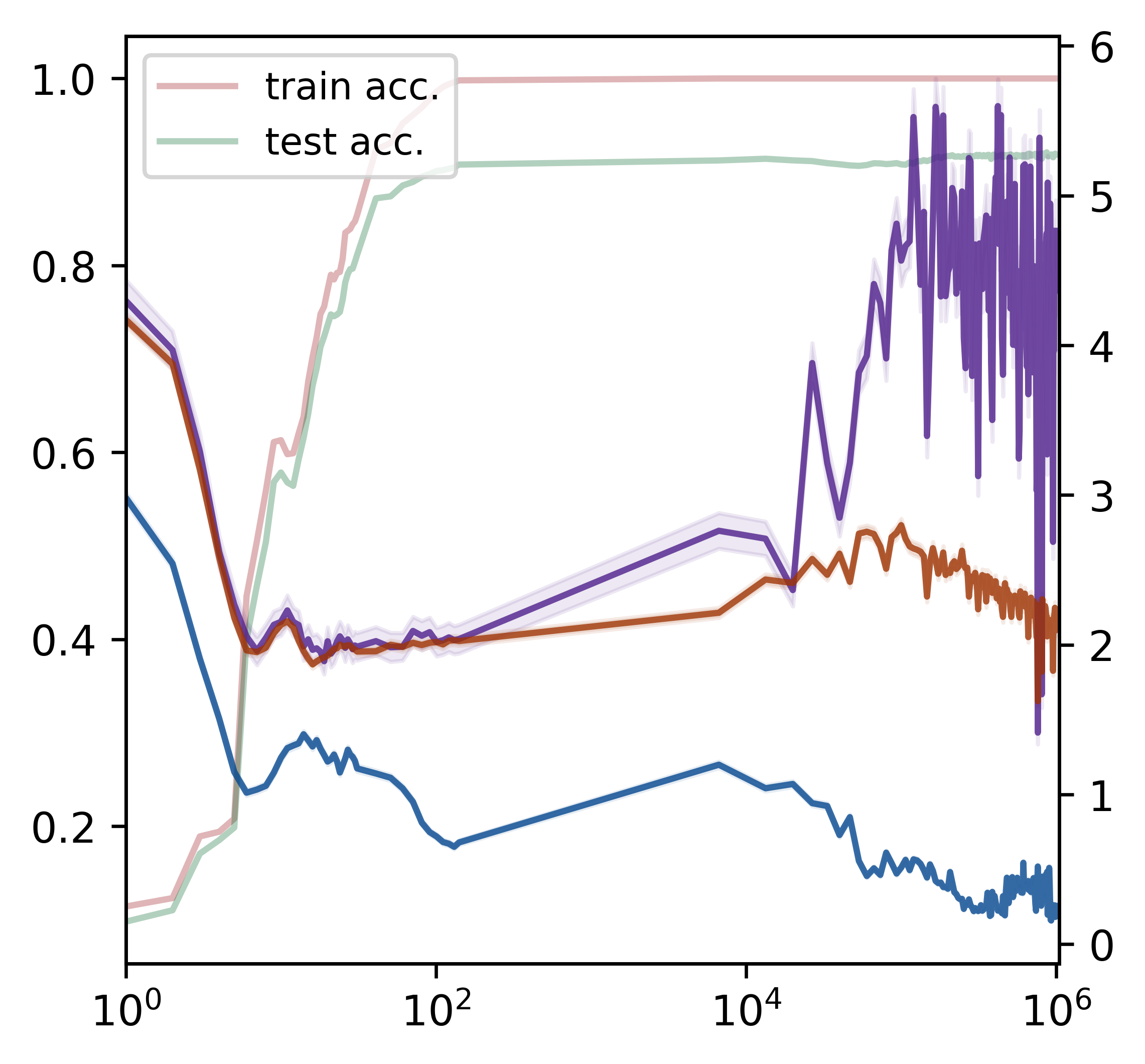}%
    \includegraphics[width=.33\linewidth]{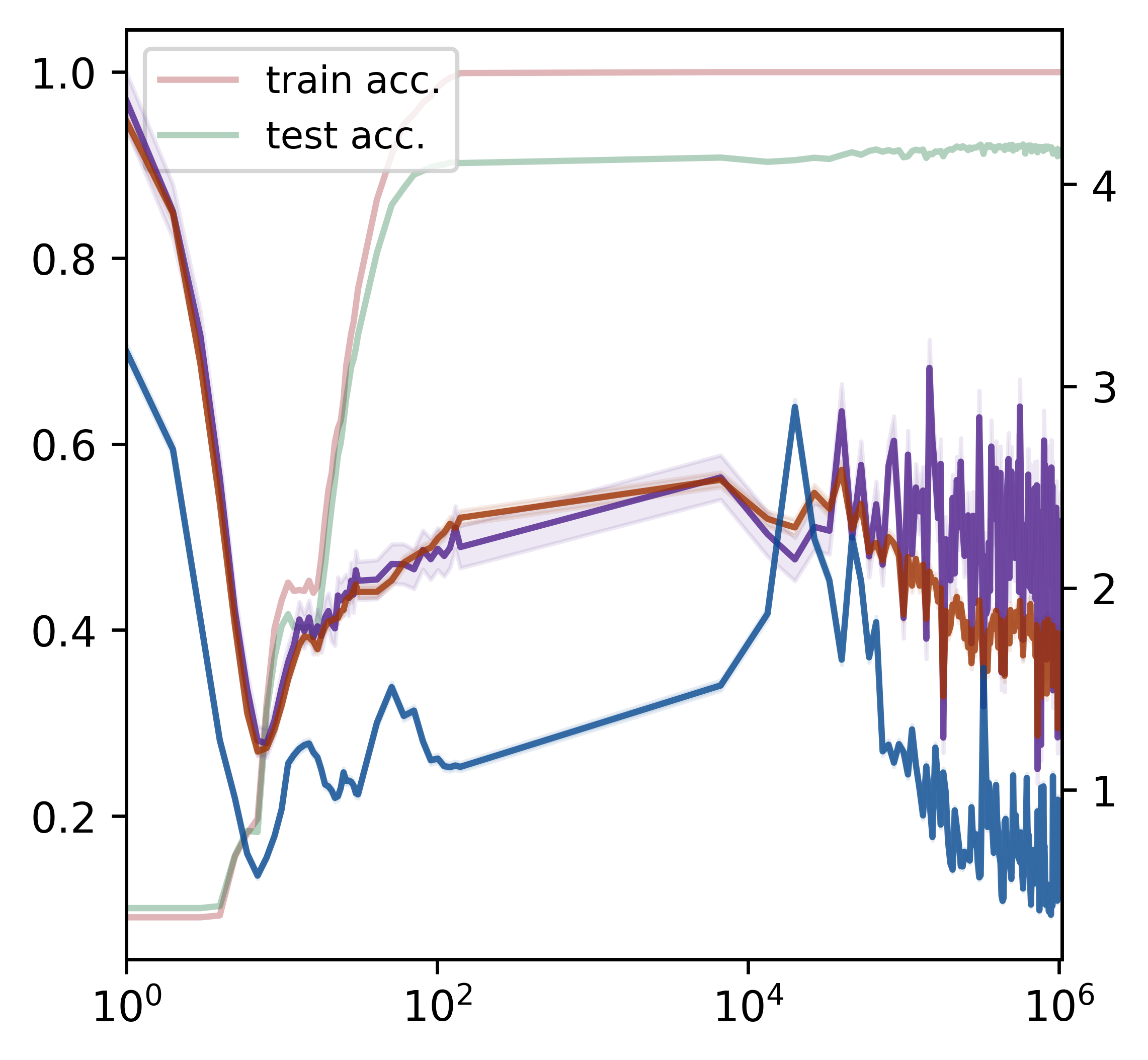}%
\end{minipage}
\begin{minipage}{0.03\linewidth}
    \centering
    \rotatebox[]{270}{{\small{Local Complexity}}}
\end{minipage}\\
\begin{minipage}{\linewidth}
    \centering
    Optimization Steps
\end{minipage}
     \caption{LC dynamics for a GeLU-MLP with width $200$ and depth $\{3,4,5\}$ presented from left to right. LC is calculated at $1000$ training points and $10000$ test and random points during training on MNIST.
     }
    \label{fig:gelu_mlp}
\end{figure}

\begin{figure}
\begin{minipage}{0.03\linewidth}
\centering
    \rotatebox[]{90}{{\small{Local Complexity}}}
\end{minipage}%
\begin{minipage}{.94\textwidth}
    \centering
    \includegraphics[width=0.2\linewidth]{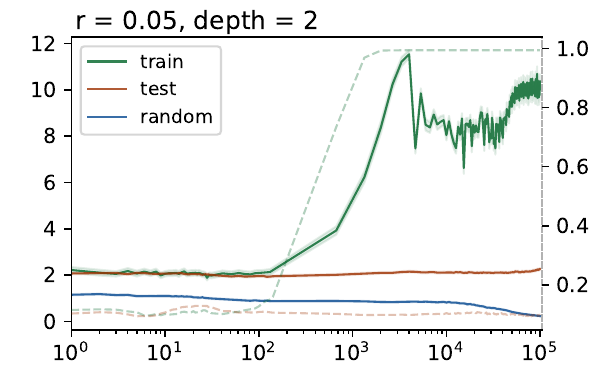}%
    \includegraphics[width=0.2\linewidth]{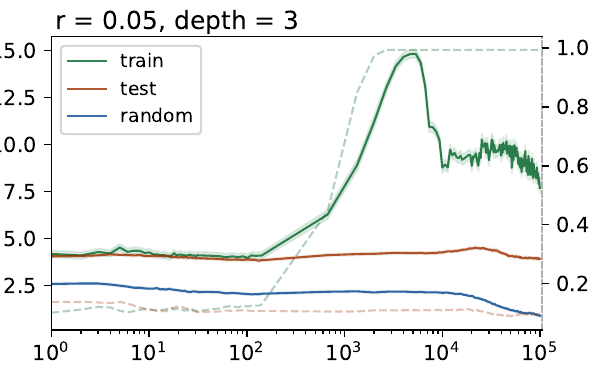}%
    \includegraphics[width=0.2\linewidth]{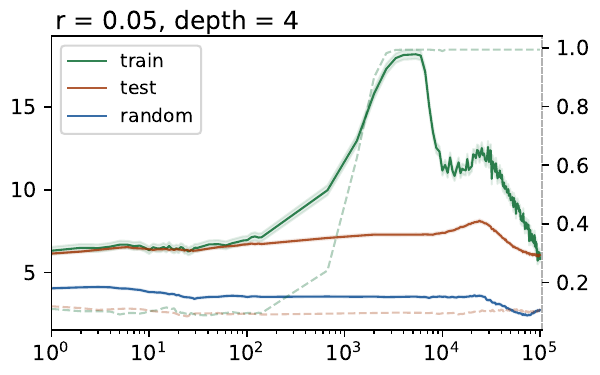}%
    \includegraphics[width=0.2\linewidth]{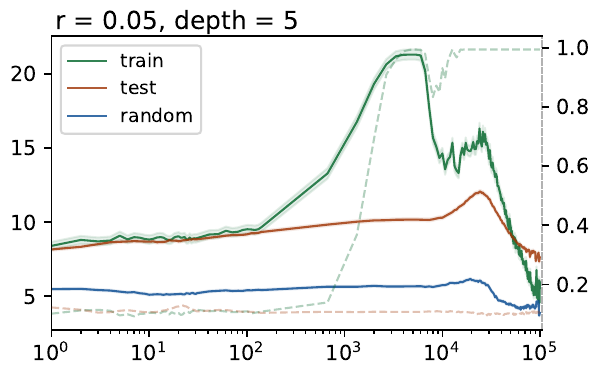}%
    \includegraphics[width=0.2\linewidth]{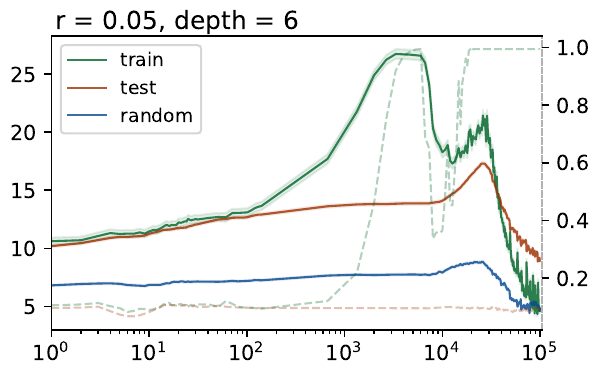}\\
    \includegraphics[width=0.2\linewidth]{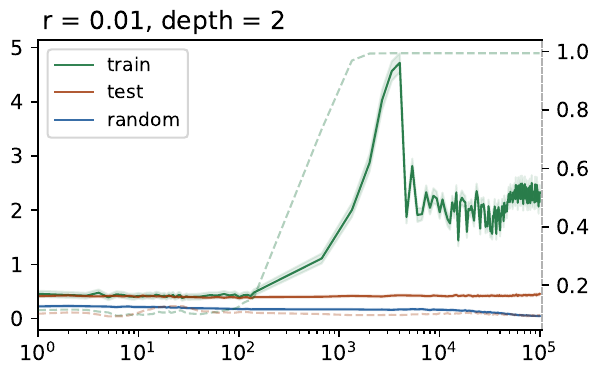}%
    \includegraphics[width=0.2\linewidth]{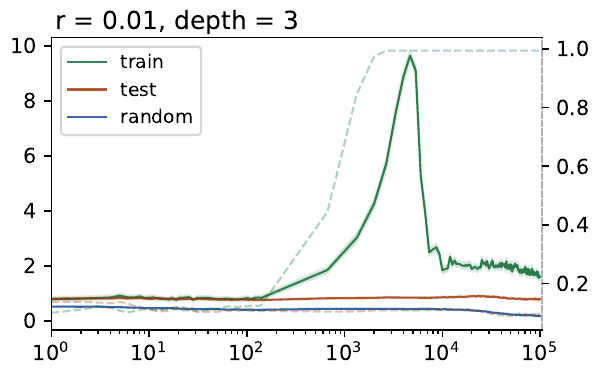}%
    \includegraphics[width=0.2\linewidth]{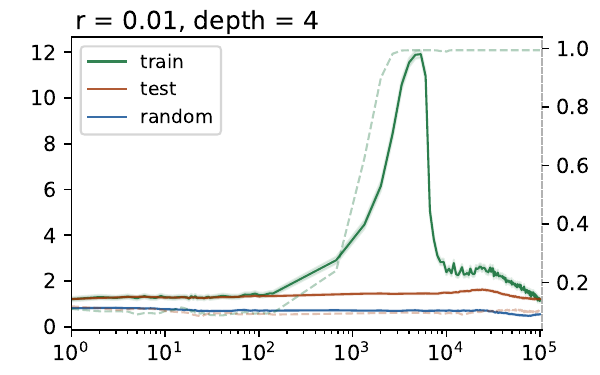}%
    \includegraphics[width=0.2\linewidth]{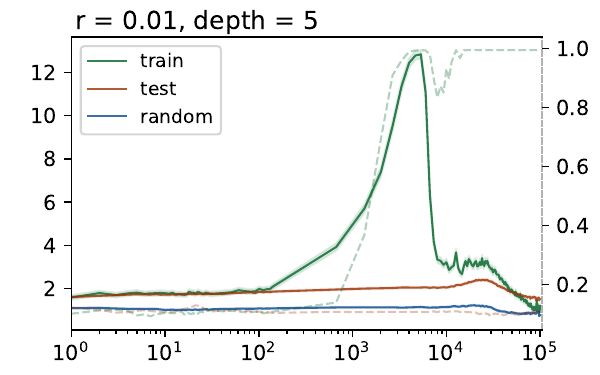}%
    \includegraphics[width=0.2\linewidth]{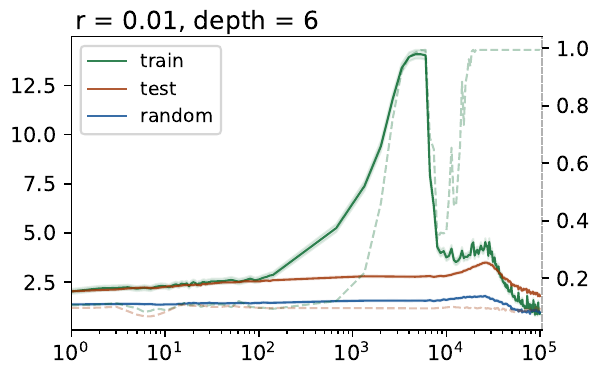}
\end{minipage}%
\begin{minipage}{0.03\linewidth}
\centering
    \rotatebox[]{270}{{\small{Accuracy}}}
\end{minipage}\\
\begin{minipage}{\linewidth}
    \centering
    \vspace{1em}
    Optimization Steps
\end{minipage}
    \caption{Random label radius and depth Sweep}
    \label{fig:mlp-randlbl-depth-sweep}
\end{figure}



\end{document}